\documentclass{article}

\def\inclapp{1}
\def\viewchanges{1}
\def\viewauthors{1}
\def\viewkeywords{0}
\def\usehyperlinks{1}
\def\addackn{1}

\usepackage[dvipsnames]{xcolor}

\usepackage{iclr2021_conference,times}
\if\viewauthors1
\iclrfinalcopy

\usepackage{natbib}

\usepackage[utf8]{inputenc} % allow utf-8 input
\usepackage[T1]{fontenc}    % use 8-bit T1 fonts
\usepackage{booktabs}       % professional-quality tables
\usepackage{amsfonts}       % blackboard math symbols
\usepackage{nicefrac}       % compact symbols for 1/2, etc.
\usepackage{microtype}      % microtypography
\usepackage{amsthm}
\usepackage{amsmath}
\usepackage{amssymb}
\usepackage{enumitem}
\usepackage{bbm}
\usepackage{url}            % simple URL typesetting

\usepackage{breakurl}
\if\usehyperlinks1
	\usepackage[colorlinks, citecolor=blue, linkcolor=magenta, breaklinks]{hyperref}       % hyperlinks
\fi

\usepackage{graphicx,wrapfig,lipsum}
\usepackage{subfigure}
\usepackage{verbatim}
\usepackage{bm}
\usepackage{adjustbox}
\usepackage{footnote}
\usepackage{algorithm}
\usepackage{algorithmic}
\usepackage{tikz}
\usepackage{cleveref}

\if\inclapp0
	\usepackage{xr}
\fi

\newtheorem{theorem}{Theorem}[section]
\newtheorem{lemma}[theorem]{Lemma}
\newtheorem{definition}[theorem]{Definition}

\newtheorem{prop}[theorem]{Proposition}
\newtheorem{rem}[theorem]{Remark}
\newtheorem{cor}[theorem]{Corollary}
\newtheorem{example}[theorem]{Example}
\newtheorem{assumption}{Assumption}

\crefname{assumption}{assumption}{assumptions}
\Crefname{assumption}{Assumption}{Assumptions}

\let\P\undefined%
\newcommand{\1}{\mathbf{1}}
\newcommand{\P}{\mathbb{P}}
\newcommand{\F}{\mathcal{F}}
\newcommand{\A}{\mathcal{A}}
\newcommand{\E}{\mathbb{E}} % expectation
\newcommand{\N}{\mathbb{N}} 
\newcommand{\R}{\mathbb{R}} 
\newcommand{\argmin}{\operatorname{argmin}} 
\newcommand{\id}{\operatorname{id}} 
 
\newcommand{\omb}{(\omega)}

\let\del\undefined
\let\com\undefined

\if\viewchanges1
	
	\newcommand{\del}[1]{{\color{red}{#1}}}
	\newcommand{\com}[1]{{\color{orange}{#1}}}
	
\else
	
	\newcommand{\del}[1]{}
	\newcommand{\com}[1]{}
	
\fi

\allowdisplaybreaks

\title{{Optimal estimation of generic dynamics by path-dependent neural jump ODEs}\thanks{This paper was published in Stochastic Processes and their Applications where it can be access via \url{https://doi.org/10.1016/j.spa.2026.105058}.}}

\if\viewauthors1
	\author{%
	  Florian Krach$^1$ \quad Marc Nübel%$^2$ 
	  \quad Josef Teichmann$^1$ \\
	  ${}^1$Department of Mathematics, 
	  ETH Zurich, Switzerland, \\ \texttt{\{firstname.lastname\}@math.ethz.ch}\\
	}
\else
	\author{}
\fi

\providecommand{\keywords}[1]{\textbf{{Keywords:}} \textit{#1}}

\begin{document}

\maketitle

\begin{abstract}
This paper studies the problem of forecasting general stochastic processes using a path-dependent extension of the Neural Jump ODE (NJ-ODE) framework \citep{herrera2021neural}.
While NJ-ODE was the first framework to establish convergence guarantees for the prediction of irregularly observed time series, these results were limited to data stemming from It\^o-diffusions with complete observations, in particular Markov processes, where all coordinates are observed simultaneously.
In this work, we generalise these results to generic, possibly non-Markovian or discontinuous, stochastic processes with incomplete observations, by utilising the reconstruction properties of the signature transform.
These theoretical results are supported by empirical studies, where it is shown that the path-dependent NJ-ODE outperforms the original NJ-ODE framework in the case of non-Markovian data.
Moreover, we show that PD-NJ-ODE can be applied successfully to classical stochastic filtering problems and to limit order book (LOB) data.
\end{abstract}

\if\viewkeywords1
\fi

%%% ==================================================================
\section{Introduction}\label{sec:Introduction}
The processing and prediction of time series data are of great importance in many data-driven fields, such as economics, finance, and medicine. In recent years, considerable progress has been made in improving machine learning techniques, particularly neural-network-based methods, to address more complicated problem settings.
Recurrent neural networks (RNNs) constituted the starting point for dealing with discrete time series of variable and possibly unbounded length.
Their main limitation is the underlying assumption that observations occur at regular time intervals. A first step towards handling irregular observation times was to define the RNN's latent variable continuously in time, with some time decay (e.g. exponential decay) towards $0$ \citep{che2018recurrent, cao2018brits}. However, since this framework is rather rigid, neural ODEs \citep{chen2018neural} marked a new milestone by making the continuous-time latent dynamics trainable through a neural network. Finally, combining this trainable continuous-time latent framework with an RNN cell led to a framework for irregularly sampled time series data \citep{ODERNN2019, Brouwer2019GRUODEBayesCM}.

In machine learning applications to time series data, we distinguish between two different problem settings. First, in the \emph{labelling problem}, the entire time series is processed to determine a class or value describing some feature of this time series.
An example is a time series consisting of the health parameters of a hospital patient, with the aim of predicting whether the patient will develop a certain disease within the next few days.
Second, in the \emph{forecasting problem}, the goal is to process the known past values of the time series to predict how it will develop in the future. If the entire time series of a predefined length is processed and values at certain future time points are predicted, this can be viewed as a special case of the labelling problem (with a possibly infinite-dimensional output). We refer to this as \emph{offline forecasting}. By contrast, if the goal is to forecast continuously in time, such that at every time point a prediction can be made based only on past observations, this constitutes a different problem, which we refer to as \emph{online forecasting}. Here, the algorithm must dynamically generate predictions based on the known past observations while no new information is available and then process new observations (and adjust itself accordingly) whenever they become available.
Returning to our previous example, this would mean continuously forecasting how the health parameters will evolve, given the observations made up to the current time.
In this work, which is the second part and generalisation of the Neural Jump ODE (NJ-ODE) framework \citep{herrera2021neural}, we consider precisely this problem. In particular, our goal is to make optimal forecasts, where optimality in this work is  meant in terms of the $L^2$-norm.

While NJ-ODE was the first framework for which theoretical guarantees of convergence of the model output to the optimal prediction were derived, relatively restrictive assumptions on the underlying dynamics of the time series data were needed. In particular, the data must stem from an It\^o diffusion subject to several constraints on its drift and diffusion. This implies that the paths in the dataset are continuous and Markovian (without path dependence). Moreover, the observations must be complete, i.e., all coordinates must be observed simultaneously at each observation time.
In the present work, we extend the NJ-ODE such that it can be applied to a very general set of stochastic processes that satisfy only weak regularity conditions.
Since some of these constraints might seem a bit abstract at first, we provide several examples for which we prove that the assumptions are satisfied.
Some examples are processes with jumps, fractional Brownian motion (rough paths with path-dependence) and multidimensional correlated processes with incomplete observations.
Moreover, we show how the NJ-ODE framework can be used to perform not only prediction but also uncertainty estimation and how it can be applied to the stochastic filtering problem.
Our theoretical results are based on the universal approximation property of neural networks and the signature transform.

\subsection{Related work}\label{sec:Related Work}
Recurrent neural networks \citep{rumelhart1985learning, jordan1997serial} and the neural ODE \citep{chen2018neural} are the two main ingredients for the (path-dependent) NJ-ODE model.
The first works in which they were combined in a model similar to ours were \citet{ODERNN2019, Brouwer2019GRUODEBayesCM}.
In contrast to our framework, the latent ODE  \citep{ODERNN2019} is a model for the offline forecasting problem, where an encoder-decoder type model is used. First, the ODE-RNN encoder processes the entire time series of observations to generate an initial latent variable, which is then used as starting point for a neural ODE that produces the forecasts.
By comparison, GRU-ODE-Bayes \citep{Brouwer2019GRUODEBayesCM} uses the same model framework as we do for online forecasting. The main difference from NJ-ODE lies in the objective function and training framework. In particular, no convergence guarantees exist for GRU-ODE-Bayes, as discussed in detail in \citet{herrera2021neural}.

Being the predecessor, NJ-ODE \citep{herrera2021neural} clearly is the most related work to the present one. While the main structure of the model and of the theoretical results is the same, we make many important changes to extend the theory from a class of It\^o processes to a large class of generic stochastic processes.
A major ingredient in this extension is the signature transform \citep{Chevyrev2016APO, KiralyOberhauser2019, fermanian2020embedding}, which allows us to approximate path-dependent behaviour.

The most related work in the context of the labelling problem, besides \citet{ODERNN2019}, is the neural controlled differential equation (NCDE) \citep{Kidger2020NeuralCD}. Similar to neural ODEs, it integrates a neural network, however, not against time but against the linear or spline interpolation of the observed time series.
The NCDE can only be used for the labelling problem (including offline forecasting), but not for the online forecasting problem, since its interpolation of the observations depends on future data (cf.\ \citet{morrill2022on}). Hence, in general its output at an intermediate time $t$ is not measurable with respect to the information available at this time. In \citet{morrill2022on}, the NCDE was extended by using the rectilinear (instead of linear or spline) interpolation of the data to circumvent this problem and make it applicable to online forecasting tasks.
Nevertheless, the authors applied it only to labelling problems (noting that these labelling problems can then be addressed online), rather than to the type of online prediction task of greatest interest here, in which the value of a stochastic process is predicted from previous discrete observations of that process. Accordingly, no convergence guarantees are provided for such problems.

Neural rough differential equations (NRDE) \citep{morrill2021neural} are yet another extension of NCDEs, where a neural network is piecewise integrated on intervals against time and multiplied with the depth-$N$ log-signature transform computed over the respective interval.
The advantage of this method over NCDEs is that the intervals can be chosen to be larger than those between consecutive observation times, since the log-signature can compress the path information over the entire interval. In particular, no information is lost, whereas increasing the step size of the NCDE model, which corresponds to subsampling the data, does result in information loss. Therefore, NRDEs are well suited to labelling problems involving long time series (by choosing larger integration intervals), for which NCDEs exhibit worsening accuracy and prohibitively long training times. However, like the original NCDE model, the NRDE method is not suitable for online forecasting, since its output at an intermediate time $t$ is generally not measurable with respect to the information available at that time.

While all of the previously discussed frameworks are prediction models that are deterministic once they are trained, neural stochastic differential equations (NSDEs) \citep{tzen2019neural, li2020scalable} are rather generative models with a standardized stochastic input generating stochastic output. In particular, NSDEs produce sample paths of a stochastic process, which can be useful when either stochastic samples are needed or if the generation of samples is easier than the computation of the distribution of the process.
However, training generative models is usually more complicated than training prediction models. In particular, NSDEs can be interpreted as (infinite-dimensional) GANs \citep{kidger2021neural}, which are known to be difficult to train (with vanishing gradients, mode collapse, and failure to converge being the most common problems) \citep{saxena2021generative}.
Monte Carlo techniques can be used to apply NSDEs when the main interest is the distribution of the underlying process. The disadvantage of this approach is the comparatively long computation time caused by the need for many independent samples of the NSDE to obtain reasonably small Monte Carlo errors. By contrast, in Section~\ref{sec:Conditional Variance, Moments and Moment Generating Function}, we explain how the NJ-ODE framework can be used to predict the conditional distribution more directly (without sampling).

\subsection{Outline of the work}\label{sec:Outline}
First, we establish the problem setting in Section~\ref{sec:Problem Setting}, where we outline which information is available to learn from and what assumptions are needed. Moreover, we show that the theoretically optimal solution (cf. Section~\ref{sec:Optimal Approximation}) to the online forecasting problem, which we aim to approximate, is given by the conditional expectation process (cf. Section~\ref{sec:Notation and assumptions}).
In Section~\ref{sec:proposed method}, we introduce our model as a signature based extension of NJ-ODE \citep{herrera2021neural} for which we prove in Section~\ref{sec:Convergence Guarantees} that it converges to the theoretically optimal solution. This proof is based on the respective proof for the NJ-ODE model.
In Section~\ref{sec:Conditional Variance, Moments and Moment Generating Function} we explain how to bridge from learning only the conditional expectation to approximating the conditional distribution, using the same framework.
We continue by explaining how our model can be applied to the stochastic  filtering problem in Section~\ref{sec:Stochastic Filtering with PD-NJ-ODE}.

Although the needed assumptions formulated in Section~\ref{sec:Notation and assumptions} are weak, they are a bit technical and might seem hard to be verified in concrete problems. Therefore, we provide several  examples of processes for which we prove that they satisfy these assumptions in Section~\ref{sec:Examples of Processes Satisfying the Assumptions}.
In particular we show that the setting used in \citet{herrera2021neural} is a special case that satisfies our assumptions here, justifying that we speak of a generalisation of \citet{herrera2021neural}.
In Section~\ref{sec:Experiments}, we provide empirical evidence that our model works well, with experiments performed on synthetic datasets based on some of the examples of Section~\ref{sec:Examples of Processes Satisfying the Assumptions} as well as on real-world datasets.
Additional details are given in the appendices.

\section{Problem setting}\label{sec:Problem Setting}
We assume that we have a dataset of time series samples comprising irregular, possibly incomplete, discrete-time observations of a continuous-time stochastic process, where the observation times and observed coordinates are random.
Importantly, we do not assume to have any knowledge about the underlying  stochastic process or the observation times and masks, except for the data we observe and that they satisfy the assumptions formulated in Section~\ref{sec:Notation and assumptions}. In particular, we do not assume to know the distribution of the stochastic process or the distribution of the observation times and masks.
The goal is to use the given data to train a model such that it approximates the optimal solution of the online forecasting problem (cf. Section~\ref{sec:Introduction}), which is shown to be given by the conditional expectation process (cf. Section~\ref{sec:Optimal Approximation}).
In the following, we give precise definitions and the assumptions needed to establish our theoretical guarantees, following the descriptions in \citet{herrera2021neural}.

\subsection{Stochastic process, random observation times and observation mask}
\label{sec:Stochastic Process, Random Observation Times and Observation Mask}
Let $d_X \in \N$ be the dimension and $T > 0$ be the fixed time horizon. Consider a filtered probability space  $(\Omega, \F, \mathbb{F} := \{\F_t\}_{0 \leq t \leq T}, \P )$, on which an adapted c\`adl\`ag  stochastic process\footnote{A stochastic process is a collection of random variables $X_t : \Omega \to \R^{d_X}, \omega \mapsto X_t(\omega)$ for $0 \leq t \leq T$.} $X :={(X_t)}_{t \in [0,T]}$ taking values in $\R^{d_X}$ is defined.
We define the running-maximum process
\[X^\star_t := \sup_{0 \leq s \leq t} |X_s|_1, \quad 0 \leq t \leq T,\]
where $|\cdot|_p$ denotes the standard $p$-norm for vectors.
Moreover, let $\mathcal{J}$ be the random set of discontinuity times of $X$, defined for every $\omega \in \Omega$ as $\mathcal{J}(\omega) := \{ t \in [0,T] | \Delta X_t(\omega) \neq 0 \}$.
Here, $\Delta X_t := X_t - X_{t-}$ is the jump of the process $X$ at time $t$, where $X_{t-} := \lim_{\epsilon \downarrow 0 } X_{t-\epsilon}$ denotes the left limit of $X$ at time $t$ (or equivalently, the value of the left-continuous version of $X$ at time $t$).

We consider another filtered probability space  $(\tilde\Omega, \tilde\F,  \tilde{\mathbb{F}} := \{ \tilde{\F}_t \}_{0 \leq t \leq T}, \tilde\P )$, on which the random observation framework of the stochastic process is defined. In particular, we define the following objects.
\begin{itemize}
\item $n: \tilde\Omega \to \N_{\geq 0}$, an $\tilde{\F}$-measurable random variable, is the random number of observations.
\item $K := \sup \left\{k \in \N \, | \, \tilde\P(n \geq k) > 0 \right\} \in \N \cup\{\infty\}$ is the maximal value of $n$.
\item  $t_i: \tilde\Omega \to [0,T] \cup \{ \infty \}$ for $0 \leq i \leq K$ are the \emph{sorted}\footnote{For all $\tilde\omega \in \tilde\Omega$, $0=t_0 < t_1(\tilde\omega) < \dotsb < t_{n(\tilde\omega)}(\tilde\omega) \leq T$.} stopping times,\footnote{In particular, $t_i$ is a random variable s.t. $\{ t_i \leq t\} \in \tilde{\F}_t$ for all $1 \leq i \leq n$ and $t \in [0,T]$.} describing the random observation times, with $t_i(\tilde{\omega}) = \infty$ if and only if $n(\tilde{\omega}) < i$.
\item $\tau : [0,T] \times \tilde\Omega \to [0,T], \quad (t, \tilde\omega) \mapsto \tau(t, \tilde\omega) := \max\{ t_i(\tilde\omega) | 0 \leq i \leq n(\tilde\omega), t_i(\tilde\omega) \leq t \}$ is the time of the last observation before a certain time $t$.
\item $M = (M_k)_{0 \leq k \leq K}$ is the observation mask, which is a sequence of random variables on  $(\tilde\Omega, \tilde\F, \tilde\P )$ taking values in $\{ 0,1 \}^{d_X}$ such that $M_k$ is $\tilde{\mathcal{F}}_{t_k}$-measurable.
The $j$th coordinate of the $k$th element of the sequence $M$, i.e., $M_{k,j}$, indicates whether $X_{t_k, j}$, the $j$th coordinate of the stochastic process at observation time $t_k$, is observed. In particular, $M_{k,j} = 1$ means that it is observed, whereas $M_{k,j} = 0$ means that it is not. By abuse of notation, we also write $M_{t_k} := M_{k}$.
\end{itemize}

\subsection{Information \texorpdfstring{$\sigma$}{sigma}-algebra}\label{sec:information sigma-algebra}
$(\Omega \times \tilde\Omega , \F \otimes \tilde\F, \mathbb{F} \otimes \tilde{\mathbb{F}}, \P \times \tilde\P)$ is the filtered product probability space which, intuitively speaking, combines the randomness of the stochastic process with the randomness of the observations.
Here, $\mathbb{F} \otimes \tilde{\mathbb{F}}$ consists of the tensor-product $\sigma$-algebras $(\F \otimes \tilde\F)_t := \F_t \otimes \tilde\F_t$ for $t \in [0,T]$.
On this probability space, we define the filtration of the currently available information $\mathbb{A} := (\mathcal{A}_t)_{t \in [0,T]}$ by
\begin{equation*}
\mathcal{A}_t := \boldsymbol{\sigma}\left(X_{t_i, j}, t_i, M_{t_i} | t_i \leq t,\, j \in \{1 \leq l \leq d_X | M_{t_i, l} = 1  \} \right),
\end{equation*}
where  $t_i$ are the random observation times and $\boldsymbol\sigma(\cdot)$ denotes the generated $\sigma$-algebra.
By the definition of $\tau$ we have $\mathcal{A}_t = \mathcal{A}_{\tau(t)}$ for all $t \in [0,T]$.
Moreover, we have for any fixed observation (stopping) time $t_k$ that the stopped and pre-stopped $\sigma$-algebras at $t_k$ are \citep[Definitions~2.37 and~8.1]{KarandikarRao2018}
\begin{equation*}
\begin{split}
\mathcal{A}_{t_k} &:= \boldsymbol{\sigma}\left(X_{t_i, j}, t_i, M_{t_i} \,\middle|\, i\leq k,\, j \in \{1 \leq l \leq d_X | M_{t_i, l} = 1  \} \right), \\
\mathcal{A}_{t_k-} &:= \boldsymbol{\sigma}\left(X_{t_i, j}, t_i, M_{t_i}, t_k \,\middle|\, i < k,\, j \in \{1 \leq l \leq d_X | M_{t_i, l} = 1  \} \right).
\end{split}
\end{equation*}

\subsection{Notation and assumptions on the stochastic process \texorpdfstring{$X$}{X}}\label{sec:Notation and assumptions}
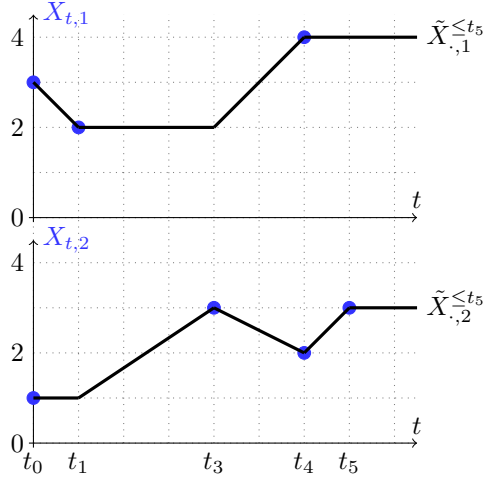
\begin{figure}
    \centering
    \begin{tikzpicture}[scale=.60]
		\draw[gray, dotted, step=1.0] (0,0) grid (8.5,5);
		\draw[->] (-0.1,0) -- (8.5,0);
		\draw[->] (0,-0.1) -- (0,4.5);

        \draw[gray, dotted, step=1.0] (0,5) grid (8.5,9.5);
		\draw[->] (-0.1,5) -- (8.5,5);
		\draw[->] (0,4.9) -- (0,9.5);

        \node[below] at (0,0) {$t_0$};
        \node[below] at (1,0) {$t_1$};
        \node[below] at (4,0) {$t_3$};
        \node[below] at (6,0) {$t_4$};
        \node[below] at (7,0) {$t_5$};

        % nodes
		\node[above] at (8.5,0) {$t$};
		\node[right, blue!80] at (0,4.5) {$X_{t,2}$};
        \node[above] at (8.5,5) {$t$};
		\node[right, blue!80] at (0,9.5) {$X_{t,1}$};

		% \foreach \i in {0,2,...,8} {
		% 	\pgfmathsetmacro\result{5*\i};
		% 	\node[below] at (\i,0) { \pgfmathprintnumber{\result}};
		% }
		\foreach \i in {0,2,4} {
			\pgfmathsetmacro\result{\i};
			\node[left] at (0,\i) { \pgfmathprintnumber{\result}};
		}
        \foreach \i in {0,2,4} {
			\pgfmathsetmacro\result{\i+5};
			\node[left] at (0,\result) { \pgfmathprintnumber{\i}};
		}

        % coordinate 1
        \filldraw [blue!80] (0,8) circle (4pt);
        \filldraw [blue!80] (1,7) circle (4pt);
        \filldraw [blue!80] (6,9) circle (4pt);
        % coordinate 2
        \filldraw [blue!80] (0,1) circle (4pt);
        \filldraw [blue!80] (4,3) circle (4pt);
        \filldraw [blue!80] (6,2) circle (4pt);
        \filldraw [blue!80] (7,3) circle (4pt);
        
        % interpolation coordinate 1
        \draw[very thick] (0,8) -- (1,7);
        \draw[very thick] (1,7) -- (4,7);
        \draw[very thick] (4,7) -- (6,9);
        \draw[very thick] (6,9) -- (8.5,9);
        % interpolation coordinate 2
        \draw[very thick] (0,1) -- (1,1);
        \draw[very thick] (1,1) -- (4,3);
        \draw[very thick] (4,3) -- (6,2);
        \draw[very thick] (6,2) -- (7,3);
        \draw[very thick] (7,3) -- (8.5,3);

        \node[right] at (8.5,9) {$\tilde X^{\leq t_5}_{\cdot, 1}$};
        \node[right] at (8.5,3) {$\tilde X^{\leq t_5}_{\cdot, 2}$};

	\end{tikzpicture}
    \caption{Schematic representation of the interpolated observation path $\tilde X^{\leq t}$ of a two dimensional process $X$ with discrete and incomplete observations (blue dots) at the observation times $t_i$.}
    \label{fig:interpolated observation path}
\end{figure}

We denote the conditional expectation process of $X$ by $\hat X = (\hat X_t)_{0 \leq t \leq T}$, defined by $\hat{X}_t := \E_{\P\times\tilde\P}[X_t | \A_t]$, and note that, in contrast to the setting in \citet{herrera2021neural}, $\hat X_{\tau(t)} \ne X_{\tau(t)}$ in general, since observations are incomplete.
Moreover, we define
for any $0 \leq t \leq T$ the process $\tilde X^{\leq t}$ to be the
interpolation of the observations of $X$ made until time $t$. Its $j$th coordinate at time $0 \leq s \leq T$ is given by
\begin{equation*}
\tilde X^{\leq t}_{s,j} := \begin{cases}
	X_{t_{a(s,t,j)},j} \frac{t_{b(s,t,j)} - s}{t_{b(s,t,j)} - t_{b(s,t,j)-1}} + X_{t_{b(s,t,j)},j} \frac{s - t_{b(s,t,j)-1}}{t_{b(s,t,j)} - t_{b(s,t,j)-1}}, & \text{if }  t_{b(s,t,j)-1} < s \leq t_{b(s,t,j)}, \\
	X_{t_{a(s,t,j)},j}, & \text{if } s \leq t_{b(s,t,j)-1} ,
\end{cases}
\end{equation*}
where
\begin{equation*}
\begin{split}
a(s,t, j) &:=  \max\{ 0 \leq a \leq n \vert t_a \leq \min(s,t), M_{t_a,j}  = 1 \}, \\
b(s,t,j) &:=  \inf\{ 1 \leq b \leq n \vert s \leq t_b \leq t, M_{t_b,j}  = 1 \},
\end{split}
\end{equation*}
with the standard definition that the infimum of the empty set is $\infty$ and the additional definition that $t_\infty := T$.
A schematic representation of this \emph{interpolated observation path} is given in Fig.~\ref{fig:interpolated observation path}.
In particular, $\tilde X^{\leq t}$ is the rectilinear interpolation (sometimes denoted as forward-fill), except that its jumps at $t_{b(s,t,j)}$ are replaced by linear interpolations between the previous observation time $t_{b(s,t,j)-1}$ and $t_{b(s,t,j)}$. It is important to note that this is not merely a coordinate-wise interpolation, since the given coordinate might not have been observed at the previous observation time. In particular,  $b(s,t,j)-1 \ne a(s,t,j)$ in general. Moreover, by this definition, $\tilde X^{\leq \tau(t)}$ is $\mathcal{A}_{\tau(t)}$-measurable for all $t$, and for any $r \geq t$ and all $s \leq \tau(t)$ we have $\tilde X^{\leq \tau(t)}_s = \tilde X^{\leq \tau(r)}_s$.
We denote the set of continuous $\R^{d_X}$-valued paths of bounded variation (cf. Definition~\ref{def:p-variation}) on $[0,T]$ by $BV^c([0,T])$. By the definition of $\tilde X^{\leq t}$ it is clear that its paths belong to $BV^c([0,T])$.

Using $\mathcal{A}_t = \mathcal{A}_{\tau(t)}$ and the fact that $\tilde X^{\leq \tau(t)} \in \mathcal{A}_{\tau(t)}$ carries all available information\footnote{More precisely, this is only true if the probability of any two consecutive observations being equal is zero. Otherwise, one can add coordinates to $\tilde{X}^{\leq t}$ that describe the coordinate-wise number of observations as paths that increase by $1$ whenever a new observation is made for the corresponding coordinate. The same interpolation as for $\tilde{X}^{\leq t}$ can be used to make these paths continuous.}
in $\mathcal{A}_{\tau(t)}$, we know by the Doob-Dynkin Lemma (\citet[Lemma~1.14]{kallenberg2021foundations} or \citet[Lemma~2]{taraldsen2018optimal}) that there exist measurable functions $F_j : [0,T] \times [0,T] \times BV^c([0,T]) \to \R$ such that $\hat X_{t,j} = F_j(t, \tau(t), \tilde X^{\leq \tau(t)}) $. The following assumptions are central to establish our theoretical results. Some examples where these assumptions are satisfied are given in Section~\ref{sec:Examples of Processes Satisfying the Assumptions}.
\begin{assumption} \label{assumption:1}
For every $1\leq k, l \leq K$, $M_k$ is independent of  $t_l$ and of $n$,  $\tilde \P (M_{k,j} =1 ) > 0$ and $M_{0,j}=1$ for all  $1 \leq j \leq d_X$  (every coordinate can be observed at any observation time and $X$ is completely observed at $0$) and $|M_k|_1 > 0$ for every $1 \leq k \leq K$ $\tilde{\P}$-almost surely (at every observation time at least one coordinate is observed).
\end{assumption}
\begin{assumption} \label{assumption:2}
The probability that any two observation times are closer than $\epsilon>0$ converges to $0$ when $\epsilon$ does, i.e., if $\delta(\tilde \omega) := \min_{0 \leq i \leq n(\tilde \omega)} |t_{i+1}(\tilde \omega) - t_i(\tilde \omega)|$ then $\lim_{\epsilon \to 0} \tilde \P (\delta < \epsilon) = 0$.
\end{assumption}
\begin{assumption} \label{assumption:3}
Almost surely $X$ is  not observed at a jump, i.e., $(\P \times \tilde{\P})( t_j \in \mathcal{J} | j\leq n ) = (\P \times \tilde{\P})( \Delta X_{t_j} \neq 0 | j\leq n) = 0$ for all $ 1 \leq j \leq K$.
\end{assumption}
\begin{assumption} \label{assumption:4}
$F_j$ are  continuous and differentiable in their first coordinate $t$ such that their partial derivatives with respect to $t$, denoted by $f_j$, are again continuous and there exists a $B >0$ and $p \in \N_{\geq 1}$ such that for every $t \in [0,T]$ the functions $f_j, F_j$ are polynomially bounded in $X^\star$, i.e.,
\begin{equation}\label{equ:assumption4 bound}
|F_j(\tau(t), \tau(t), \tilde X^{\leq \tau(t)})| +  | f_j(t, \tau(t), \tilde X^{\leq \tau(t)})  | \leq B (X_t^\star +1)^p .
\end{equation}
\end{assumption}
\begin{assumption} \label{assumption:5}
$X^\star$ is $L^{2 p}$-integrable, i.e., $\E[(X^\star_T)^{2 p}] < \infty$.
\end{assumption}
\begin{assumption} \label{assumption:6}
The random number of observations $n$ is integrable, i.e., $\E_{\tilde\P}[n] < \infty$.
\end{assumption}

\begin{rem}
Assumption~\ref{assumption:2} is a technical condition that is needed to get compact subsets of $BV([0,T])$. It is actually equivalent to $\tilde \P (\delta = 0) = 0$, which is always satisfied when the observation times are strictly increasing.
\end{rem}
\begin{rem}\label{rem:extension assumptions M0}
The assumption that $X_0$ is observed completely, i.e., that $M_0 = 1$, can be weakened (at least) in two possible ways.
\begin{enumerate}
\item We assume instead that $X_0$ is not observed at all, i.e., $M_0 = 0$.
\item We assume instead that a fixed subset of coordinates $I_0 \subseteq \{ 1, \dotsc, d_X \}$ is always observed at $t=0$ and the others not, i.e., $M_{0, j} = \1_{I_0}(j)$.
\end{enumerate}
In both cases, the definition of $\tilde{X}^{\leq t}$ can easily be adjusted accordingly.
\end{rem}

With Assumption~\ref{assumption:4}  we can rewrite $\hat X$ by the fundamental theorem of calculus as
\begin{equation*}
\hat X_{t,j} = F_j(\tau(t), \tau(t), \tilde X^{\leq \tau(t)}) + \int_{\tau(t)}^t f_j(s, \tau(t),  \tilde X^{\leq \tau(t)}) ds ,
\end{equation*}
implying that it is c\`adl\`ag. We remark that jumps of $\hat X$ occur only at new observation times, i.e., at $t_i$, for $1 \leq i \leq n$.

\begin{rem}
In principle, the assumption that $F_j$ are continuous would be enough to use neural networks to approximate them.
The reason why we make the stronger assumption that $f_j$ exist and are continuous is that by subsequently rewriting  $\hat{X}$ with the fundamental theorem of calculus, we can make use of additional domain knowledge which simplifies the learning task. For more details see Appendix~\ref{sec:Direct Neural Network Approximating of the Conditional Expectation}.
\end{rem}

\begin{rem}
    The assumption that $F_j$ and $f_j$ are bounded by $B (X_t^\star +1)^p$, together with the assumption that $X^\star$ is $L^{2 p}$-integrable, ensures that our loss function (defined later) is well defined and does not explode. In particular, these assumptions are crucial for showing that the maximal distance between $X$ and our model's approximation of it is integrable, which is needed to prove convergence of our method.
\end{rem}

\subsection{Optimal approximation}\label{sec:Optimal Approximation}
As in \citet{herrera2021neural}, we are interested in the $L^2$-optimal approximation of $X_t$ at any time $t \in [0,T]$ given the  currently available information $\mathcal{A}_t$. The following result was proven in  \citet[Proposition B.1]{herrera2021neural} and shows that this approximation is given by the conditional expectation process $\hat X$.

\begin{prop}\label{prop:best estimator}
The optimal ($L^2$-norm minimizing) process in $L^2(\Omega \times \tilde\Omega, \mathbb{A} , \P \times \tilde\P)$ approximating $(X_t)_{t \in [0,T]}$ is given by\footnote{
While we gave a pointwise definition, \citet[Theorem 7.6.5]{cohen2015stochastic} allows $\hat X$ to be defined directly as the optional projection. By \citet[Remark 7.2.2]{cohen2015stochastic}, this implies that the process $\hat X$ is progressively measurable and, in particular, jointly measurable in $t$ and $\omega \times \tilde{\omega}$.
However, as we have seen, even from the pointwise definition, it follows that $\hat{X}$ is c\`adl\`ag, hence optional \citep[Theorem 7.2.7]{cohen2015stochastic}.
}
$\hat{X}$. Moreover, this process is unique up to $(\P \times \tilde\P)$-null-sets.
\end{prop}

\subsection{Push-forward measure of known information and implied metric}
\label{sec:Push-forward Measure of Known Information and Implied Metric}
In our analysis, the left limits of c\`adl\`ag processes at observation times play an important role. In the following, we define (pseudo)metrics for the processes of interest. We start by constructing probability measures and then define the (pseudo)metrics using their induced $L^1$-norm. We note that, to understand the rest of the paper, it is enough to consider Definition~\ref{def:indistinguishability} and Remark~\ref{rem:equivalent definition pseudo metric}; the other parts of this section can be skipped.

Fix any $r \in \N$ and $1 \leq k \leq K$; we use the notation $t_k-$ to denote the left limit at $t_k$.
Any $\mathbb{A}$-adapted process $Z : [0,T] \times (\Omega \times \tilde{\Omega}) \to \R^{r}$ can  be written as a function of $\tilde{X}^{\leq t}$ and $t$ (cf. Section~\ref{sec:Notation and assumptions}). By abuse of notation, we will use the same symbol to write
\begin{equation}\label{equ:equivalent notations Z}
Z_t(\omega, \tilde{\omega}) = Z(t, (\omega, \tilde{\omega})) = Z ( \tilde{X}^{\leq t} (\omega, \tilde{\omega}), t ) = Z ( \tilde{X}^{\leq t}, t )
\end{equation}
and we will simply write $Z$ whenever the context is clear.
Moreover, for any such process, we define $Z_\infty := 0$ so that $Z_{t_k}$ is well defined.
The available information at $t_k-$, i.e., just before new information becomes available, is given through $\tilde{X}^{\leq t_k-} := \lim_{\epsilon \downarrow 0}\tilde{X}^{\leq t_k-\epsilon} = \tilde{X}^{\leq t_{k-1}}$. Therefore, the process $Z$ at time $t_k-$ can be written as $Z_{t_k-} = Z(\tilde{X}^{\leq t_{k-1}}, t_k-) = \lim_{\epsilon \downarrow 0} Z(\tilde{X}^{\leq t_{k-1}}, t_k-\epsilon)$, which is standard notation for stochastic processes.
We define the joint push-forward measure
\[\lambda_k := \lambda_{(\tilde{X}^{\leq t_{k-1}}, t_k-)} := (\P \times \tilde{\P}) \circ (\tilde{X}^{\leq t_{k-1}}, t_k-)^{-1}\]
 through the following limit of the joint push-forward measures $\lambda_{k, \epsilon} := \lambda_{(\tilde{X}^{\leq t_{k-1}}, t_k-\epsilon)}$ for $\epsilon > 0$, which are probability measures on their induced image probability space with the sample space $\hat{\Omega}=BV^c([0,T])\times ([0,T]\cup\{ \infty\})$.
For any bounded function $Z: \hat{\Omega} \to \R^r$ for which left limits in terms of its second argument exist, we define
\begin{equation}\label{equ:def lambda_k}
\E_{\lambda_k}[Z]
	:= \lim_{\epsilon \downarrow 0} \E_{\lambda_{k, \epsilon}}[Z]
	= \lim_{\epsilon \downarrow 0} \E_{\P \times \tilde{\P}}\left[ Z(\tilde{X}^{\leq t_{k-1}}, t_k-\epsilon) \right]
	= \E_{\P \times \tilde{\P}}\left[ Z(\tilde{X}^{\leq t_{k-1}}, t_k-) \right] ,
\end{equation}
where we used \citet[Theorem 1.6.9]{Durrett:2010:PTE:1869916} for the second and dominated convergence in the third equality. It is easy to show that $\lambda_k$ satisfies the properties of a probability measure (i.e., that it integrates to $1$ and satisfies the additivity property) by choosing $Z$ to be the appropriate indicator function.
For any c\`adl\`ag $\mathbb{A}$-adapted process $Z$, for which $Z^\star$ is integrable, \eqref{equ:def lambda_k} holds similarly (by replacing boundedness with integrability of $Z^\star$ for dominated convergence). Moreover, by combining \eqref{equ:equivalent notations Z} and \eqref{equ:def lambda_k}, we then have for such a $Z$ that
\begin{equation}\label{equ:expectation lambda_k for Z}
\E_{\lambda_k}[Z]  = \E_{\P \times \tilde{\P}}\left[ Z(\tilde{X}^{\leq t_{k-1}}, t_k-) \right] = \E_{\P \times \tilde{\P}}\left[ Z_{t_k-}((\omega, \tilde{\omega})) \right]
= \E_{\P \times \tilde{\P}}\left[ Z_{t_k-} \right].
\end{equation}

In the following we are interested in the probability measure $\lambda_k$ conditioned on the event that $n \geq k$ (which is equivalent to $t_k \ne \infty$).
In particular, we define $\mu_k( \cdot ) := \lambda_k( \cdot | t_k- \ne \infty)$.

\begin{prop}\label{prop:mu_k expectation equiv}
Let $c_0 := c_0(k) := (\tilde{\P}(n \geq k))^{-1} $. For any c\`adl\`ag $\mathbb{A}$-adapted process $Z$, for which $Z^\star$ is integrable, we have
\begin{equation*}
\E_{\mu_k}[Z]  = c_0(k)\,  \E_{\P \times \tilde{\P}}\left[ \1_{\{n \geq k\}} Z(\tilde{X}^{\leq t_{k-1}}, t_k-) \right].
\end{equation*}
\end{prop}
\begin{proof}
Let $B \subseteq BV^c([0,T]) \times [0,T]$ be measurable such that its pre-image under $(\tilde{X}^{\leq t_{k-1}}, t_k-)$ is $\mathcal{A}$-measurable and define $Z = \1_B$.
Then
\begin{align*}
\E_{\mu_k}[Z]  &= (\P \times \tilde{\P}) \left[ (\tilde{X}^{\leq t_{k-1}}, t_k-) \in B \, \big| \,  t_k- \ne \infty \right] \\
&= (\P \times \tilde{\P}) \left[ (\tilde{X}^{\leq t_{k-1}}, t_k-) \in B ,  n \geq k \right]  \, c_0(k) \\
&= c_0(k) \, \E_{\P \times \tilde{\P}} \left[ \1_B(\tilde{X}^{\leq t_{k-1}}, t_k-) \1_{\{n \geq k\}} \right],
\end{align*}
where we used that $n \geq k$ is equivalent to $t_k \ne \infty$.
For a general $Z$ the claim now follows with ``measure theoretic induction'' \citep[Case 1--4 of Proof of Theorem 1.6.9]{Durrett:2010:PTE:1869916}, of which the first step was presented above.
\end{proof}
We use $\mu_k$ to define a pseudometric $d_k$ on the set of  c\`adl\`ag $\mathbb{A}$-adapted processes. In particular, for any two such processes $Z, \xi$, we define
\begin{equation}\label{equ:definition of d_k}
d_k(Z, \xi) := \E_{\mu_k} \left[ |Z - \xi|_2 \right].
\end{equation}
By looking at the equivalence relation induced by this pseudometric, $d_k$ defines a metric on the corresponding quotient space. Moreover, this equivalence relation defines indistinguishability between processes in our setting in the following sense.

\begin{definition}\label{def:indistinguishability}
We call two c\`adl\`ag $\mathbb{A}$-adapted processes $Z, \xi  : [0,T] \times (\Omega \times \tilde{\Omega}) \to \R^{r}$ indistinguishable, if $d_k(Z, \xi)=0$ holds for every  $1 \leq k \leq K$.
\end{definition}

\begin{rem}\label{rem:equivalent definition pseudo metric}
The (pseudo) metrics $d_k$ can also be defined directly as
\[ d_k (Z, \xi) = c_0(k)\,  \E_{\P \times \tilde{\P}}\left[ \1_{\{n \geq k\}} | Z_{t_k-} - \xi_{t_k-} | \right], \]
however, \eqref{equ:definition of d_k} allows the interpretation of $d_k$ as distance induced by an $L^1$-norm.
\end{rem}

\section{Extension of the neural jump ODE model}\label{sec:proposed method}
After briefly revisiting the original Neural Jump ODE model (Section~\ref{sec:Recall: Neural Jump ODE}) and the definition of the signature transform of paths (Section~\ref{sec:Signature Transform}), we introduce a signature-based extension of NJ-ODE, which we call the \emph{Path-Dependent Neural Jump ODE} (PD-NJ-ODE), in Section~\ref{sec:Signature NJ-ODE}.

\subsection{Recall: neural jump ODE}\label{sec:Recall: Neural Jump ODE}
We define $\mathcal{X} \subseteq \R^{d_X}$ and $\mathcal{H} \subseteq \R^{d_H}$ to be the observation and latent spaces for $d_X, d_H \in \N$. Moreover, we define three feedforward neural networks (with at least one hidden layer and, e.g., sigmoid activation functions),
\begin{itemize}
\item $f_{\theta_1}: \R^{d_H} \times \R^{d_X} \times [0,T] \times [0,T] \to \R^{d_H}$ modelling the ODE dynamics,
\item $\rho_{\theta_2}: \R^{d_X} \to \R^{d_H} $  modelling the jumps when new observations are made, and
\item $g_{\theta_3}: \R^{d_H} \to \R^{d_Y}$ the readout map, mapping into the target space $\mathcal{Y} \subseteq \R^{d_Y}$ for $d_Y \in \N$,
\end{itemize}
where $\theta := (\theta_1, \theta_2, \theta_3) \in \Theta$ are the trainable weights.
We define the pure jump stochastic process
\begin{equation}\label{equ:u}
u : \tilde\Omega \times [0,T] \to \R, (\tilde\omega, t) \mapsto u_t(\tilde\omega) := \sum_{i=1}^{n(\tilde\omega)} \1_{[t_i(\tilde\omega), \infty)}(t).
\end{equation}
Then the \emph{Neural Jump ODE} (NJ-ODE) is defined by the latent process $H := (H_t)_{t\in [0,T]}$ and the output process $Y := (Y_t)_{t\in [0,T]}$ which are the solution of the SDE system
\begin{equation}\label{equ:NJ-ODE}
\begin{split}
H_0 &= \rho_{\theta_2}\left(X_{0} \right), \\
dH_t &= f_{\theta_1}\left(H_{t-}, X_{\tau(t)},\tau(t), t - \tau(t) \right) dt + \left( \rho_{\theta_2}\left(X_{t} \right) - H_{t-} \right) du_t, \\
Y_t &= g_{\theta_3}(H_t).
\end{split}
\end{equation}
The latent process $H$ of the NJ-ODE before and after each observation can equivalently be written as
\begin{equation}\label{equ:NJ-ODE equ way writing}
\left\{
    \begin{array}{lll}
        h_{t_{i+1-}}  &:=& \operatorname{ODESolve}(f_{\theta_1}, (h_{t}, x_{t_i},t_i, t - t_i), (t_i, t_{i+1})) \\
       h_{t_{i+1}} &:= & \rho_{\theta_2}(x_{t_{i+1}})\,.
    \end{array}
\right.
\end{equation}

\subsection{Signature transform}\label{sec:Signature Transform}
The variation of a path is defined as follows.

\begin{definition}\label{def:p-variation}
    Let  $J$ be a closed interval in $\R$ and $d\geq 1$.
    Let $\mathbf{X}:J\rightarrow\R^{d}$ be a path on $J$.
    The variation of $\mathbf{X}$ on the interval $J$ is defined by
    \begin{equation*}
        \|\mathbf{X}\|_{var, J}
        =  \sup_{D}\sum_{t_j \in D} |\mathbf{X}_{t_{j}}-\mathbf{X}_{t_{j-1}} |_2 ,
    \end{equation*}
    where the supremum is taken over all finite partitions $D$ of $J$.
\end{definition}

\begin{definition}We denote the set of $\R^{d}$-valued paths of bounded variation  on $J$ by $BV(J, \R^d)$ and endow it with the norm
        \begin{align*}
\lVert \mathbf{X} \rVert_{BV} := |\mathbf{X}_0|_2 +  \|\mathbf{X}\|_{var, J}.
\end{align*}

\end{definition}\vspace*{-12pt}

For continuous paths of bounded variation we can define the signature transform.
\begin{definition}\label{def:signature}
    Let $J$ denote a closed interval in $\R$.
    Let $\mathbf{X}:J\rightarrow\R^{d}$ be a continuous path with finite variation.
    The signature of $\mathbf{X}$ is defined as
    \begin{equation*}
        S(\mathbf{X}) = \left(1, \mathbf{X}_J^1, \mathbf{X}_J^2, \dots\right),
    \end{equation*}
    where, for each $m\geq 1$,
    \begin{equation*}
        \mathbf{X}_J^m = \int_{\substack{u_1<\dots<u_m \\ u_1,\dots,u_m\in J}} d\mathbf{X}_{u_1}\otimes\dots\otimes d\mathbf{X}_{u_m} \in (\R^d)^{\otimes m}
    \end{equation*}
    is a collection of iterated integrals.
    The map from a path to its signature is called the signature transform.
\end{definition}

A good introduction to the signature transform with its properties and examples can be found in \citet{Chevyrev2016APO, KiralyOberhauser2019, fermanian2020embedding}.
In practice, we are not able to use the full (infinite) signature, but instead use a truncated version.

\begin{definition}    Let $J$ denote a compact interval in $\R$.
    Let $\mathbf{X}:J\rightarrow\R^{d}$ be a continuous path with finite variation.
    The  truncated signature of $\mathbf{X}$ of order $m$ is defined as
    \begin{equation*}
        \pi_m(\mathbf{X}) = \left(1,\mathbf{X}_J^1,\mathbf{X}_J^2,\dots,\mathbf{X}_J^m\right),
    \end{equation*}
    i.e., the first $m+1$ terms (levels) of the signature of $\mathbf{X}$.

\end{definition}

Note that the size of the truncated signature depends on the dimension of $\mathbf{X}$, as well as the chosen truncation level.
Specifically, for a path of dimension $d$, the dimension of the truncated signature of order $m$ is given by
\begin{equation}\label{eq:sig_nb_terms}
\begin{cases}
m+1, & \text{if } d =1, \\
 \frac{d^{m+1}-1}{d-1}, & \text{if } d >1.
\end{cases}
\end{equation}
Using the truncated signature as input to a model therefore entails a trade-off between accurately describing the path and limiting model complexity.

For the following universality result of the signature we have to exclude tree-like paths.
Essentially, a path is tree-like if it can be reduced to a constant path by successively removing pieces of the form $W\ast\overleftarrow{W}$, where $\overleftarrow{W}$ is the time-reversal of $W$ and $\ast$ the concatenation of paths.

\begin{definition}    A path $\mathbf{X}:[0,T]\rightarrow\R^{d}$ is tree-like if there exists a function $h:[0,T]\rightarrow[0,\infty)$ such that $h(0)=h(T)=0$ and such that, for all $s,t\in[0,T]$ with $s\leq t$,
    \begin{equation*}
       | \mathbf{X}_t-\mathbf{X}_s|_2 \leq h(s) + h(t) -2\inf_{u\in[s,t]}h(u).
      \end{equation*}
Two paths are called tree-like equivalent if following forward the first one and concatenating with the backwards running second one leads to a tree-like path.

\end{definition}

\begin{rem}
By adding time as one component to a given path, the set of tree-like equivalent paths reduces to a singleton.
\end{rem}

For a Hilbert space $\mathcal{H}$, we denote the set of continuous, $\mathcal{H}$-valued paths of bounded variation starting at the origin ($X_0 = 0$) on $[a,b]$ by $BV_0^c([a,b],\mathcal{H})$. The main reason for using the signature transform is its universal approximation property.
It states that any continuous function invariant under tree-like equivalences of a path can be approximated arbitrarily well by a linear function of the truncated signature transform for some truncation level and is proven, e.g., in \citet[Theorem 1]{KiralyOberhauser2019}.
\begin{theorem}\label{thm:universal_approx_sig}
    Let $\mathcal{P}$ be a compact subset of $BV_0^c([0,1],\mathcal{H})$ of paths that are not tree-like equivalent.
    Let $f:\mathcal{P}\rightarrow\R$ be continuous in variation norm.
    Then, for any $\varepsilon>0$, there exists $M>0$ and a linear functional $\mathbf{w}$ acting on the truncated signature of degree $M$ such that
    \[
        \sup_{\mathbf{x}\in\mathcal{P}}
        \left|f(\mathbf{x})- \left\langle \mathbf{w},\pi_M(\mathbf{x})\right\rangle\right|
        <\varepsilon.
    \]
\end{theorem}

In the following proposition we show that the result of Theorem~\ref{thm:universal_approx_sig} can be extended to functions with additional inputs.

\begin{prop}\label{prop:universal_approx_sig}
Let $\mathcal{P}$ be a compact subset of $BV_0^c ([0,1],\mathcal{H})$ of paths that are not tree-like equivalent and let $C \subseteq \R^m$ for $m \in \N$ be compact. We consider the Cartesian product  $BV_0^c ([0,1],\mathcal{H})  \times \R^m$ with the product norm given by the sum of the single norms (variation norm and $1$-norm).
    Let $f:\mathcal{P} \times C \rightarrow\R$ be continuous.
    Then, for any $\varepsilon>0$, there exists $M>0$ and a continuous function $\tilde f$ such that
    \[
        \sup_{(x,c) \in\mathcal{P} \times C}  \left|f(x, c) - \tilde f (\pi_M(x), c)  \right|  <\varepsilon.
    \]
  \end{prop}

  \begin{rem}
    We could with equal effort prove that there is a continuous selection of weights $ c \mapsto  \mathbf{w}(c) $ such that
    $ \left\langle \mathbf{w}(c),\pi_M(\mathbf{x})\right\rangle $ is close to  $ f(x,c) $ uniformly on compacts $\mathcal{P}$. For later purposes
    we shall need the proposition's assertion.
  \end{rem}

\begin{proof}
Since $f$ is a continuous function on a compact metric space, it is uniformly continuous by the Heine--Cantor theorem.
Hence, there exists $\delta > 0$ such that for all $c, \tilde c \in C$ with $|c - \tilde c| < \delta $ we have
\[| f(x, c) - f(x, \tilde c) | < \epsilon/2\]
for all $x \in \mathcal{P}$.
Since $C$ is compact, there exist finitely many open balls $(U_i)_{1 \leq i \leq N}$ with $U_i \subseteq \R^m$  of radius $\delta$ such that they cover $C$. Let the points $(c_i)_{1 \leq i \leq N} \in C^N$ be the centres of these balls.
By the partition of unity, there exist continuous functions $(\rho_i)_{1 \leq i \leq N}$, with $\rho_i : C \to [0,1]$ and $\operatorname{supp}(\rho_i) \subseteq U_i$ such that $\sum_{i=1}^N \rho_i(c) = 1$ for all $c \in C$.
For each $c_i$, Theorem~\ref{thm:universal_approx_sig} implies that there exist $M_i$ and $\mathbf{w}_i \in \R^{M_i}$ such that the function $x \mapsto f(x,c_i)$ is approximated well by $\tilde f_i (  \pi_{M_i}({x}) ) := \left\langle \mathbf{w}_i, \pi_{M_i}({x})\right\rangle$, i.e.,
\[
 \sup_{{x}\in\mathcal{P}}
        \left|f({x, c_i})-\left\langle \mathbf{w}_i, \pi_{M_i}({x})\right\rangle\right|
        <\varepsilon/2.
\]
W.l.o.g. we can assume that all $M_i = M$ are the same, by concatenating  $\mathbf{w}_i$ with $0$s.
Then the function $\tilde f (\pi_M(x), c) := \sum_{i=1}^N \rho_i(c) f_i ( \pi_{M}({x}) ) $ is continuous as a sum of products of continuous functions and satisfies the claim. Indeed, for any $(x,c) \in \mathcal{P} \times C$ we have
\begin{equation*}
\begin{split}
 \left|f({x, c})- \tilde f (\pi_M(x) ,c) \right|
  &=  \left| \sum_{i=1}^N \rho_i(c) \left[ f({x, c})- \tilde f_i (\pi_M(x)) \right] \right| \\
 & \leq \sum_{i=1}^N \rho_i(c) \left| f({x, c})- \tilde f_i (\pi_M(x)) \right| \\
 & \leq \sum_{i=1}^N \rho_i(c) \left( \left| f({x, c})- f({x, c_i}) \right| + \left| f({x, c_i})- \tilde f_i (\pi_M(x)) \right|  \right) \\
 & \leq \sum_{i=1}^N \rho_i(c) \left( \epsilon/2 + \epsilon/2 \right) \leq \epsilon,
\end{split}
\end{equation*}
where we used that $\rho_i(c) = 0$ if $c \notin U_i$ implying that $|c - c_i| < \delta$ if $\rho_i(c) > 0$ and therefore $| f(x, c) - f(x, c_i) | < \epsilon/2$.
\end{proof}

To apply this result, we need a tractable description of certain compact subsets of $BV_0^c ([0,1],\mathcal{\R}^d)$ that include suitable paths for our considerations. Since $BV_0^c ([0,1],\mathcal{\R}^d)$ is not finite dimensional, not every closed and bounded subset is compact. \citet[Theorem 2]{BUGAJEWSKI2020123752} characterizes relatively compact subsets of $BV_0^c ([0,1],\mathcal{\R})$, i.e., subsets such that their closure is compact. Moreover, they prove in \citet[Example 4]{BUGAJEWSKI2020123752} that the following set of functions is relatively compact.
\begin{prop}\label{prop:compact subset BV}
For every $N \in \N$ the family  $A_N \subseteq BV_0^c([0,1], \R)$ of all piecewise linear, bounded and continuous functions that can be written as
\begin{equation*}
f(t) = (a_1 t ) \1_{[t_0, t_1]}(t) + \sum_{i=2}^N  (a_i t + b_i ) \1_{(t_{i-1}, t_i]}(t),
\end{equation*}
is relatively compact, where $a_i, b_i \in [-N,N]$, $b_1=0$, $a_i t_i + b_i = a_{i+1} t_i + b_{i+1}$, for all $1 \leq i < N$, and $0=t_0 < t_1 < \dotsb < t_N = 1$.
\end{prop}
The following remark shows that this result can be extended to $\R^d$-valued paths, which we will need for our considerations.

\begin{rem}
\label{rem:compact subset BV}
Since the product of compact sets is compact and $BV_0^c ([0,1],\mathcal{\R}^d)$ can be identified with $BV_0^c ([0,1],\mathcal{\R})^d$, Proposition~\ref{prop:compact subset BV} can be extended to $\R^d$-valued paths. Moreover, the generalisation from $BV_0^c ([0,1],\mathcal{\R}^d)$ to $BV_0^c ([0,T],\mathcal{\R}^d)$ is immediate.
These are the compact subsets $A_N$ of $BV_0^c ([0,T],\mathcal{\R}^d)$ that we will use.
\end{rem}

\subsection{Path-dependent neural jump ODE}\label{sec:Signature NJ-ODE}
We adapt \eqref{equ:NJ-ODE} by using the signature as an additional input to both the neural ODE $f_{\theta_1}$ and the jump network $ \rho_{\theta_2}$. Moreover, since the process $X$ is no longer necessarily Markovian, we return to a recurrent structure for the jump network $ \rho_{\theta_2}$, as in \citet{Brouwer2019GRUODEBayesCM, ODERNN2019}.
Note that we cannot use the signature of the true path $(X_s)_{0 \leq s \leq t}$ of the data up to time $t$ as input, since we only have discrete observations of $X$ at the observation times $t_i$ (which is not sufficient to calculate the signature of $X$).
Instead, we use the shifted interpolation $\tilde X^{\leq t} - X_0 \in BV_0^c([0,T])$ up to time $t$ and compute the truncated signature $\pi_m(\tilde X^{\leq t} - X_0)$.
Together with the starting point $X_0$, this signature includes all available information (whereas the signature of $(X_s)_{0 \leq s \leq t}$ would include much more than the available information, i.e., it would not be $\mathcal{A}_t$-measurable). Moreover, the interpolation $\tilde X^{\leq t}$ has bounded variation, regardless of whether this is true for the original path $X$. Hence, Theorem~\ref{thm:universal_approx_sig} applies if $\tilde X^{\leq t}$ lies in some compact subset.
The advantage (besides Theorem~\ref{thm:universal_approx_sig}) of using the truncated signature rather than the discrete observations directly is that it accumulates information from an arbitrary number of observations in a vector of fixed size.
Additionally, we use the following random variables, which are running summary statistics of the observations, as inputs; $\tilde X^{\star}_t := \sup_{0\leq s \leq t} | \tilde X^{\leq t}_s|_1 \leq X^\star_T$, $n_t := \max\{ i \in \N \, | \, t_i \leq t \} \leq n$ and $\delta_t := \min\{ t_i - t_{i-1} | t_i \leq t \} \geq \delta$. They are useful, because with their help we can define necessary conditions implying that the interpolated observation path $\tilde X^{\leq t}$ lies within a compact subset of $BV_0^c([0,T])$.

Our proof requires the neural networks to be bounded so that we can derive an upper bound for the worst-case difference between the true conditional expectation and our model's approximation of it. Therefore, we introduce the following special class of neural networks with bounded outputs, based on any standard class of neural networks.

\begin{definition}[Bounded output neural networks]\label{def:bounded output NN}
For any dimension $d \in \N$ we define the bounded output activation function with trainable parameter $\gamma > 0$ as the Lipschitz  continuous function
\begin{equation*}
\Gamma_\gamma : \R^d \to \R^d, x \mapsto x \cdot \min\left(1, \frac{\gamma}{|x|_2}\right).
\end{equation*}
Then we define the class of \emph{bounded output neural networks} as
\begin{equation*}
\mathcal{N} := \{  f_{(\vartheta, \gamma)} := \Gamma_\gamma \circ \tilde f_{\tilde \theta} \, | \, \gamma > 0, \tilde f_{\tilde \theta} \in \tilde{\mathcal{N}}  \},
\end{equation*}
where $\tilde{\mathcal{N}}$ can be any set of neural networks. We use the notations $\tilde f_{\tilde \theta} \in \tilde{\mathcal{N}}$ and $ f_\theta \in \mathcal{N}$ for $\theta=(\tilde \theta, \gamma)$ to highlight the trainable weights $\tilde \theta$ (and $\gamma$) of the respective (bounded output) neural networks.
If needed, one could also consider appropriately smoothened versions of  $\Gamma_\gamma$.

\end{definition}

In the following we assume that $\tilde{\mathcal{N}}$ is a set of feedforward neural networks with Lipschitz continuous activation functions
such that, for any $d,D \in \N$ and any compact subset $\mathcal{X} \subset \R^d$, $\tilde{\mathcal{N}}$ is dense in the space of continuous functions $C(\mathcal{X}, \R^D)$ (with respect to the supremum norm). In particular, $\tilde{\mathcal{N}}$ must satisfy the standard universal approximation theorem, as is the case, e.g., for the set of single-hidden-layer neural networks with continuous, bounded, and non-constant activation functions \cite[Theorem~2]{hornik1991approximation}. Moreover, we assume that $\operatorname{id} \in \tilde{\mathcal{N}}$.

\begin{definition}\label{def:PD-NJ-ODE}
    The  \textit{Path-dependent neural jump ODE (PD-NJ-ODE)} model is given by
    \begin{equation}\label{equ:PD-NJ-ODE}
\begin{split}
H_0 &= \rho_{\theta_2}\left(0, 0, \pi_m (0), X_0 \right), \\
dH_t &= f_{\theta_1}\left(H_{t-}, t, \tau(t), \pi_m (\tilde X^{\leq \tau(t)} -X_0 ), X_0, \tilde X^{\star}_t, n_t,\delta_t \right) dt  \\
& \quad + \left( \rho_{\theta_2}\left( H_{t-}, t, \pi_m (\tilde X^{\leq \tau(t)}-X_0 ), X_0, \tilde X^{\star}_t, n_t,\delta_t \right) - H_{t-} \right) du_t, \\
Y_t &= \tilde g_{\tilde \theta_3}(H_t).
\end{split}
\end{equation}
    The functions $f_{\theta_1}, \rho_{\theta_2} \in \mathcal{N}$ are bounded output feedforward neural networks and $\tilde g_{\tilde \theta_3} \in \tilde{\mathcal{N}}$ is a feedforward neural network.  They have trainable parameters $\theta = (\theta_1, \theta_2, \tilde \theta_3) \in \Theta$, where $\theta_i = (\tilde \theta_i, \gamma_i)$ for $i \in \{1,2 \}$ and $\Theta$ is the set of all possible weights for the PD-NJ-ODE model; $m \in \N$ is the signature truncation level and $u$ is the jump process counting the observations defined in \eqref{equ:u}.
\end{definition}

\citet[Thm. 7, Chap. V]{Pro1992} implies that a unique solution of \eqref{equ:PD-NJ-ODE} exists. We write $Y^\theta(\tilde{X}^{\leq \tau(\cdot)})$ to emphasize the dependence of $Y$ on $\theta$ and $\tilde{X}^{\leq \tau(\cdot)}$.
Moreover, we present an implementable version of this model in Algorithm~\ref{algo:1}.
		\begin{figure*}[t]
		\begin{algorithm}[H]
   \caption{The path-dependent NJ-ODE.
	A small step size $\Delta t$ is fixed and we denote $t_{n+1} := T$. 
    $\hbox{ODESolve}(f, x, (a,b))$ numerically solves the first-order ODE defined by $f$, taking the inputs $x$, on the interval $(a,b)$.
	}
   \label{algo:1}
\begin{algorithmic}
   \STATE {\bfseries Input:} Data points {with} timestamps and masks $\{(X_i, t_i, M_i)\}_{i=0\dots {n}}$, 
   \STATE set $H_{0-} = 0$
   \FOR{$i=0$ {\bfseries to} {$n$}} 
        \STATE construct $\tilde X^{\leq t_i}$ from data
        \STATE $S_i = \pi_m(\tilde X^{\leq t_i} - X_0)$  \hfill $\triangleright$ compute truncated signature
		\STATE ${H_{t_i}} = {\rho_{\theta_2}(H_{t_i-}, t_{i}, S_i, X_0, \tilde X^{\star}_t, n_t,\delta_t})$ \hfill $\triangleright$ Update hidden state given next observation $x_i$
		\STATE {$Y_{t_i} = \tilde g_{\tilde{\theta}_3}(H_{t_i})$}  \hfill $\triangleright$ compute output
		\STATE $s \leftarrow t_i$
		
		\WHILE {$s + \Delta t < t_{i+1}$}
        		\STATE ${H_{s + \Delta t}} = \hbox{ODESolve}(f_{\theta_1},  (H_{s}, s, t_i, S_i, X_0, \tilde X^{\star}_t, n_t,\delta_t), (s, s + \Delta t))$ \hfill $\triangleright$ get next hidden state
        		\STATE {$Y_{s + \Delta t} = \tilde g_{\tilde{\theta}_3}(H_{s + \Delta t})$}  \hfill $\triangleright$ compute output
        		\STATE $s \leftarrow s + \Delta t$
        \ENDWHILE
        \STATE ${H_{t_{i+1}-}} = \hbox{ODESolve}(f_{\theta_1},  (H_{s-}, s, t_i, S_i, X_0, \tilde X^{\star}_t, n_t,\delta_t), (s, t_{i+1}))$
        \STATE {$Y_{t_{i+1}-} = \tilde g_{\tilde{\theta}_3}(H_{(s + \Delta t)-})$}
   \ENDFOR
\end{algorithmic}
\end{algorithm}
\vspace{-0.5cm}
\end{figure*}

\begin{rem}[Continuation of Remark {\ref{rem:extension assumptions M0}}]\label{rem:extension PD-NJ-ODE for other M0}
In the case that $X_0$ is not observed completely, we have to slightly extend the model architecture \eqref{equ:PD-NJ-ODE} by another bounded output neural network $\zeta_{\theta_4} \in \mathcal{N}$, where we distinguish between the two cases.
\begin{enumerate}
\item If $M_0 = 0$, then $\zeta_{\theta_4}: \{ 0 \} \to \R^{d_H}$ and $H_0 = \zeta_{\theta_4}(0)$.
\item If $M_{0, j} = \1_{I_0}(j)$, then $\zeta_{\theta_4}: \R^{|I_0|} \to \R^{d_H}$ and $H_0 = \zeta_{\theta_4}( \operatorname{proj}_{I_0} (X_0) )$, where $\operatorname{proj}_{I_0} : \R^{d_X} \to \R^{|I_0|}$ is the projection to the coordinates in $I_0$.
\end{enumerate}
\end{rem}

\subsubsection{Objective function}\label{sec:Objective Function}
Let $\mathbb{D}$  be the set of all c\`adl\`ag $\R^{d_X}$-valued $\mathbb{A}$-adapted processes on the probability space $(\Omega \times \tilde\Omega , \F \otimes \tilde\F, \mathbb{F} \otimes \tilde{ \mathbb{F}}, \P \times \tilde\P)$. Then we define our objective functions
\begin{align}
\Psi: \, &\mathbb{D} \to \R, \nonumber \\
&Z \mapsto \Psi(Z) := \E_{\P\times\tilde\P}\left[ \frac{1}{n} \sum_{i=1}^n  \left(  \left\lvert M_i \odot ( X_{t_i} - Z_{t_i} ) \right\rvert_2 + \left\lvert M_i \odot (Z_{t_i} - Z_{t_{i}-} ) \right\rvert_2 \right)^2 \right], \label{equ:Psi} \\
\Phi : \, &\Theta \to \R, \theta \mapsto \Phi(\theta) := \Psi(Y^{\theta}(\tilde{X}^{\leq \tau(\cdot)})), \label{equ:Phi}
\end{align}
where $\odot$ is the element-wise multiplication (Hadamard product) and $\Phi$ will be our (theoretical) loss function.
Note that the definition of $Y^\theta$ directly implies that it is an element of $\mathbb{D}$; hence, $\Phi$ is well defined.

Let us assume that we observe $N \in \N$ independent realizations of the path $X$  together with independent realizations of the observation mask $M$ at times $( t_1^{(j)}, \dotsc,  t_{n^{(j)}}^{(j)})$, $1 \leq j \leq N$, which are themselves independent realizations of the random vector $(n, t_1, \dotsc, t_n)$. In particular, let us assume that $X^{(j)} \sim X$, $M^{(j)} \sim M$ and $(n^{(j)}, t_1^{(j)}, \dotsc, t_{n^{(j)}}^{(j)}) \sim ( n, t_1, \dotsc, t_{n})$ are i.i.d. random processes (respectively variables) for $1 \leq j \leq N$ and that our training data is one realization of them.
We write $Y^{\theta, j} := Y^{\theta }(\tilde{X}^{\leq \tau(\cdot), (j)})$. Then the Monte Carlo approximation of our loss function
\begin{equation}\label{equ:appr loss function}
\hat\Phi_N(\theta) := \frac{1}{N} \sum_{j=1}^N  \frac{1}{n^{(j)}}\sum_{i=1}^{n^{(j)}} \left(  \left\lvert M_{i}^{(j)} \odot \left( X_{t_i^{(j)}}^{(j)} - Y_{t_i^{(j)}}^{\theta, j } \right) \right\rvert_2 + \left\lvert M_{i}^{(j)} \odot \left( Y_{t_i^{(j)}}^{\theta, j } - Y_{t_{i}^{(j)}-}^{\theta, j } \right) \right\rvert_2 \right)^2,
\end{equation}
converges $(\P\times\tilde\P)$-a.s. to $\Phi(\theta)$ as $N \to \infty$, by the law of large numbers (cf. Theorem~\ref{thm:MC convergence Yt}).

\section{Convergence guarantees}\label{sec:Convergence Guarantees}
As in \citet{herrera2021neural}, we first give the convergence result for the objective function $\Phi$ and then show that its Monte Carlo approximation $\hat \Phi$ converges to it. The proofs mainly follow the proofs therein, with extensions for the more general setting. Let $\hat{\Theta}_m \subset \Theta$ be the set of possible weights $\theta = (\theta_1, \theta_2, \tilde \theta_3) \in \Theta$ with $\theta_i = (\tilde{\theta}_i, \gamma_i)$ for $i \in \{ 1,2\}$, for the three (bounded output) neural networks of the PD-NJ-ODE \eqref{equ:NJ-ODE}, such that their widths and depths\footnote{The width of a neural network is the maximal number of nodes in any of its hidden layers and the depth is the number of hidden layers.} are at most $m$  and such that the truncated signature of level $m$ or smaller is used as input.
We define the compact subset $\Theta_m := \{ \theta = ((\tilde{\theta}_1, \gamma_1), (\tilde{\theta}_2, \gamma_2), \tilde{\theta}_3 ) \in \hat{\Theta}_m \, | \, |\tilde \theta_i|_2 \leq m, \gamma_i \leq m \} \subset \hat{\Theta}_m$.
Furthermore, we use the notation $\hat{\Theta}_m^i$,  $\Theta_m^i$ and  $\tilde{\Theta}_m^i$  if we speak of the projections of the sets on the weights $\theta_i$ and $\tilde{\theta}_i$ respectively.

\subsection{Convergence of theoretical loss function}\label{sec:Convergence of Theoretical loss function}
\begin{theorem}\label{thm:1}
Let $\theta^{\min}_m \in \Theta_m^{\min} := \argmin _{\theta \in \Theta_m}\{ \Phi(\theta) \}$ for every $m \in \N$. If Assumptions~\ref{assumption:1} to \ref{assumption:6} are satisfied, then, for $m \to \infty$, the value of the loss function $\Phi$ \eqref{equ:Phi} converges to the minimal value of $\Psi$ \eqref{equ:Psi} which is uniquely achieved by $\hat{X}$ up to indistinguishability (cf. Definition~\ref{def:indistinguishability}), i.e.,
\begin{equation*}
\Phi(\theta_m^{\min}) \xrightarrow{m \to \infty} \min_{Z \in \mathbb{D}} \Psi(Z) = \Psi(\hat{X}).
\end{equation*}
Furthermore, for every $1 \leq k \leq K$ we have that $Y^{\theta_m^{\min}}$ converges to $\hat{X}$ in the metric $d_k$ \eqref{equ:definition of d_k} as $m \to \infty$.
\end{theorem}

Before proving the theorem, we derive the following useful result, which is an extension of a result in \citet{herrera2021neural}.

\begin{lemma}\label{lem:L2 identity}
For any $\mathbb{A}$-adapted process $Z$ it holds that
\begin{align*}
&\E_{\P \times\tilde\P}\left[\tfrac{1}{n} \sum_{i=1}^n \left\lvert M_{t_i} \odot ( X_{t_i} - Z_{t_i-} ) \right\rvert_2^2\right] \\&
	= \E_{\P \times\tilde\P}\left[ \tfrac{1}{n}\sum_{i=1}^n \left\lvert M_{t_i} \odot ( X_{t_i} - \hat{X}_{t_i-} ) \right\rvert_2^2\right] + \E_{\P \times\tilde\P}\left[\tfrac{1}{n}\sum_{i=1}^n \left\lvert M_{t_i} \odot (  \hat{X}_{t_i-} - Z_{t_i-}) \right\rvert_2^2\right] .
\end{align*}
\end{lemma}

\begin{proof}
First note that by Assumption~\ref{assumption:3} we have that $X_{t_i} = X_{t_i-}$ almost surely.
Then
\begin{align*}
&\E_{\P \times\tilde\P}\left[\tfrac{1}{n}\sum_{i=1}^n \left\lvert M_{t_i} \odot ( X_{t_i-} - Z_{t_i-} ) \right\rvert_2^2\right]
	=  \E_{\tilde\P}\left[\tfrac{1}{n}\sum_{i=1}^n \sum_{j=1}^{d_X} M_{t_i, j} \E_{\P}\left[ \left\lvert   X_{t_i-,j} - Z_{t_i-,j}  \right\rvert^2\right]\right] \\&
	= \E_{\tilde\P}\left[\tfrac{1}{n}\sum_{i=1}^n \sum_{j=1}^{d_X} M_{t_i, j} \left( \E_{\P}\left[ \left\lvert X_{t_i-,j} - \hat{X}_{t_i-,j} \right\rvert^2\right]  + \E_{\P}\left[ \left\lvert  \hat{X}_{t_i-,j} - Z_{t_i-,j}  \right\rvert^2 \right] \right) \right] \\&
	 = \E_{\P \times\tilde\P}\left[\tfrac{1}{n}\sum_{i=1}^n \left\lvert M_{t_i} \odot ( X_{t_i-} - \hat{X}_{t_i-} ) \right\rvert_2^2\right] + \E_{\P \times\tilde\P}\left[\tfrac{1}{n}\sum_{i=1}^n \left\lvert M_{t_i} \odot (  \hat{X}_{t_i-} - Z_{t_i-} ) \right\rvert_2^2\right],
\end{align*}
where we used \citet[Proposition B.2, Lemma B.3]{herrera2021neural} and Fubini's theorem.
\end{proof}

Moreover, we will make use of the following construction.

\begin{lemma}\label{lem:expectation weighted sum over t_i terms}
    Let $U \sim \operatorname{Unif}([0,1])$ and denote by $\mu_U$ the corresponding probability measure. Let $\varphi : [0,T] \times (\Omega\times\tilde\Omega) \to \R$ be a measurable function such that, for every $t\in [0,T]$, $\varphi (t)$ is $\mathcal{F} \otimes \tilde{\mathcal{F}}$-measurable (note that this condition is satisfied by all adapted processes).
    If $U$ is independent of $\mathcal{F} \otimes \tilde{\mathcal{F}}$, i.e., if we consider the product measure $\mu_U \times (\P\times \tilde{\P})$, then we have for $\bar t := \sum_{i=1}^n \1_{(\frac{i-1}{n}, \frac{i}{n}]}(U) \, t_i$ that
    \begin{equation*}
        \E_{ \mu_U \times (\P\times \tilde\P)} [\varphi\left(\bar t\right)] = \E_{\P\times \tilde\P}\left[\sum_{i=1}^n \frac{1}{n} \varphi(t_i)\right].
    \end{equation*}
\end{lemma}

\begin{proof}
    We have
    \begin{align*}
        \E_{ \mu_U \times (\P\times \tilde\P)} \left[\varphi\left(\bar t\right)\right] &=
        \E_{ \mu_U \times (\P\times \tilde\P)} \left[\varphi\left( \sum_{i=1}^n \1_{(\frac{i-1}{n}, \frac{i}{n}]}(U) \, t_i \right)\right] \\&
        = \E_{ \P\times \tilde\P } \left[ \int_0^1 \varphi\left( \sum_{i=1}^n \1_{(\frac{i-1}{n}, \frac{i}{n}]} (u) \, t_i \right) \, du\right]
        = \E_{ \P\times \tilde\P } \left[ \sum_{i=1}^n \int_{\frac{i-1}{n}}^{\frac{i}{n}} \varphi\left( t_i \right) \, du \right]
        = \E_{\P\times \tilde\P}\left[\sum_{i=1}^n \frac{1}{n} \varphi(t_i)\right],
    \end{align*}
    by Fubini's theorem.
\end{proof}

\begin{cor}\label{cor:expectation weighted sum over t_i terms}
    Similarly, we have for $\bar t := \sum_{i=1}^n \1_{(\frac{i-1}{2n}, \frac{i}{2n}]}(U) \, t_i + \sum_{i=1}^n \1_{(\frac{n+i-1}{2n}, \frac{n+i}{2n}]}(U) \, t_{i-1}$,
    \begin{equation*}
        \E_{ \mu_U \times (\P\times \tilde\P)} [\varphi\left(\bar t\right)] = \frac{1}{2}\left( \E_{\P\times \tilde\P}\left[\sum_{i=1}^n \frac{1}{n} \varphi(t_i)\right] + \E_{\P\times \tilde\P}\left[\sum_{i=1}^n \frac{1}{n} \varphi(t_{i-1})\right] \right).
    \end{equation*}
\end{cor}

\begin{proof}[Proof of Theorem {\ref{thm:1}}]
We start by showing that $\hat{X} \in \mathbb{D}$ is the unique minimizer of $\Psi$ up to indistinguishability (as defined in Definition~\ref{def:indistinguishability}). First, note  that for every $t_i$ we have $M_{t_i} \odot \hat{X}_{t_i} = M_{t_i} \odot X_{t_i}$ and that $X_{t_i} = X_{t_i-}$ if $t_i \notin \mathcal{J}$, hence with probability 1. Therefore,
\begin{equation*}
\begin{split}
\Psi(\hat{X})
	&= \E_{\P\times\tilde\P}\left[\frac{1}{n}\sum_{i=1}^n \left\lvert M_{t_i} \odot (X_{t_i} - \hat{X}_{t_i-} ) \right\rvert_2^2\right] \\
	&= \min_{Z \in \mathbb{D}} \E_{\P\times\tilde\P}\left[\frac{1}{n}\sum_{i=1}^n \left\lvert M_{t_i} \odot (X_{t_i } - Z_{t_i-}) \right \rvert_2^2\right] \\
	&\leq \min_{Z \in \mathbb{D}} \E_{\P\times\tilde\P}\left[\frac{1}{n}\sum_{i=1}^n \left( \left\lvert M_{t_i} \odot (X_{t_i} - Z_{t_i} ) \right \rvert_2 + \left\lvert M_{t_i} \odot (Z_{t_i} - Z_{t_i-}) \right \rvert_2 \right)^2\right] \\
	&= \min_{Z \in \mathbb{D}}\Psi(Z),
\end{split}
\end{equation*}
where we use Lemma~\ref{lem:L2 identity}  for the second line and the triangle inequality for the third line. Hence, $\hat{X}$ is a minimizer of $\Psi$.

Before we can show uniqueness of $\hat{X}$, we need some additional results. For those, let $Z \in \mathbb{D}$.
Let $c_1 := \E_{\P \times \tilde\P}\left[ n \right]^{1/2} \in (0, \infty)$, then  the H\"older inequality, together with the fact that $n \geq 1$, yields
\begin{equation}\label{equ:HI}
\E_{\P \times \tilde\P}\left[ \left\lvert Z \right\rvert_2 \right]
	= \E_{\P \times \tilde\P}\left[ \frac{\sqrt{n}}{\sqrt{n}} \left\lvert Z \right\rvert_2 \right]
	\leq c_1 \, \E_{\P \times \tilde\P}\left[ \frac{1}{n} \left\lvert Z \right\rvert_2^2 \right]^{1/2}.
\end{equation}
By Assumption~\ref{assumption:1} we know that $c_2 := \min_{1 \leq j \leq d_X} \tilde\P (M_{k,j} = 1) > 0$. Hence, we have for any $1 \leq k \leq K$ by the independence of $M_{k,j}$ from $t_k$, $n$ and $\mathcal{A}_{t_k-}$ that
\begin{align*}
\E_{\P \times \tilde\P}\left[ \1_{\{n \geq k\}} \left\lvert M_{t_k} \odot ( \hat{X}_{t_k-} - Z_{t_k-} ) \right\rvert_1 \right]
	&= \E_{\P \times \tilde\P}\left[ \1_{\{n \geq k\}} \sum_{j=1}^{d_X} M_{k,j} \left\lvert  \hat{X}_{t_k-,j} - Z_{t_k-,j}  \right\rvert \right] \\&
	=\sum_{j=1}^{d_X} \E_{\P \times \tilde\P}\left[ M_{k,j} \right] \, \E_{\P \times \tilde\P}\left[ \1_{\{n \geq k\}}  \left\lvert  \hat{X}_{t_k-,j} - Z_{t_k-,j}  \right\rvert \right]
	\geq c_2 \, \E_{\P \times \tilde\P}\left[ \1_{\{n \geq k\}} \left\lvert \hat{X}_{t_k-} - Z_{t_k-}  \right\rvert_1 \right] ,
\end{align*}
and by the equivalence of $1$- and $2$-norm, we therefore have for some constant $c_3 > 0$
\begin{equation}\label{equ:M split}
 \E_{\P \times \tilde\P}\left[ \1_{\{n \geq k\}} \left\lvert \hat{X}_{t_k-} - Z_{t_k-}  \right\rvert_2 \right] \leq \frac{c_3}{c_2} \E_{\P \times \tilde\P}\left[ \1_{\{n \geq k\}} \left\lvert M_{t_k} \odot ( \hat{X}_{t_k-} - Z_{t_k-} ) \right\rvert_2 \right].
\end{equation}

To see that $\hat{X}$ is the unique minimiser up to indistinguishability, let $Z \in \mathbb{D}$ be a process which is not indistinguishable from $\hat{X}$. Hence, there exists some $1 \leq k \leq K$ such that $d_k(\hat{X}, Z) > 0$.
We have
\begin{equation*}
\begin{split}
\Psi(Z) &= \E_{\P\times\tilde\P}\left[\frac{1}{n}\sum_{i=1}^n \left( \left\lvert M_{t_i} \odot (X_{t_i} - Z_{t_i} ) \right \rvert_2 + \left\lvert M_{t_i} \odot ( Z_{t_i} - Z_{t_i-} ) \right \rvert_2 \right)^2\right]\\
	& \geq \E_{\tilde\P}\left[\frac{1}{n}\sum_{i=1}^n \E_{\P}\left[\left\lvert M_{t_i} \odot ( X_{t_i} - Z_{t_i-} ) \right\rvert_2^2\right] \right] \\
	&=  \E_{\P \times\tilde\P}\left[ \tfrac{1}{n}\sum_{i=1}^n \left\lvert M_{t_i} \odot ( X_{t_i} - \hat{X}_{t_i-} ) \right\rvert_2^2\right] + \E_{\P \times\tilde\P}\left[\tfrac{1}{n}\sum_{i=1}^n \left\lvert M_{t_i} \odot (  \hat{X}_{t_i-} - Z_{t_i-}) \right\rvert_2^2\right] \\
	&= \Psi(\hat{X})  + \E_{\P \times\tilde\P}\left[\tfrac{1}{n}\sum_{i=1}^n \left\lvert M_{t_i} \odot (  \hat{X}_{t_i-} - Z_{t_i-}) \right\rvert_2^2\right] ,
\end{split}
\end{equation*}
where we used triangle-inequality for the second and Lemma~\ref{lem:L2 identity} in the third line. Hence, it is enough to note that the second term is greater than $0$. Indeed,
 \begin{equation}\label{equ:thm1-positive expectation}
\begin{split}
 \E_{\P \times\tilde\P}\left[\tfrac{1}{n}\sum_{i=1}^n \left\lvert M_{t_i} \odot (  \hat{X}_{t_i-} - Z_{t_i-}) \right\rvert_2^2\right]
	&=  \E_{\P \times\tilde\P}\left[\tfrac{1}{n}\sum_{i=1}^K \1_{\{ n \geq i \}} \left\lvert M_{t_i} \odot (  \hat{X}_{t_i-} - Z_{t_i-}) \right\rvert_2^2\right] \\
	& \geq \E_{\P \times\tilde\P}\left[\tfrac{1}{n}  \1_{\{ n \geq k \}} \left\lvert M_{t_k} \odot (  \hat{X}_{t_k-} - Z_{t_k-}) \right\rvert_2^2\right] \\
	& \geq c_1^{-2} \E_{\P \times\tilde\P}\left[ \1_{\{ n \geq k \}} \left\lvert M_{t_k} \odot (  \hat{X}_{t_k-} - Z_{t_k-}) \right\rvert_2\right]^2 \\
	& \geq \left( \frac{c_2}{c_1 c_3} \right)^2 \E_{\P \times\tilde\P}\left[ \1_{\{ n \geq k \}} \left\lvert  \hat{X}_{t_k-} - Z_{t_k-}  \right\rvert_2\right]^2 \\
	& = \left( \frac{c_2}{c_0 c_1 c_3} \right)^2 d_k(\hat{X}, Z)^2 > 0,
\end{split}
\end{equation}
where we used \eqref{equ:HI} for the 3rd, \eqref{equ:M split} for the 4th and Proposition~\ref{prop:mu_k expectation equiv} together with \eqref{equ:definition of d_k} for the last line. Hence, $\Psi(Z) > \Psi(\hat{X})$.

Next we show that \eqref{equ:PD-NJ-ODE} can approximate $\hat{X}$ arbitrarily well. Since the dimension $d_H$ can be chosen freely, let us fix it to $d_H := d_X$. Furthermore, let us fix  $\tilde \theta_3^{\star}$ such that $ \tilde g_{\theta_3^{\star}} = \id$, which is possible since we assumed that $\id \in \tilde{\mathcal{ N}}$.
Let $\varepsilon > 0$,   $N_\varepsilon := \lceil 2 (T+1) \varepsilon^{-2} \rceil  $ (implying that $\lim_{\varepsilon \to 0} N_\varepsilon = \infty$) and $\mathcal{P}_\varepsilon$ be the closure of the set $A_{ N_\varepsilon  }$ of Remark~\ref{rem:compact subset BV}, which is compact.
For any $1 \leq j \leq d_X$, the function $f_j$ is continuous by Assumption~\ref{assumption:4} and can (by abuse of notation) equivalently be written as (continuous) function $ f_j(t, \tau(t), \tilde X^{\leq \tau(t) } -X_0, X_0 )$.
Therefore, Proposition~\ref{prop:universal_approx_sig} implies that there exists an $m_0 = m_0(\varepsilon) \in \N$ and a continuous function $\hat f_j$ such that
\begin{equation*}
\sup_{(t, \tau, X) \in [0,T]^2\times \mathcal{P}_\varepsilon } \left| f_j(t, \tau, X ) - \hat f_j(t, \tau, \pi_{m_0}( X -X_0 ), X_0 )\right| \leq \varepsilon/2.
\end{equation*}
Since the variation of functions in $\mathcal{P}_\varepsilon$ is uniformly bounded by a  finite constant,  the set of their truncated signatures $\pi_{m_0}(\mathcal{P}_\varepsilon)$ is a bounded subset in $\R^d$ for some $d \in \N$ (depending on $d_X$ and $m_0$), hence its closure, denoted by $\Pi_\varepsilon$, is compact.
Therefore, the universal approximation theorem for neural networks  \citep[Theorem~2]{hornik1991approximation} implies that there exists an $m_{1,1} = m_{1,1}(\varepsilon) \in \N$ and neural network weights $\tilde{\theta}_{1,1}^{\star, m_{1,1}} \in \tilde{\Theta}_{m_{1,1}}^{1}$ such that for every $1 \leq j \leq d_X$ the function $\hat f_j$ is approximated up to $\varepsilon/2$ by the $j$th coordinate of the neural network $\tilde f_{\tilde{\theta}_{1,1}^{\star, m_{1,1}}} \in \tilde{\mathcal{N}}$ (denoted by $\tilde f_{\tilde{\theta}_{1,1}^{\star, m_{1,1}}, j}$) on the compact set $[0,T]^2\times \Pi_\varepsilon$. Hence, combining the two approximations we get (by triangle inequality)
\begin{equation*}
\sup_{(t, \tau, X) \in [0,T]^2\times \mathcal{P}_\varepsilon } \left| f_j(t, \tau, X ) -  \tilde f_{\tilde \theta_{1,1}^{\star, m_{1,1}}, j}  (t, \tau, \pi_{m_0}( X -X_0 ), X_0 )\right| \leq \varepsilon.
\end{equation*}
Similarly we get that  there exists an $m_2 = m_{2,1}(\varepsilon) \in \N$ and neural network weights $\tilde \theta_{2,1}^{\star, m_{2,1}} \in \tilde \Theta_{m_{2,1}}^2$ such that for every $1 \leq j \leq d_X$
\begin{equation*}
\sup_{(t, X) \in [0,T] \times \mathcal{P}_\varepsilon } \left| F_j(t, t, X ) -  \tilde \rho_{\tilde \theta_{2,1}^{\star, m_{2,1}}, j}  (t, \pi_{m_0}( X -X_0 ), X_0 )\right| \leq \varepsilon.
\end{equation*}
Extending the inputs of the neural networks does not worsen the approximation, since the corresponding weights can simply be set to $0$; hence, $H_{t-}$ can also be used as an additional input. To simplify the derivation, we will not explicitly write $H_{t-}$ as an input, but keep in mind that this argument permits its inclusion.

Next we define the bounded output neural networks based on these neural networks. For this let us define
\begin{equation*}
\gamma_1 := \max_{(t, \tau, X) \in [0,T]^2\times \mathcal{P}_\varepsilon } \left|  \tilde f_{\tilde \theta_{1,1}^{\star, m_{1,1}}}  (t, \tau, \pi_{m_0}( X -X_0 ), X_0 )\right|
\end{equation*}
and $\gamma_2$ equivalently for $\tilde \rho_{\tilde \theta_{2,1}^{\star, m_{2,1}}}$. Since the neural networks are continuous functions, they attain a finite maximum on the compact sets; hence, $\gamma_1$ and $\gamma_2$ are finite.
Then we define the bounded output neural networks  $\bar f_{\bar \theta_{1,1}^{\star, m_{1,1}}}, \bar \rho_{ \bar \theta_{2,1}^{\star, m_{2,1}}} \in \mathcal{N}$ with $\bar \theta_{i,1}^{\star, m_{i,1}} := (\tilde \theta_{i,1}^{\star, m_{i,1}}, \gamma_i)$.
Clearly, these bounded output neural networks coincide with the neural networks on the compact sets. Therefore, they satisfy the same $\varepsilon$-approximation.

It is important to note, however, that this boundedness is insufficient to show convergence, since the bounds are not implied by boundedness of $f_j$ and $F_j$ and depend on $\epsilon$. In particular, as $\epsilon$ decreases and $m$ therefore grows, $\gamma_i$ may grow so rapidly that the expectation of the neural networks outside the compact set does not converge to $0$ as the compact set expands.
Therefore, we compensate the neural networks outside the compact set with another neural network. To be able to tell whether the input path is in the relevant compact set, we use the random variables $\tilde X^{\star}_t \leq X^\star_T$, $n_t \leq n$ and $\delta_t  \geq \delta$ defined in Section~\ref{sec:Signature NJ-ODE} as additional inputs for the neural networks. By setting the corresponding weights to $0$ in the neural networks defined before, the $\epsilon$-approximation still holds.
Importantly, if $\tilde X^{\star}_t \leq 1/\epsilon$, $n_t \leq 1/\epsilon$ and $\delta_t \geq \epsilon$ then $\tilde X^{\leq t} - X_0 \in A_{N_\epsilon} \subset \mathcal{P}_\epsilon$ hence the $\epsilon$-approximation holds. Otherwise, we have $X^\star_T > 1/\epsilon$ or $n > 1/\epsilon$ or $\delta < \epsilon$.
Let $\tilde \mu_1$ be the push-forward measure of $dt \times (\P \times \tilde{\P})$ through the measurable map defining the inputs to our neural networks
\begin{equation}\label{equ:thm1 input to NN 1}
\begin{split}
    [0,T] \times (\Omega \times \tilde \Omega)  & \to \R^{2+d+d_X+3},\\
    (t, (\omega, \tilde \omega)) &\mapsto \left(t, \tau(t), \pi_{m_0}((\tilde{X}^{\leq t} - X_0)(\omega, \tilde \omega)), (X_0, \tilde X^{\star}_t, n_t, \delta_t)(\omega, \tilde \omega) \right)
\end{split}
\end{equation}
and define $D_1 := [0,T]^2\times \R^d \times \R^{d_X}\times [0, 1/\epsilon]^2 \times [\epsilon, T]$.
If we denote the complement as $D_1^\complement$ then we know by the argument above that on $D_1$ the $\epsilon$-approximation holds for any input of the form \eqref{equ:thm1 input to NN 1} and that
\begin{equation}
    \begin{split}
        \1_{D_1^\complement} \leq \left( \1_{\{ X_T^\star \geq 1/\varepsilon \}} + \1_{\{ n \geq  1/\varepsilon \}} + \1_{\{ \delta \leq \epsilon \}} \right) =: \zeta.
    \end{split}
\end{equation}
If we define
\begin{equation*}
    \upsilon := \bar f_{\theta_{1,1}^{\star, m_{1,1}}} \1_{ D_1^\complement } : \R^{2+d+d_X} \to \R^{d_X},
\end{equation*}
then $\tilde \mu_1 (\R^{2+d+d_X+3}) = T$, i.e., $\tilde\mu_1$ is a finite measure and $\upsilon$ is a function in $L^2(\tilde\mu_1)$, since it is bounded by $\gamma_1$.
Hence, \citet[Theorem~1]{hornik1991approximation} implies that there exists $m_{1,2} =  m_{1,2}(\epsilon) \in \N$ and neural network weights $\tilde \theta^{\star, m_{1,2}}_{1,2} \in \tilde{\Theta}_{m_{1,2}}^{1}$ such that the corresponding neural network $\tilde f_{\tilde \theta^{\star, m_{1,2}}_{1,2}}$ satisfies $\int |\upsilon - \tilde f_{\tilde \theta^{\star, m_{1,2}}_{1,2}} |_2^2 d\tilde\mu_1 < \epsilon$ (note that w.l.o.g.\ we can assume that $\tilde f_{\tilde \theta^{\star, m_{1,2}}_{1,2}} \in \mathcal{N}$, i.e., that it is a bounded output neural network). Then we define the neural network $f_{ \theta^{\star, m_{1}}_{1}} = \bar f_{\bar \theta_{1,1}^{\star, m_{1,1}}} - \tilde f_{\tilde \theta^{\star, m_{1,2}}_{1,2}} \in \mathcal{N}$, with the weights $\theta^{\star, m_{1}}_{1} = (\bar \theta_{1,1}^{\star, m_{1,1}}, \tilde \theta^{\star, m_{1,2}}_{1,2})$ and $m_1 = m_{1,1} + m_{1,2}$.\\
Similarly (but not identically, since the push-forward map differs), we proceed for $F_j$. Let $\tilde \mu_2$ be the push-forward measure of $du \times (\P \times \tilde{\P})$ through the measurable map
\begin{equation}\label{equ:thm1 input to NN 2}
\begin{split}
    [0,1] \times (\Omega \times \tilde \Omega)  & \to \R^{1+d+d_X+3},\\
    (U, (\omega, \tilde \omega)) &\mapsto \left(\bar t, \pi_{m_0}((\tilde{X}^{\leq \bar t} - X_0)(\omega, \tilde \omega)), (X_0,\tilde X^{\star}_t, n_t, \delta_t)(\omega, \tilde \omega) \right),
\end{split}
\end{equation}
where $\bar t := \bar t (U, (\omega, \tilde \omega)) := \sum_{i=1}^n \1_{(\frac{i-1}{2n}, \frac{i}{2n}]}(U) \, t_i(\tilde \omega) + \sum_{i=1}^n \1_{(\frac{n+i-1}{2n}, \frac{n+i}{2n}]}(U) \, t_{i-1}(\tilde \omega)$ as in Corollary~\ref{cor:expectation weighted sum over t_i terms}.
Define $D_2 := [0,T]\times \R^d \times \R^{d_X} \times [0, 1/\epsilon]^2 \times [\epsilon, T]$ and
\begin{equation*}
    \Upsilon :=  \bar \rho_{\theta_{2,1}^{\star, m_{2,1}}}  \1_{ D_2^\complement } : \R^{2+d+d_X} \to \R^{d_X}
\end{equation*}
and note that on $D_2$ the $\epsilon$-approximation holds for any input of the form \eqref{equ:thm1 input to NN 2} and $\1_{D_2^\complement} \leq \zeta$.
Moreover, $\tilde \mu_2 (\R^{1+d+d_X}) = 1$, i.e., $\tilde\mu_2$ is a finite measure and $\Upsilon$ is a function in $L^2(\tilde\mu_2)$, since it is bounded by $\gamma_2$.
Hence, \citet[Theorem~1]{hornik1991approximation} implies that there exists $m_{2,2} =  m_{2,2}(\epsilon) \in \N$ and neural network weights $\tilde \theta^{\star, m_{2,2}}_{2,2} \in \tilde{\Theta}_{m_{2,2}}^{2}$ such that the corresponding neural network $\tilde \rho_{\tilde \theta^{\star, m_{2,2}}_{2,2}}$ satisfies $\int |\Upsilon - \tilde \rho_{\tilde \theta^{\star, m_{2,2}}_{2,2}} |_2^2 d\tilde\mu_2 < \epsilon$. Then we define the neural network $\rho_{ \theta^{\star, m_{2}}_{2}} = \bar \rho_{\bar \theta_{2,1}^{\star, m_{2,1}}} - \tilde \rho_{\tilde \theta^{\star, m_{2,2}}_{2,2}} \in \mathcal{N}$, with the weights $\theta^{\star, m_{2}}_{2} = (\bar \theta_{2,1}^{\star, m_{2,1}}, \tilde \theta^{\star, m_{2,2}}_{2,2})$ and $m_2 = m_{2,1} + m_{2,2}$.
Note that we cannot approximate $f,F$ or the difference between them and their primal approximating neural networks directly, since the input space of $f,F$ is infinite dimensional, which is not covered by the universal approximation results.
Setting $m := \max(m_0, m_1, m_2, \gamma_1, \gamma_2, | \theta_1^{\star, m_1}|_2, | \theta_2^{\star, m_2}|_2)$, it follows that $\theta^\star_m := (\theta^{\star, m_1}_1, \theta^{\star, m_2}_2, \tilde \theta^\star_3 ) \in \Theta_m$.

Now we can bound the distance between $Y_t^{\theta_m^\star}$ and $\hat{X}$.
Let $F = (F_j)_{1 \leq j \leq d_X}$ and $f = (f_j)_{1 \leq j \leq d_X}$.
Then we have for $t \in \{ t_1, \dotsc, t_n \} \subset [0,T]$,
\begin{equation*}
\begin{split}
    \left\lvert Y_t^{\theta_m^\star} - \hat{X}_t  \right\rvert_1
    & = \left\lvert  \rho_{\theta_2^{\star, m_2}}\left(t,\pi_{m}( \tilde X^{\leq t} - X_0 ), X_0, \tilde X^\star_t, n_t, \delta_t \right) - F \left(t, t, \tilde X^{\leq t} \right) \right\rvert_1  \\
    &= \left\lvert  \rho_{\theta_2^{\star, m_2}} - F \right\rvert_1  (\1_{D_2} + \1_{D_2^\complement}) \\
    &\leq \left( \left\lvert  \bar \rho_{\bar \theta_{2,1}^{\star, m_{2,1}}} - F \right\rvert_1  + \left\lvert  \tilde \rho_{\tilde \theta^{\star, m_{2,2}}_{2,2}} \right\rvert_1 \right) \1_{D_2} + \left( \left\lvert  \bar \rho_{\bar \theta_{2,1}^{\star, m_{2,1}}} - \tilde \rho_{\tilde \theta^{\star, m_{2,2}}_{2,2}} \right\rvert_1 +  \left\lvert F \right\rvert_1 \right) \1_{D_2^\complement} \\
    &\leq \left( \epsilon d_X  + \left\lvert   \Upsilon - \tilde \rho_{\tilde \theta^{\star, m_{2,2}}_{2,2}} \right\rvert_1 \right) \1_{D_2} + \left( \left\lvert  \Upsilon - \tilde \rho_{\tilde \theta^{\star, m_{2,2}}_{2,2}} \right\rvert_1  +  \left\lvert F \right\rvert_1 \right) \1_{D_2^\complement} \\
    & \leq \epsilon d_X  + \left\lvert   \left(\Upsilon - \tilde\rho_{\tilde \theta^{\star, m_{2,2}}_{2,2}}\right) (t) \right\rvert_1 +  \left\lvert F \right\rvert_1 \zeta,
\end{split}
\end{equation*}
where we write $\left(\Upsilon - \tilde\rho_{\tilde \theta^{\star, m_{2,2}}_{2,2}}\right) (t)$ as an abbreviation to clarify the first input ($t$).
If $t \in [0,T] \setminus \{ t_1, \dotsc, t_n \}$,
\begin{equation*}
\begin{split}
     \big\lvert  Y_t^{\theta_m^\star} &- \hat{X}_t  \big\rvert_1  \\
    &\leq \left\lvert Y_{\tau(t)}^{\theta_m^\star} - \hat{X}_{\tau(t)}  \right\rvert_1 \\
    & \quad +  \int_{\tau(t)}^t \left\lvert f_{\theta_1^{\star, m_1}}\left(s, \tau(t), \pi_{m} (\tilde X^{\leq \tau(t)} -X_0 ), X_0, \tilde X^\star_t, n_t, \delta_t \right)
		- f(s, \tau(t), \tilde X^{\leq \tau(t)})  \right\rvert_1 ds\\
    & \leq \left( \epsilon d_X  + \left\lvert  \left( \Upsilon - \tilde\rho_{\tilde \theta^{\star, m_{2,2}}_{2,2}} \right) (\tau(t))\right\rvert_1 +  \left\lvert F \right\rvert_1 \zeta\right)
    + \int_{0}^T \left( \epsilon d_X  + \left\lvert   \upsilon - \tilde f_{\tilde \theta^{\star, m_{1,2}}_{1,2}} \right\rvert_1 +  \left\lvert f \right\rvert_1 \zeta \right) ds,
\end{split}
\end{equation*}
using the approximation steps as in the previous bound.
By equivalence of the $1$- and $2$-norm, there exists a constant $c > 0$ such that for all $t \in [0,T]$ we have with the two bounds above that
\begin{equation*}
\begin{split}
\left\lvert Y_t^{\theta_m^\star} - \hat{X}_t  \right\rvert_2
	& \leq c \left( \epsilon d_X (T+1) + (\left\lvert F \right\rvert_2 + T \left\lvert f \right\rvert_2)  \zeta + \left\lvert   \Upsilon - \tilde \rho_{\tilde \theta^{\star, m_{2,2}}_{2,2}} \right\rvert_2 + \int_{0}^T \left\lvert   \upsilon - \tilde f_{\tilde \theta^{\star, m_{1,2}}_{1,2}} \right\rvert_2 ds \right).
\end{split}
\end{equation*}
Hence, we have with $C>0$ absorbing all constant terms (and therefore changing from line to line)
\begin{equation}\label{equ:thm1 bounding the difference Y - hat X}
\begin{split}
    &\E_{\P\times\tilde\P} \left[ \frac{1}{n} \sum_{i=1}^n \left\lvert Y_{t_i-}^{\theta_m^\star} - \hat{X}_{t_i-}  \right\rvert_2^2 \right]
    + \E_{\P\times\tilde\P} \left[ \frac{1}{n} \sum_{i=1}^n \left\lvert Y_{t_i}^{\theta_m^\star} - \hat{X}_{t_i}  \right\rvert_2^2 \right]\\
    & \leq C \left(
        \epsilon^2
        + \E_{\P\times\tilde\P}\left[ (\left\lvert F \right\rvert_2^2 + \left\lvert f \right\rvert_2^2)  \zeta  \right]
        + \E_{\P\times\tilde\P} \left[ \frac{1}{n} \sum_{i=1}^n \left\lvert   \left( \Upsilon - \tilde\rho_{\tilde \theta^{\star, m_{2,2}}_{2,2}} \right) (t_{i-1}) \right\rvert_2^2 \right] \right. \\
        &\quad + \E_{\P\times\tilde\P} \left[ \frac{1}{n} \sum_{i=1}^n \left\lvert   \left( \Upsilon - \tilde\rho_{\tilde \theta^{\star, m_{2,2}}_{2,2}} \right) (t_{i}) \right\rvert_2^2 \right]
        + \left. \E_{\P\times\tilde\P} \left[ T^2 \left( \int_{0}^T \left\lvert   \upsilon - \tilde f_{\tilde \theta^{\star, m_{1,2}}_{1,2}} \right\rvert_2 \tfrac{1}{T} ds \right)^2 \right]
    \right) \\
    & \leq C \left(
        \epsilon^2
        + \E_{\P\times\tilde\P}\left[ (X_T^\star +1)^{2p}  \zeta  \right]
        + \E_{\mu_u \times (\P\times\tilde\P)} \left[ \left\lvert   \left( \Upsilon - \tilde\rho_{\tilde \theta^{\star, m_{2,2}}_{2,2}} \right) (\bar t) \right\rvert_2^2 \right] \right. \\
        &\quad\qquad+ \left. \E_{\P\times\tilde\P} \left[ T  \int_{0}^T \left\lvert   \upsilon - \tilde f_{\tilde \theta^{\star, m_{1,2}}_{1,2}} \right\rvert_2^2  ds \right]
    \right) \\
    & = C \left(
        \epsilon^2
        + \E_{\P\times\tilde\P}\left[ (X_T^\star +1)^{2p}  \zeta  \right]
        + \int \left\lvert    \Upsilon - \tilde\rho_{\tilde \theta^{\star, m_{2,2}}_{2,2}} \right\rvert_2^2~d \tilde\mu_2
        + T \int \left\lvert   \upsilon - \tilde f_{\tilde \theta^{\star, m_{1,2}}_{1,2}} \right\rvert_2^2  d \tilde\mu_1
    \right) \\
    & = C \left(
        \epsilon^2
        + \E_{\P\times\tilde\P}\left[ (X_T^\star +1)^{2p}  \zeta  \right] + \epsilon
    \right) ,
\end{split}
\end{equation}
where we used Cauchy-Schwarz inequality for the first inequality. For the second inequality we used Assumption~\ref{assumption:4} to bound $f,F$ in the second term, Corollary~\ref{cor:expectation weighted sum over t_i terms} for the 3rd and 4th term and Jensen's inequality for the last term. For the 3rd (in)equality we used the definition of the measures $\tilde \mu_1, \tilde\mu_2$ and for the last one we used the $L^2$-approximation of $\upsilon, \Upsilon$ with the respective neural networks.

So far, we have fixed an $\varepsilon >0$ and argued that there exists some $ m \in \N$ such that the neural network approximation bounds hold with $\varepsilon$-error.
However, what we actually need to show is that this error converges to $0$ when increasing the  truncation level and network size $m$.
Therefore, we define $\varepsilon_m \geq 0$ to be the smallest number such that the above bounds hold with error $\varepsilon_m$ when using an architecture with signature truncation level  $m \in \N$ and weights in $\Theta_m$. Since increasing $m$ can only make the approximations better (by setting the new weights to $0$, the same approximation error as before is achieved, but potentially there exists a better choice), we have $\varepsilon_{m_1} \geq \varepsilon_{m_2}$ for any $m_1 \leq m_2$. In particular $(\varepsilon_m)_{m \geq 0}$ is a decreasing sequence, hence, our derivations before prove that $\lim_{m \to \infty} \varepsilon_m = 0 $. In the following we denote by $\theta^\star_m \in \Theta_m$ the best choice for the weights within the set $\Theta_m$ to approximate the functions $F_j, f_j$.

Note that Assumptions~\ref{assumption:4} and~\ref{assumption:5} imply that there exists $c_4>0$ such that
\begin{equation}\label{equ:thm1 bound for X - hat X}
    \E_{\P\times\tilde\P} \left[ \frac{1}{n}  \sum_{i=1}^n |X_{t_i} - \hat X_{t_i-}|_2^2 \right] \leq c_4
\end{equation}
and that $M_{t_i} \odot \hat{X}_{t_i} = M_{t_i} \odot X_{t_i}$.
Since $\theta_m^{\min} \in \argmin _{\theta \in \Theta_m}\{ \Phi(\theta) \}$  (note that at least one minimum exists in the compact set $\Theta_m$ since $\Phi$ is continuous), we get for $\bar t := \sum_{i=1}^n \1_{(\frac{i-1}{n}, \frac{i}{n}]}(U) \, t_i$ with $U \sim \operatorname{Unif}([0,1])$ independent of $\mathcal{F}\otimes\tilde{\mathcal{F}}$ and  $R_{t_i} := \left\lvert  \hat{X}_{t_{i-}} - \ Y_{t_{i-}}^{\theta_m^{\star}} \right\rvert_2 + \left\lvert  \hat{X}_{t_i} - \ Y_{t_i}^{\theta_m^{\star}} \right\rvert_2$
\begin{align}\label{equ:phi convergence 1}
\min_{Z \in \mathbb{D}}
    & \Psi(Z)
    \leq \Phi(\theta_m^{\min}) \leq \Phi(\theta_m^{\star}) \notag \\
    & = \E_{\P\times\tilde\P}\left[ \frac{1}{n} \sum_{i=1}^n \left(  \left\lvert M_{t_i} \odot ( X_{t_i} - Y_{t_i}^{\theta_m^{\star}} ) \right\rvert_2 + \left\lvert M_{t_i} \odot (  Y_{t_i}^{\theta_m^{\star}} - Y_{t_{i}-}^{\theta_m^{\star}} ) \right\rvert_2 \right)^2 \right] \notag \\
    & \leq \E_{\P\times\tilde\P}\left[ \frac{1}{n} \sum_{i=1}^n \left(  \left\lvert M_{t_i} \odot ( \hat{X}_{t_i} - \ Y_{t_i}^{\theta_m^{\star}}) \right\rvert_2 + \left\lvert M_{t_i} \odot ( Y_{t_i}^{\theta_m^{\star}} - \hat{X}_{t_i} ) \right\rvert_2 \right. \right. \notag\\
    &			\left.\left. \qquad \qquad \qquad \quad + \left\lvert M_{t_i} \odot ( \hat{X}_{t_i} -  \hat{X}_{t_i-} ) \right\rvert_2  + \left\lvert M_{t_i} \odot ( \hat{X}_{t_i-} - Y_{t_{i}-}^{\theta_m^{\star}} ) \right\rvert_2 \right)^2 \right] \notag\\
    & \leq \E_{\P\times\tilde\P}\left[ \frac{1}{n} \sum_{i=1}^n \left( \left\lvert M_{t_i} \odot ( X_{t_i} -  \hat{X}_{t_i-} ) \right\rvert_2  + 2 R_{t_i} \right)^2 \right] \\
    & = \E_{\mu_U \times (\P\times\tilde\P)}\left[  \left( \left\lvert M_{\bar t} \odot ( X_{\bar t} -  \hat{X}_{\bar t-} ) \right\rvert_2  + 2 R_{\bar t} \right)^2 \right] \notag\\
    & \leq \left(
        \E_{\mu_U \times (\P\times\tilde\P)}\left[  \left\lvert M_{\bar t} \odot ( X_{\bar t} -  \hat{X}_{\bar t-} ) \right\rvert_2^2 \right]^{1/2}
        + \E_{\mu_U \times (\P\times\tilde\P)}\left[  4 R_{\bar t}^2 \right]^{1/2}
    \right)^2 \notag\\
    & = \Psi(\hat{X}) + 2 c_4^{1/2} \E_{\P\times\tilde\P}\left[ \frac{1}{n} \sum_{i=1}^n  4 R_{t_i}^2 \right]^{1/2} + \E_{\P\times\tilde\P}\left[ \frac{1}{n} \sum_{i=1}^n  4 R_{t_i}^2 \right] \notag
\end{align}
where we used Lemma~\ref{lem:expectation weighted sum over t_i terms} in the 6th (in)equality, the triangle inequality for the $L^2$-norm in the $7$th (in)equality and again Lemma~\ref{lem:expectation weighted sum over t_i terms} as well as \eqref{equ:thm1 bound for X - hat X} in the last (in)equality.
Integrability of $|X_T^\star|_2$ and $|n|$ (Assumptions~\ref{assumption:5} and~\ref{assumption:6}) together with Assumption~\ref{assumption:2} on $\delta$ imply that
\[\zeta = \1_{\{ X_T^\star \geq 1/\varepsilon_m \}} + \1_{\{ n \geq  1/\varepsilon_m \}} + \1_{\{ \delta \leq \varepsilon_m \}} \xrightarrow[m \to \infty]{\P\times\tilde\P-a.s.} 0.\]
Therefore, we have for a suitable constant $C>0$ (not depending on $\varepsilon_m$ and $m$), using Cauchy-Schwarz inequality and \eqref{equ:thm1 bounding the difference Y - hat X}
\begin{equation*}
\begin{split}
    \E_{\P\times\tilde\P}\left[ \frac{1}{n} \sum_{i=1}^n  R_{t_i}^2 \right]
    & \leq 2 \left( \E_{\P\times\tilde\P} \left[ \frac{1}{n} \sum_{i=1}^n \left\lvert Y_{t_i-}^{\theta_m^\star} - \hat{X}_{t_i-}  \right\rvert_2^2 \right]
    + \E_{\P\times\tilde\P} \left[ \frac{1}{n} \sum_{i=1}^n \left\lvert Y_{t_i}^{\theta_m^\star} - \hat{X}_{t_i}  \right\rvert_2^2 \right] \right) \\
    & \leq C \left(
        \epsilon_m^2 + \epsilon_m
        + \E_{\P\times\tilde\P}\left[ (X_T^\star +1)^{2p}  \zeta  \right]
    \right) \xrightarrow{m \to \infty} 0,
\end{split}
\end{equation*}
where convergence follows by dominated convergence, since $X_T^\star$ is $L^{2p}$-integrable by Assumption~\ref{assumption:5} (and using the inequality $|a+b|^q \leq 2^{q-1} (|a|^q+|b|^q)$ for  $q \geq 1$).
Using this and $\Psi(\hat{X}) = \min_{Z \in \mathbb{D}} \Psi(Z)$, we get from \eqref{equ:phi convergence 1}
\begin{equation*}
\min_{Z \in \mathbb{D}} \Psi(Z)
	 \leq \Phi(\theta_m^{\min}) \leq \Phi(\theta_m^{\star}) \xrightarrow{m \to \infty} \min_{Z \in \mathbb{D}} \Psi(Z)  .
\end{equation*}

Finally, we show that $\lim_{m\to\infty} d_k \left( \hat{X},  Y^{\theta_m^{\min}} \right) = 0$ for all $1 \leq k \leq K$.
First note that the triangle inequality and Lemma~\ref{lem:L2 identity} yield
\begin{equation}\label{equ:proof thm bound 1}
\begin{split}
\Phi(\theta_m^{\min}) - \Psi(\hat{X})
	&\geq \E_{\P\times\tilde\P}\left[ \frac{1}{n}\sum_{i=1}^n   \left\lvert M_{t_i} \odot ( X_{t_i} - Y^{\theta_m^{\min}}_{t_{i}-} )\right\rvert_2^2 \right] - \Psi(\hat{X}) \\
	&= \E_{\P\times\tilde\P}\left[\frac{1}{n} \sum_{i=1}^n  \left\lvert M_{t_i} \odot ( \hat{X}_{t_i-} - Y^{\theta_m^{\min}}_{t_{i}-} ) \right\rvert_2^2 \right] .
\end{split}
\end{equation}
Hence, applying \eqref{equ:HI}, \eqref{equ:M split}, Proposition~\ref{prop:mu_k expectation equiv} and \eqref{equ:definition of d_k} in  reverse order than it was done in  \eqref{equ:thm1-positive expectation} and finally \eqref{equ:proof thm bound 1}, yields
\begin{equation}\label{equ:convergence in L1}
\begin{split}
d_k \left( \hat{X} , Y^{\theta_m^{\min}}  \right)
	& \leq \frac{c_0 \, c_1 \, c_3}{c_2} \, \E_{\P \times \tilde\P}\left[ \frac{1}{n} \sum_{i=1}^n \left\lvert M_{t_i} \odot ( \hat{X}_{t_i-} - Y^{\theta_m^{\min}}_{t_i-} ) \right\rvert_2^2 \right]^{1/2} \\
	& \leq   \frac{c_0 \, c_1 \, c_3}{c_2} \, \left( \Phi(\theta_m^{\min}) - \Psi(\hat{X}) \right)^{1/2}  \xrightarrow{m \to \infty} 0,
\end{split}
\end{equation}
which completes the proof.
\end{proof}

\begin{rem}[Continuation of Remark {\ref{rem:extension assumptions M0}} and Remark {\ref{rem:extension PD-NJ-ODE for other M0}}] \label{rem:extension convergence result for other M0}
In the case that $X_0$ is not observed completely, we only need to check that  $\left\lvert Y_t^{\theta_m^\star} - \hat{X}_t  \right\rvert_2  \leq  c_m$ is still satisfied, which amounts to showing that $\left\lvert Y_0^{\theta_m^\star} - \hat{X}_0  \right\rvert_1  \leq  \frac{c_m}{c (T+1)}$. We distinguish again between the two cases.
\begin{enumerate}
\item If $M_0=0$, we have
\begin{equation*}
\left\lvert Y_0^{\theta_m^\star} - \hat{X}_0  \right\rvert_1 = \left\lvert \zeta_{\theta_4^{\star, m}}(0) - \E[X_0]  \right\rvert_1 \leq \frac{c_m}{c (T+1)} ,
\end{equation*}
using that the constant $\E[X_0]$ can be approximated arbitrarily  well by the bounded output neural network on the compact subset $\{  0\}$.
\item If $M_{0, j} = \1_{I_0}(j)$, then $\E[X_0 | \operatorname{proj}_{I_0} (X_0) ]$ is a continuous function in the input $\operatorname{proj}_{I_0} (X_0)$ (by Assumption~\ref{assumption:4}) and can therefore be approximated arbitrarily well by $\zeta_{\theta_4} \in \mathcal{N}$ on any compact subset. It is enough to note that $X_T^\star < 1/\varepsilon$ implies that $\operatorname{proj}_{I_0} (X_0)$ lies in a compact set to conclude that
\begin{equation*}
\left\lvert Y_0^{\theta_m^\star} - \hat{X}_0  \right\rvert_1 = \left\lvert \zeta_{\theta_4^{\star, m}}( \operatorname{proj}_{I_0} (X_0) ) - \E[X_0 | \operatorname{proj}_{I_0} (X_0) ]  \right\rvert_1 \leq \frac{c_m}{c (T+1)}.
\end{equation*}
\end{enumerate}
\end{rem}

\subsection{Convergence of the Monte Carlo approximation}\label{sec:Convergence of the Monte Carlo approximation}
We now  assume the size $m$ of the neural network and of the signature truncation level is fixed and we study the convergence of the Monte Carlo approximation when the number of samples $N$ increases.
Moreover, we show that both types of convergence can be combined.
The convergence analysis is based on \citet[Chapter 4.3]{lapeyre2019neural} and follows \citet[Theorem E.13]{herrera2021neural}.
\begin{theorem}\label{thm:MC convergence Yt}
Let $\theta^{\min}_{m,N} \in \Theta^{\min}_{m,N} := \argmin_{\theta \in \Theta_m}\{ \hat\Phi_N(\theta)\}$ for every $m, N \in \N$.
Then, for every $m \in \N$, $(\P\times\tilde\P)$-a.s.
\begin{equation*}
\hat\Phi_N \xrightarrow{N \to \infty} \Phi \quad \text{uniformly on } \Theta_m.
\end{equation*}
Moreover, for every $m \in \N$, we have a.s.,
\begin{equation*}
\Phi(\theta^{\min}_{m,N}) \xrightarrow{N \to \infty} \Phi(\theta^{\min}_{m}) \quad \text{and} \quad \hat\Phi_N(\theta^{\min}_{m,N}) \xrightarrow{N \to \infty} \Phi(\theta^{\min}_{m}) .
\end{equation*}
In particular, one can define an increasing random sequence $(N_m)_{m \in \N}$ in $\N$ such that for every $1 \leq k \leq K$ we have almost surely that $Y^{\theta_{m, N_m}^{\min}}$ converges to $\hat{X}$
in the metric $d_k$  as $m \to \infty$.

\end{theorem}

We define the separable Banach space $\mathcal{S} := \{ x = (x_i)_{ i \in \N} \in \ell^1(\R^{d}) \; \vert \; \lVert x \rVert_{\ell^1} < \infty \}$ for a suitable $d$ (see below) with the norm $\lVert x \rVert_{\ell^1} := \sum_{i \in \N} \lvert x_i \rvert_2$, the function
\begin{equation*}
F(x,y,z,m) :=  \left\lvert m \odot ( x - y ) \right\rvert_2 + \left\lvert m \odot (y - z) \right\rvert_2
\end{equation*}
and $\xi_j := (\xi_{j, 0}, \dotsc, \xi_{j, n^{(j)}}, 0, \dotsc)$, where
\[\xi_{j,k} := (t_{k}^{(j)}, X_{t_k^{(j)}}^{(j)}, M_{t_k^{(j)}}^{(j)}, \pi_m(\tilde X^{\leq t_k^{(j)}, {(j)}} - X_0^{(j)}), \tilde X^{\star, (j)}_t, k, \delta_{t_k^{(j)}}^{(j)}) \in \R^d\]
and $t_k^{(j)}$, $M_{t_k^{(j)}}^{(j)}$ and $X^{(j)}_{t^{(j)}_i}$ (with $0$ entries for coordinates which are not observed) are random variables describing the $j$th realization of the training data, as defined in Section~\ref{sec:Objective Function}.
Let $n^{j}(\xi_j) := \max_{k \in \N}\{ \xi_{j,k} \neq 0\}$, $t_k(\xi_j):= t_{k}^{(j)}$, $X_k (\xi_j) := X_{t_k^{(j)}}^{(j)}$ and $M_k(\xi_j) := M_{t_k^{(j)}}^{(j)}$.
By this definition we have  $n^{(j)} = n^{j}(\xi_j)$ $(\P \times \tilde{\P})$-almost-surely. Moreover, we have that $\xi_j$ are i.i.d. random variables taking values in $\mathcal{S}$.
Let us write $Y_t^{\theta}(\xi)$ to make the dependence of $Y$ on the input and the weight $\theta$ explicit.
Then we define
\begin{equation*}
h(\theta, \xi_j) := \frac{1}{n^{j}(\xi_j)}\sum_{i=1}^{n^{j}(\xi_j)}  F \left( X_i(\xi_j), Y^\theta_{t_i(\xi_j)}(\xi_j), Y^\theta_{t_i(\xi_j)-}(\xi_j), M_i(\xi_j) \right)^2.
\end{equation*}

\begin{lemma}\label{lem:properties for MC conv thm}
Almost-surely the random function $\theta\in\Theta_M \mapsto Y_{t}^{\theta}$ is uniformly continuous for every $t \in [0,T]$.
\end{lemma}

\begin{proof}
Since the activation functions of the neural networks are continuous, also the neural networks are continuous with respect to their weights $\theta$, which implies that also $\theta\in \Theta_M \mapsto Y_{t}^{\theta}$ is continuous. Since ${\Theta}_M$ is compact, this automatically yields uniform continuity.
\end{proof}

The following lemma is a consequence of \cite[Corollary 7.10]{ledoux1991m} and \cite[Sec. 2.6, Lemma A1 \& Theorem A1 and discussion thereafter]{rubinstein1993discrete}.

\begin{lemma}
\label{lemma:convergencelocallyuniform}
Let $(\xi_i)_{i \geq 1}$ be a sequence of $i.i.d.$ random variables with values in $\mathcal{S}$ and $h:\mathbb{R}^d\times \mathcal{S}\to \mathbb{R}$ be a measurable function.
Assume that a.s., the function $\theta\in \mathbb{R}^d \mapsto h(\theta, \xi_1)$ is continuous and for all $C>0$, $\E(\sup_{|\theta|_2 \leq C}|h(\theta, \xi_1)|)< + \infty$. Then, a.s. $f_N:  \mathbb{R}^d \to \R, \theta  \mapsto \frac{1}{N}\sum_{i=1}^{N}h(\theta, \xi_i)$ converges locally uniformly to the continuous function $f: \mathbb{R}^d \to \R, \theta \mapsto \E(h(\theta, \xi_1))$, i.e.,
\begin{equation*}
\lim_{N\to\infty} \sup_{|\theta|_2\leq C} \left|\frac{1}{N}\sum_{i=1}^{N}h(\theta, \xi_i) -   \E(h(\theta, \xi_1))\right| = 0 \qquad a.s.
\end{equation*}
Moreover, let $K \subset \R^d$ be compact and define the random variables $v_n := \inf_{x\in K} f_n(x)$. We consider a minimizing sequence of random variables $(x_n)_{n=0}^{\infty}$, given by $f_n(x_n) = \inf_{x\in K} f_n(x) $ and let $v^\star = \inf_{x\in K}f(x)$ and $\mathcal{K}^\star = \{ x\in K: f(x) =v^\star \}$. Then $v_n \to v^\star$ and $d(x_n, \mathcal{K}^\star) \to 0 $ a.s.
\end{lemma}

\begin{proof}[Proof of Theorem {\ref{thm:MC convergence Yt}}.]
First, we note that $Y^\theta_t$ is obtained by integrating the output of (bounded-output) neural networks and is therefore bounded in terms of the input, the weights (which are bounded by $m$), $T$, and a constant depending on the architecture and activation functions of the neural network. In particular, $|Y^\theta_t(\xi_j)| \leq \tilde B(1+ X^{\star,(j)})^p$ for all $t \in [0,T]$ and $\theta \in \Theta_m$, for some constant $\tilde B$ (possibly depending on $m$), where $X^{\star,(j)}$ corresponds to the input $\xi_j$.
Hence,
\begin{align*}
&F \left( X_i(\xi_j), Y^\theta_{t_i(\xi_j)}(\xi_j), Y^\theta_{t_i(\xi_j)-}(\xi_j), M_i(\xi_j) \right)^2 \\&
	= \left(\left\lvert M_i(\xi_j) \odot (X_i(\xi_j) - Y^\theta_{t_i(\xi_j)}(\xi_j) ) \right\rvert_2 + \left\lvert M_i(\xi_j) \odot ( Y^\theta_{t_i(\xi_j)}(\xi_j) - Y^\theta_{t_i(\xi_j)-}(\xi_j) ) \right\rvert_2 \right)^2\\&
	\leq (4 (B+\tilde B) (1+ X^{\star,(j)})^p)^2 = 16 (B+\tilde B)^2 (1+ X^{\star,(j)})^{2p}.
\end{align*}

Hence,
\begin{equation}
\label{equ:dominating bound loss function}
\E_{\P\times\tilde\P}\left[\sup_{\theta \in \Theta_m} h(\theta, \xi_j)\right]
\leq   \E_{\P\times\tilde\P}\left[\frac{1}{n}\sum_{i=1}^{n} 16 (B+\tilde B)^2 (1+ X^{\star,(j)})^{2p} \right]  < \infty,
\end{equation}
by Assumption~\ref{assumption:5}.
By Lemma~\ref{lem:properties for MC conv thm}, the function $\theta \mapsto h(\theta, \xi_1)$ is continuous, hence, we can apply Lemma~\ref{lemma:convergencelocallyuniform}, yielding that almost-surely for $N \to \infty$ the function
\begin{equation}\label{equ:unif conv 1}
\theta \mapsto \frac{1}{N} \sum_{j=1}^{N} h(\theta, \xi_j) = \hat\Phi_N(\theta)
\end{equation}
converges uniformly on $\Theta_m$ to
\begin{equation}\label{equ:unif conv 2}
\theta \mapsto \E_{\P \times \tilde\P}[h(\theta, \xi_1)] = \Phi(\theta).
\end{equation}

Moreover, Lemma~\ref{lemma:convergencelocallyuniform} yields that $d(\theta^{\min}_{m,N},\Theta^{\min}_m)\to 0$ a.s. when $N\to \infty$.
Then there exists a sequence $(\hat\theta^{\min}_{m,N})_{N \in \N}$ in $\Theta_m^{\min}$ such that $\lvert \theta^{\min}_{m,N} - \hat\theta^{\min}_{m,N} \rvert_2 \to 0$ a.s. for $N \to \infty$.
The uniform continuity of the random functions $\theta \mapsto Y_{t}^{\theta}$ on ${\Theta}_m$ implies that for any fixed  deterministic and bounded $\xi_0$ (taking values in the same space as the $\xi_j$)
\[\lvert Y_{t}^{\theta^{\min}_{m,N}}(\xi_0) - Y_{t}^{\hat\theta^{\min}_{m,N}}(\xi_0) \rvert_2 \to 0 \text{ a.s. for all  } t \in [0,T] \text{ as } N \to \infty.\]
By continuity of $F$ this yields $\lvert h(\theta^{\min}_{m,N}, \xi_0) - h(\hat\theta^{\min}_{m,N}, \xi_0) \rvert \to 0$ a.s.\ as $N \to \infty$.
Let $\xi_0$ now be a random variable which is independent of and identically distributed as the $\xi_j$ defined on a copy $(\Omega_0, \mathbb{F}_0, \mathcal{F}_0, \P_0)$ of the filtered probability space $(\Omega\times\tilde\Omega, \mathbb{F}\times\tilde{\mathbb{F}}, \mathcal{F}\times\tilde{\mathcal{F}}, \P\times\tilde\P)$. Then the above statements hold for $\xi_0(\omega_0)$ for $\P_0$-a.e.\ fixed $\omega_0$.
Hence, we have for $(\P\times\tilde{\P})$-a.e.\ fixed $\omega \in \Omega\times\tilde\Omega$, that $\lvert h(\theta^{\min}_{m,N}\omb, \xi_0) - h(\hat\theta^{\min}_{m,N}\omb, \xi_0) \rvert \to 0$ $\P_0$-a.s.\ as $N \to \infty$.
With \eqref{equ:dominating bound loss function} we can apply dominated convergence which yields
\begin{equation*}
\lim_{N \to \infty} \E_{\xi_0}\left[ \lvert h(\theta^{\min}_{m,N}\omb, \xi_0) - h(\hat\theta^{\min}_{m,N}, \xi_0) \rvert \right] = 0 \text{ for $(\P\times\tilde{\P})$-a.e. } \omega \in \Omega\times\tilde\Omega.
\end{equation*}
Since for every integrable random variable $Z$ we have $0 \leq \lvert \E[Z] \rvert \leq \E[\lvert Z \rvert] $ and since $\hat\theta^{\min}_{m,N}\in \Theta_m^{\min}$ we can deduce that for $(\P\times\tilde{\P})$-a.e.\ fixed $\omega \in \Omega\times\tilde\Omega$,
\begin{equation}
\label{equ: MC convergence}
\lim_{N \to \infty} \Phi(\theta^{\min}_{m,N}\omb) = \lim_{N \to \infty} \E_{\xi_0}\left[  h(\theta^{\min}_{m,N}\omb, \xi_0) \right] = \lim_{N \to \infty} \E_{\xi_0}\left[  h(\hat\theta^{\min}_{m,N}\omb, \xi_0) \right] = \Phi(\theta^{\min}_m).
\end{equation}
Now by triangle inequality, for $(\P\times\tilde{\P})$-a.e.\ fixed $\omega$, we have for $\hat\Phi_{\tilde N}$ and $\Phi$ defined through test samples $\tilde{\xi}_j$ on $\Omega_0$, i.e., independent of and identically distributed as the training samples $\xi_j$ yielding $\theta^{\min}_{m,N}\omb$,
\begin{equation}\label{equ: MC convergence 2}
\lvert \hat\Phi_{\tilde N}(\theta^{\min}_{m, N}\omb) -  \Phi(\theta^{\min}_{m}) \rvert \leq \lvert \hat\Phi_{\tilde N}(\theta^{\min}_{m, N}\omb) -  \Phi(\theta^{\min}_{m, N}\omb) \rvert + \lvert \Phi(\theta^{\min}_{m, N}\omb) -  \Phi(\theta^{\min}_{m}) \rvert.
\end{equation}
\eqref{equ:unif conv 1} and \eqref{equ:unif conv 2} imply that the first term on the right hand side converges to 0 when $\tilde N \to \infty$ and \eqref{equ: MC convergence} implies that the second term on the right hand side converges to 0 a.s.\ when $ N \to \infty$.
Moreover, the uniform convergence in \eqref{equ:unif conv 1} and \eqref{equ:unif conv 2} yields the same result when setting $\tilde N = N$. Furthermore, Lemma~\ref{lemma:convergencelocallyuniform} yields the same result for $\hat\Phi_N(\theta^{\min}_{m, N})\omb$, i.e., when $\hat\Phi_{N}$ and $\Phi$ are defined through the $\xi_j$ on the probability space corresponding to the training data. This finishes the proof of the first part of the theorem.

We define $N_0 := 0$ and for every $m \in \N$
\begin{equation*}
N_m\omb := \min\left\{ N \in \N \; \vert \; N > N_{m-1}\omb, \lvert \Phi(\theta^{\min}_{m,N}\omb) - \Phi(\theta^{\min}_{m}) \rvert  \leq \tfrac{1}{m} \right\},
\end{equation*}
which is possible due to \eqref{equ: MC convergence} for $(\P\times\tilde{\P})$-a.e.\ $\omega \in \Omega\times\tilde\Omega$. Then Theorem~\ref{thm:1} implies that for $(\P\times\tilde{\P})$-a.e.\ $\omega \in \Omega\times\tilde\Omega$
\begin{equation*}
\lvert \Phi(\theta^{\min}_{m,N_m\omb}\omb) - \Psi(\hat{X}) \rvert  \leq \tfrac{1}{m} +  \lvert \Phi(\theta^{\min}_{m}) - \Psi(\hat{X}) \rvert \xrightarrow{m \to \infty} 0.
\end{equation*}
Therefore, we can apply the same arguments as in the proof of Theorem~\ref{thm:1} (starting from \eqref{equ:proof thm bound 1}) to show that
\begin{equation*}
d_k \left(  \hat{X} , Y^{\theta^{\min}_{m,N_m\omb}\omb}  \right)
\leq   \frac{c_0 \, c_1 \, c_3}{c_2} \, \left( \Phi(\theta^{\min}_{m,N_m\omb}\omb) - \Psi(\hat{X}) \right)^{1/2}  \xrightarrow{m \to \infty} 0,
\end{equation*}
for every $1 \leq k \leq K$ and for $(\P\times\tilde{\P})$-a.e.\ $\omega \in \Omega\times\tilde\Omega$.
\end{proof}

\begin{cor}\label{cor:1}
In the setting of Theorem~\ref{thm:MC convergence Yt}, we also have that $(\P \times \tilde{\P})$-a.s.
\begin{equation*}
\Phi(\theta^{\min}_{m,N_m}) \xrightarrow{m \to \infty} \Psi(\hat{X}) \quad \text{and} \quad \hat\Phi_{\tilde{N}_m}(\theta^{\min}_{m,\tilde{N}_m}) \xrightarrow{m \to \infty} \Psi(\hat{X}),
\end{equation*}
where $(\tilde{N}_m)_{m \in \N}$ is a suitable increasing random sequence in $\N$.
\end{cor}

\begin{proof}
The first convergence result was already shown in the proof of Theorem~\ref{thm:MC convergence Yt} and the second one can be shown similarly, when defining $\tilde{N}_m$ by $\tilde{N}_0 := 0$ and for every $m \in \N$
\begin{equation*}
\tilde{N}_m := \min\left\{ N \in \N \; \vert \; N > \tilde{N}_{m-1}, \lvert \hat\Phi_N(\theta^{\min}_{m,N}) - \Phi(\theta^{\min}_{m}) \rvert  \leq \tfrac{1}{m}  \right\},
\end{equation*}
which is possible due to \eqref{equ: MC convergence 2}.
\end{proof}

\begin{rem}\label{rem:equivalent objective function}
Theorems~\ref{thm:1}, {\ref{thm:MC convergence Yt}} and Corollary~\ref{cor:1} hold equivalently, when replacing the terms $(Z_{t_i} - Z_{t_i -})$ and $\left( Y_{t_i^{(j)}}^{\theta, j } - Y_{t_{i}^{(j)}-}^{\theta, j } \right)$ in \eqref{equ:Psi}, \eqref{equ:Phi} and \eqref{equ:appr loss function}, by $(X_{t_i} - Z_{t_i -})$ and $\left( X_{t_i^{(j)}}^{(j) } - Y_{t_{i}^{(j)}-}^{\theta, j } \right)$ respectively.
We will refer to this adjustment of the objective function as \emph{equivalent objective (or loss) function}.
\end{rem}

\subsection{Dependence between the process and the observation framework \& noisy observations}\label{sec:Overcoming the Independence of the Underlying Process and its Observation Times}

So far, we have focused on the case in which the observation times $t_i$ and masks $M_i$ are independent of the underlying process $X$. This was modelled using the product of two independent probability spaces. Although this still permits considerable generality, it keeps the derivations relatively simple, since Fubini's theorem can be used to split expressions into components associated with the two probability spaces.
Additionally, we have considered only noise-free observations in our framework. Although the process itself is stochastic and may therefore have noisy components, we have assumed throughout that it is observed without measurement noise and that the goal is to predict the process itself, without filtering out noise. One way to address noisy observations is through stochastic filtering, as described in Section~\ref{sec:Stochastic Filtering with PD-NJ-ODE}. However, this approach requires stronger assumptions: in particular, the distributions of the underlying process and the noise must be known, or, equivalently, training samples separated into the noise-free observation and the noise term must be available.

In the companion paper \citep{NJODE3}, we prove the equivalent theoretical results when there is dependence between the process $X$ and the observation framework and when observations are noisy under mild additional assumptions. In particular, we show that both of these generalisations do not pose a problem for our PD-NJ-ODE framework.

\section{Conditional variance, moments and moment generating function}
\label{sec:Conditional Variance, Moments and Moment Generating Function}

\subsection{Uncertainty estimation: conditional variance}
\label{sec:Uncertainty Estimation: Conditional Variance}
Let $X$ be a $d_X$-dimensional process satisfying Assumptions~\ref{assumption:1} to~\ref{assumption:6}. If $X^2$ satisfies Assumptions~\ref{assumption:4} and~\ref{assumption:5}, then also the joint process $Z:= (Z_1, Z_2) := (X, X^2)^\top$ satisfies all assumptions.
Therefore, the conditional expectation of $Z$ can be used to get an uncertainty estimate for the prediction of the process $X$ (not to be confused with an uncertainty estimate of our model output) by computing its conditional variance
\begin{equation}\label{equ:conditional variance}
\operatorname{Var}[X_t \, | \, \mathcal{A}_{\tau(t)}] = \E[X_t^2 \, | \, \mathcal{A}_{\tau(t)} ] - \E[X_t \, | \, \mathcal{A}_{\tau(t)} ]^2 = \E[(Z_2)_t \, | \, \mathcal{A}_{\tau(t)} ] - \E[ (Z_1)_t \, | \, \mathcal{A}_{\tau(t)} ]^2.
\end{equation}
In particular, when using the $2d_X$-dimensional input $Z$ for PD-NJ-ODE (where the observation mask is the same for $Z_1$ and $Z_2$), \eqref{equ:conditional variance} yields a way to compute the conditional variance of $X$.

An example of a process $X$ for which also its conditional variance can be estimated in our framework is given below.
\begin{example}[Brownian Motion and its Conditional Variance]
\label{example:BM and Cond Var}
If $X$ is a standard Brownian motion and $Z := (X, X^2)^\top$ then we have that $\E[(Z^\star_T)^p] < \infty$ for every $1 \leq p <  \infty$ (compare with Section~\ref{sec:Multivariate Process with Incomplete Observations: Correlated Brownian Motions}) and since $\operatorname{Var}[X_{t+s} \, | \, X_t] = \operatorname{Var}[X_{t+s} - X_t] = s$ we get
\[\E[Z_{t} | \mathcal{A}_{\tau(t)}] =  (X_{\tau(t)} , X_{\tau(t)}^2 + (t - \tau(t)))^\top = ((Z_1)_{\tau(t)} , (Z_2)_{\tau(t)}+ (t - \tau(t)))^\top.\]
Hence, $f_1(s, \tau(t), \tilde X^{\leq \tau(t)} ) = 0$ and $f_2(s, \tau(t), \tilde X^{\leq \tau(t)} ) = 1$ and therefore Assumptions~\ref{assumption:4} and~\ref{assumption:5} are satisfied for $Z$.

\end{example}

\subsection{Towards the conditional distribution: moments and the moment generating function}
\label{sec:Towards the Conditional Distribution: Moments and the Moment Generating Function}
For simplicity, we assume in the following that the process $X$ is one-dimensional, although the considerations generalize to higher dimensions.
If the process $X$ is such that all its moments $X^n$ for $n \in \N$ satisfy Assumptions~\ref{assumption:4} and~\ref{assumption:5}, then, in principle, PD-NJ-ODE can be used to compute all conditional moments of $X$. Theorem 3.3.11 and the subsequent remark in \citet{Durrett:2010:PTE:1869916} give two conditions on the moments under which they uniquely characterize the corresponding distribution. Hence, if either condition is satisfied, PD-NJ-ODE can, in principle, characterize the conditional distribution of $X_t$ given $\mathcal{A}_{\tau(t)}$ for any $t \in [0,T]$.

Under the stronger assumption that there exists some $\delta > 0$ such that for all $u \in (- \delta, \delta)$ the process $\exp(u X )$ satisfies Assumptions~\ref{assumption:4} and~\ref{assumption:5}, PD-NJ-ODE can in principle be used to compute the conditional moment generating function of $X$, $\hat m_X(u)_t := \E[\exp(u X_t) | \mathcal{A}_{\tau(t)} ]$ for $u \in (- \delta, \delta)$, for any $t \in [0,T]$. Since the moment generating function on any open interval including $0$ uniquely characterizes the corresponding distribution \citep[Section~30]{Billingsley1995Prob}, PD-NJ-ODE then, in principle, characterizes the conditional distribution.

In practice, only finitely many moments can be computed, and the moment generating function can be approximated only by evaluating it at finitely many values of $u$. However, the two methods described above provide a way for PD-NJ-ODE to characterize the conditional distribution of $X$ approximately.

\section{Stochastic filtering with PD-NJ-ODE}\label{sec:Stochastic Filtering with PD-NJ-ODE}
In this section we show that the PD-NJ-ODE can be used to approximate solutions to the stochastic filtering problem with discrete observations.
In Section~\ref{sec:Stochastic Filtering Problem}, we first briefly introduce the stochastic filtering problem, following \citet{bain2009fundamentals}, and then explain how PD-NJ-ODE can be used to solve it in Section~\ref{sec:Applying PD-NJ-ODE to the Filtering Problem}. Moreover, in Appendix~\ref{sec:A Generalized Version of the Filtering Problem}, we discuss solving a generalized version of the stochastic filtering problem.

\subsection{Stochastic filtering problem}\label{sec:Stochastic Filtering Problem}
In the filtering framework, one is interested in a \emph{signal process} $X \in \R^{d_X}$, defined as in Section~\ref{sec:Stochastic Process, Random Observation Times and Observation Mask}. However, this signal process is never observed. Instead one observes the \emph{observation (or sensor) process} $Y \in \R^{d_Y}$ defined as
\begin{equation*}
Y_t := Y_0 + \int_0^t h(X_s) ds + W_t,
\end{equation*}
where $h:\R^{d_X} \to \R^{d_Y}$ is a measurable function and $W$ is a standard $\mathbb{F}$-adapted $d_Y$-dimensional Brownian motion independent of $X$.
Defining the available information at time $t \in [0,T]$ as $\tilde{ \mathcal{Y}}_t := \boldsymbol{\sigma}\left(Y_s | s \in [0,t]\right)$, the filtering problem consists of determining the conditional distribution of $X_t$ given the currently available information $\mathcal{Y}_t$. In particular, this means computing the conditional expectations
\begin{equation*}
\E[\varphi(X_t) \, | \, \tilde{ \mathcal{Y}} ],
\end{equation*}
for any measurable function $\varphi$ for which the integral is well defined \cite[Definition~3.2]{bain2009fundamentals}.
A typical assumption in the filtering problem is that the distribution of $X$ is known (e.g. in the sense that $X$ is given as the solution of an It\^o-diffusion with known coefficients \citep[Section~3.2.1]{bain2009fundamentals}) and that the function $h$ is known. In particular, this implies that the joint law of $(X,Y)$ is known.

\begin{rem}\label{rem:more general filtering formulations}
There are more general formulations of the filtering problem, as for example in \citet[Section~22]{cohen2015stochastic}. The following applications of PD-NJ-ODE work identically in these settings as long as we have access to the joint law of $(X,Y)$.
\end{rem}

In real-world filtering tasks, the observation process $Y$ can only be observed at finitely many observation times $t_i$, which we can again model as described in Section~\ref{sec:Stochastic Process, Random Observation Times and Observation Mask}. In particular, we assume here that $Y_0$ is always observed, although the framework could easily be extended to the setting where this is not the case.
The available information at any time $t \in [0,T]$ can be described similarly as in Section~\ref{sec:information sigma-algebra} by
\begin{equation*}
{ \mathcal{Y}}_t := \boldsymbol{\sigma}\left(Y_{t_i}, t_i | t_i \leq t \right).
\end{equation*}
Hence, the resulting filtering problem is to compute for any integrable measurable function $\varphi : \R^{d_X} \to \R^{d_\varphi}$ the conditional expectation
 \begin{equation*}
\E[\varphi(X_t) \, | \, {\mathcal{Y}}_t].
\end{equation*}

\subsection{Applying PD-NJ-ODE to the filtering problem}\label{sec:Applying PD-NJ-ODE to the Filtering Problem}
At first sight one might think that our model framework is not compatible with the filtering problem, since our results depend on the assumption that every coordinate is observed with positive probability at any observation time, while obviously, in the filtering task, the signal process $X$ is never observed. However, the important subtlety is that this is only true for the samples on which we want to \emph{evaluate} the trained PD-NJ-ODE model, while for the training of the  model we can make use of the knowledge of the law of $(X,Y)$ to generate a suitable training set.

In particular, we generate i.i.d. samples of $Z:=(\varphi(X),Y) \in \R^{d_\varphi + d_Y}$ together with i.i.d. samples of the number of observations and the observation times $(n, t_1, \dotsc, t_n)$. Moreover, we generate observation masks $M_k \in \{ 0,1\}^{d_\varphi + d_Y}$, where the coordinates corresponding to $Y$ are always $1$ and the coordinates corresponding to $X$ are either all $1$ or all $0$ simultaneously, with the probability of being $1$ given by $p_0 =0$ and $p_k \in (0,1)$ for $k \geq 1$.
It is crucial that $p_k \ne 0$, since otherwise this would contradict Assumption~\ref{assumption:1}, and that $p_k \ne 1$ for the following reason.
If $p_k$ was $1$, the signal process $X$ would always be observed at the $k$th observation time, implying that  the probability of having a sample with at least $k$ observations where only the $Y$-coordinates are observed is $0$ in the distribution of the training set.\footnote{Importantly, the expectations in our theoretical results are taken with respect to the (theoretical) distribution of the training set. In particular, for samples that do not lie in a subset with positive probability in the distribution of the training set, these results do not apply.}
On the other hand, for any fixed (and finite) number of observations $n$, any choice of $p_k \in (0,1)$ yields the probability $\prod_{k=1}^n (1-p_k) > 0$ that the signal process $X$ is never observed at the $n$ observation times. Hence, the samples considered when evaluating the model on the filtering task, for which only the $Y$-coordinates are observed, occur in the training set with positive probability, ensuring that the theoretical results apply.

To make this precise, let $\operatorname{proj}_X : \R^{d_\varphi + d_Y} \to \R^{d_\varphi}$ be the projection on the $X$-coordinates and let us define the probability measure $\nu_k$ similar to $\mu_k$ as
\begin{equation}\label{equ:nu-k}
\nu_k(\cdot) := \lambda_k(\cdot \, | \, t_k- \ne \infty, \forall i < k: \operatorname{proj}_X(M_i)=0 ),
\end{equation}
where we note that $\forall i < k: \operatorname{proj}_X(M_i)=0$ can equivalently be expressed in terms of $\tilde{Z}^{\leq t_{k-1}}$ and we recall that $t_k- \ne \infty$ is equivalent to $n \geq k$.
Independence of $n$ and $M_j$ together with the arguments above imply that
\[c_4 := \tilde\P(n \geq k, \forall i < k: \operatorname{proj}_X(M_i) = 0 ) > 0.\]
Hence,  \eqref{equ:nu-k} is a well defined probability measure for $1 \leq k \leq K$.
Similarly to the pseudometric $d_k$ we define the pseudometric $\tilde{d}_k$ on the set of  c\`adl\`ag $\mathbb{A}$-adapted processes by
\begin{equation}\label{equ:definition of tilde d_k}
\tilde{d}_k(Z, \xi) := \E_{\nu_k} \left[ |Z - \xi|_2 \right].
\end{equation}

To prevent confusion, we will call the PD-NJ-ODE output $G$ instead of $Y$. Then the following result is a consequence of Theorem~\ref{thm:MC convergence Yt}.

\begin{cor}
\label{cor:convergence in filtering problem}
Assume that the training samples are generated as described above and that $Z$ satisfies Assumptions~\ref{assumption:4} and~\ref{assumption:5}.
Let $\theta^{\min}_{m,N} \in \Theta^{\min}_{m,N} := \arg \inf_{\theta \in \Theta_m}\{ \hat\Phi_N(\theta)\}$ for every $m, N \in \N$.
Then, one can define an increasing sequence $(N_m)_{m \in \N}$ in $\N$ such that for every $1 \leq k \leq K$ the following statements hold.
\begin{enumerate}
\item $G^{\theta_{m, N_m}^{\min}}$ converges to $\hat{Z}$
in the metric $d_k$ as $m \to \infty$.

\item $G_X^{\theta_{m, N_m}^{\min}} := \operatorname{proj}_X \left( G^{\theta_{m, N_m}^{\min}}  \right)$ converges to $\widehat{\varphi(X)} = \operatorname{proj}_X(\hat{Z})$
in the metric $\tilde{d}_k$ as $m \to \infty$.

\item
We write $G_X^{\theta_{m, N_m}^{\min}}(Z)$ and $G_X^{\theta_{m, N_m}^{\min}}(Y)$ to emphasise whether the $X$ and $Y$-coordinates or only the $Y$-coordinates are provided as input. Similarly we distinguish between the conditional expectation given the $X$ and $Y$-coordinates of the observations $\widehat{\varphi(X)} = \left( \E[\varphi(X)_t \, | \, \mathcal{A}_t] \right)_{t \in [0,T]} $ and the one given only the $Y$-coordinates $\left( \E[\varphi(X)_t \, | \,  \mathcal{Y}_t] \right)_{t \in [0,T]}$.
It holds that
\begin{equation*}
\tilde{d}_k \left(  G_X^{\theta_{m, N_m}^{\min}}(Z), G_X^{\theta_{m, N_m}^{\min}}(Y) \right) = \tilde{d}_k \left( \widehat{\varphi(X)} ,  \left( \E[\varphi(X)_t \, | \,  \mathcal{Y}_t] \right)_{t \in [0,T]} \right) = 0.
\end{equation*}
\end{enumerate}
\end{cor}

\begin{proof}
With the given assumptions on $Z$ and on the training samples, Assumptions~\ref{assumption:1} to~\ref{assumption:6} are satisfied (with the Case~2 of Remark~\ref{rem:extension assumptions M0}). Therefore, the first item follows directly from Theorem~\ref{thm:MC convergence Yt}.

For the second item, we first note that similar to Proposition~\ref{prop:mu_k expectation equiv}, for any c\`adl\`ag $\mathbb{A}$-adapted process $Z$, for which $Z^\star$ is integrable, we have
\begin{equation*}
\E_{\nu_k}[Z]  = c_4\,  \E_{\P \times \tilde{\P}}\left[ \1_{\{n \geq k\}} \1_{\{ \forall i < k: \operatorname{proj}_X(M_i)=0 \}} Z(\tilde{X}^{\leq t_{k-1}}, t_k-) \right].
\end{equation*}
Therefore (similar to \eqref{equ:convergence in L1}), we have
\begin{equation*}
\begin{split}
\tilde{d}_k \left( \widehat{\varphi(X)} , G_X^{\theta_m^{\star}}  \right)
	& \leq \tilde{d}_k \left(  \hat{Z} , G^{\theta_m^{\star}}  \right) \\
	& =  c_4 \, \E_{\P \times \tilde{\P}}\left[ \1_{\{n \geq k\}} \1_{\{ \forall i < k: \operatorname{proj}_X(M_i)=0 \}}   \left\lvert \hat{Z}_{t_k-} - G^{\theta_m^{\star}}_{t_k-} \right\rvert_2 \right] \\
	& \leq  c_4 \, \E_{\P \times \tilde\P}\left[ \1_{\{n \geq k\}}   \left\lvert \hat{Z}_{t_k-} - G^{\theta_m^{\star}}_{t_k-} \right\rvert_2 \right] \\
	& \leq  \frac{ c_4 \, c_1 \, c_3}{c_2} \, \left( \Phi(\theta_m^{\star}) - \Psi(\hat{Z}) \right)^{1/2}  \xrightarrow{m \to \infty} 0,
\end{split}
\end{equation*}
from which item 2 follows.

For the last item we have
\begin{align*}
&\tilde{d}_k \left(  G_X^{\theta_{m, N_m}^{\min}}(Z), G_X^{\theta_{m, N_m}^{\min}}(Y) \right) \\
&=  c_4 \,  \E_{\P\times\tilde\P }\left[ \1_{\{n \geq k\}}  \1_{ \{ \forall i < k: \operatorname{proj}_X(M_i) = 0  \} } \left| \left( G_X^{\theta_{m, N_m}^{\min}}(Z) \right)_{t_k-} - \left( G_X^{\theta_{m, N_m}^{\min}}(Y) \right)_{t_k-}  \right|_2 \right] = 0,
\end{align*}
since on the subset where  $\operatorname{proj}_X(M_i) = 0$ for $i < k$,  the PD-NJ-ODE model only gets the $Y$-coordinates as input up to time $t_k-$.
The second equality holds for the same reason.
\end{proof}

\begin{rem}
It follows from Corollary~\ref{cor:convergence in filtering problem} that $G_X^{\theta_{m, N_m}^{\min}}(Y)$ converges to $\left( \E[\varphi(X)_t \, | \,  \mathcal{Y}_t] \right)_{t \in [0,T]}$ in  the metric $\tilde{d}_k$ as $m \to \infty$.
In particular, the output\footnote{More precisely, the part of the output representing the approximations of the $X$-coordinates} of the \emph{trained} model, when applied to input samples of the filtering task, where only the $Y$-coordinates are observed, converges to the conditional expectation $\left( \E[\varphi(X)_t \, | \,  \mathcal{Y}_t] \right)_{t \in [0,T]} $.
\end{rem}

\section{Examples of processes satisfying the assumptions}\label{sec:Examples of Processes Satisfying the Assumptions}
We summarise several processes that satisfy Assumptions~\ref{assumption:4} and~\ref{assumption:5} (the other assumptions need to be satisfied by the observation framework, i.e., they are not process specific). The examples in Sections~\ref{sec:Ito Diffusion with Regularity Assumptions} and \ref{sec:Other Ito Diffusions: Black-Scholes, Ornstein-Uhlenbeck, Heston} are from \citet{herrera2021neural} and recalled here, to show that the new setting truly is a generalization of the old one.
In Section~\ref{sec:Stochastic Process with Jumps: Homogeneous Poisson Point Process}, we present a process with jumps; in Section~\ref{sec:FBM}, a path-dependent process; in Section~\ref{sec:Multivariate Process with Incomplete Observations: Correlated Brownian Motions}, a multivariate process with correlated coordinates and incomplete observations; and in Section~\ref{sec:Filtering Problem with Brownian Motions}, a process in the setting of a stochastic filtering problem.
Additional processes satisfying these assumptions are discussed in Appendix~\ref{sec:Additional Examples of Processes Satisfying the Assumptions}.

\subsection{It\^o diffusion with regularity assumptions}\label{sec:Ito Diffusion with Regularity Assumptions}
Let $\{ W_t \}_{t\in [0,T] }$ be a $d_W$-dimensional Brownian motion on $(\Omega, \F, \mathbb{F} := \{\F_t\}_{0 \leq t \leq T}, \P )$, for $d_W \in \N$. Let $X :={(X_t)}_{t \in [0,T]}$  be defined as the solution of the stochastic differential equation (SDE)
\begin{equation}\label{equ:SDE X}
dX_t = \mu(t, X_t) dt + \sigma(t, X_t) dW_t\,,
\end{equation}
for all $0 \leq t \leq T$, where $X_0 = x  \in \R^{d_X}$ is the starting point and the measurable functions $\mu : [0,T] \times \R^{d_X} \to \R^{d_X}$ and $\sigma: [0,T] \times \R^{d_X} \to \R^{d_X \times d_W}$ are the \emph{drift} and the \emph{diffusion} respectively. By definition this process is continuous. Moreover, we impose the following assumptions.
\begin{itemize}
	\item {$\mu$ and $\sigma$ are both globally
 Lipschitz continuous in their second component}, i.e., for $\varphi \in \{ \mu, \sigma\}$ there exists a constant $\tilde M > 0$ such that for all $t \in [0,T]$
\begin{equation*}\label{equ:condition mu sigma extended}
\lvert \varphi(t,x) - \varphi(t,y) \rvert_2 \leq \tilde M \lvert x -y \rvert_2 \quad \text{and} \quad \lvert \varphi(t, x) \rvert_2 \leq (1 + \lvert x \rvert_2) \tilde M.
\end{equation*}
In particular, their growth is at most linear in the second component.
	\item {$\mu$ is bounded, and continuous in its first component ($t$) uniformly in its second component ($x$)},  i.e., for every $t \in [0,T]$ and $\varepsilon>0$ there exists a $\delta>0$ such that for all $s \in [0,T]$ with $\lvert t-s\rvert < \delta$ and all $x \in \R^{d_X}$ we have $\lvert \mu(t,x) - \mu(s,x) \rvert < \epsilon$.
	\item {$\sigma$ is c\`adl\`ag} (right-continuous with existing left-limit) {in the first component ($t$) and $L^2$ integrable with respect to $W$, $\sigma \in L^2(W)$}, i.e.,
    \begin{equation}
    \label{equ:L2(W)}
    \E\left[\sum_{i=1}^{d_X} \sum_{j=1}^{d_W} \int_0^T \sup_x \sigma_{i,j}(t,x)^2 d[W^j,W^j]_t\right] = \int_0^T \lvert \sup_x \sigma(t,x) \rvert_F^2 \, dt < \infty,
    \end{equation}
    where $\lvert \cdot \rvert_F$ denotes the Frobenius matrix norm. This is in particular implied if $\sigma$ is bounded.
	\item We always observe all coordinates of $X$ simultaneously (no incomplete observations).
\end{itemize}

\begin{rem}
In \citet{herrera2021neural}, there was the additional assumption that $X$ is continuous and square-integrable; we omit this assumption here because it is already implied by the others.
\end{rem}

Under these assumptions a unique continuous solution of \eqref{equ:SDE X} exists, once an initial value is fixed \citep[Thm. 7, Chap. V]{Pro1992}.
\citet[Lemma E.7]{herrera2021neural} shows that $X^\star$ is $L^2$-integrable.
Moreover, the results of \citet[Propositions B.1 and B.4]{herrera2021neural} imply that Assumption~\ref{assumption:4} is satisfied.

\begin{rem}\label{rem:generalization ito diff}
The assumption that $\mu$ is bounded and the integrability assumption on $\sigma$ can be weakened, as will be shown in Section~\ref{sec:Stochastic Functional Differential Equations}. In particular, the boundedness of $\mu$ is only needed to apply the Markov property, which is not needed in our more general setting now, where we have access to the entire past information.
\end{rem}

\subsection{Other It\^o diffusions: black-scholes, Ornstein-Uhlenbeck, Heston}\label{sec:Other Ito Diffusions: Black-Scholes, Ornstein-Uhlenbeck, Heston}
The Black-Scholes (geometric Brownian Motion), Ornstein-Uhlenbeck and Heston processes are It\^o diffusions which do not satisfy the assumptions in Section~\ref{sec:Ito Diffusion with Regularity Assumptions}. In particular, their drifts are not bounded and the Heston process does not satisfy the Lipschitz assumptions. If the Feller condition is satisfied, the Heston process is Lipschitz with high probability, otherwise not.
Nevertheless, Assumptions~\ref{assumption:4} and~\ref{assumption:5} are still satisfied for these three processes as shown below.
Experiments on these processes were performed in \citet{herrera2021neural}.

\begin{example}[Black-Scholes]
The SDE describing this model is
\begin{equation*}
dX_t = \mu X_t dt+ \sigma X_t dW_t , \quad X_0 = x_0,
\end{equation*}
where $W$ is a 1-dimensional Brownian motion and $\mu, \sigma \geq 0$. The conditional expectation of the solution process $X$ is given by $\hat X_t = E(X_{t}|X_{\tau(t)}) = X_{\tau(t)} e^{\mu (t - \tau(t))}$. Hence, $f(s, \tau(t), \tilde X^{\leq \tau(t)} ) = X_{\tau(t)} \mu e^{\mu(s - \tau(t))}$.

\citet[Lemma 16.1.4]{cohen2015stochastic} implies that $\E[(X^\star_T)^p] < \infty$ for every $2 \leq p <  \infty$ and therefore by HÃ¶lder's inequality for all $1 \leq p < \infty$.

\end{example}

\begin{example}[Ornstein-Uhlenbeck]
This model is described by the SDE
\begin{equation*}
dX_t = -k(X_t - m) dt+ \sigma dW_t , \quad  X_0 = x_0,
\end{equation*}
where $W$ is a 1-dimensional Brownian motion and $ k, m, \sigma > 0$. The conditional expectation of the solution process $X$ is given by $\hat X_t = E(X_{t}|X_{\tau(t)}) = X_{\tau(t)} e^{-k(t - \tau(t))} + m\left(1-e^{-k (t - \tau(t))}\right)$. Hence, $f(s, \tau(t), \tilde X^{\leq \tau(t)} ) = -X_{\tau(t)} k e^{-k(s - \tau(t))} + m k e^{-k(s - \tau(t))}$.

Again, \citet[Lemma 16.1.4]{cohen2015stochastic} implies that $\E[(X^\star_T)^p] < \infty$ for every $2 \leq p <  \infty$ and therefore by HÃ¶lder's inequality for all $1 \leq p < \infty$.

\end{example}

\begin{example}[Heston]
The Heston model is described by the SDE
\begin{equation*}
\begin{split}
dX_t &= \mu X_t dt+ \sqrt{v_t}X_tdW_t \\
dv_t &= -k(v_t - m) dt+ \sigma \sqrt{v_t} dB_t
\end{split}
\end{equation*}
where $W$ and $B$ are 1-dimensional Brownian motions with correlation $\rho \in (-1,1)$, $\mu \geq 0$ and $k, m, \sigma > 0$. The conditional expectation of the solution process $X$ is given by $\hat X_t = E(X_{t}|X_{\tau(t)}) = X_{\tau(t)} e^{\mu (t - \tau(t))}$. Hence, $f_1(s, \tau(t), \tilde X^{\leq \tau(t)} ) = X_{\tau(t)} \mu e^{\mu(s - \tau(t))}$.
Moreover, if also the stochastic variance should be predicted simultaneously, its conditional expectation is given by $\hat v_t = E(v_{t}|v_{\tau(t)}) = v_{\tau(t)} e^{-k(t - \tau(t))} + m\left(1-e^{-k (t - \tau(t))}\right)$. Hence, $f_2(s, \tau(t), \tilde v^{\leq \tau(t)} ) = -v_{\tau(t)} k e^{-k(s - \tau(t))} + m k e^{-k(s - \tau(t))}$.

$L^p$-integrability of the Heston model is more delicate.
While \citet[Lemma 16.1.4]{cohen2015stochastic} implies that $\E[(v^\star_T)^p] < \infty$ for every $2 \leq p <  \infty$ and therefore by HÃ¶lder's inequality for all $1 \leq p < \infty$, the moments of $X$ can explode \citep[Proposition 3.1]{andersen2007moment}.
In particular, the moment explosion time
\[T_\star(p) := \sup\{ t \geq 0 \, | \, \E[|X_t|^p] < \infty \}\]
of $X$ is infinite if and only if $\rho  \sigma p < k $ and  $(\rho  \sigma p - k)^2 - \sigma^2 (p^2 - p) \geq 0$.
By It\^o's lemma we can write $X_t = \exp(\mu t + Z_t) $, where $Z_t$ is the discounted log-price process satisfying
\begin{equation*}
d Z_t = -\frac{v_t}{2} dt + \sqrt{v_t} dW_t.
\end{equation*}
According to \citet[Corollary 2.7 and Equation 6.1]{keller2011moment}, $\exp(Z_t)$ is a martingale, hence Doob's Maximal inequality implies that for any $1 < p < \infty$
\begin{equation*}
\E[(X^\star_T)^p] \leq \frac{p}{p-1} \exp(\mu T) \E[|X_T|^p].
\end{equation*}
Therefore, $X^\star_T$ is $L^p$-integrable if $T < T_\star(p)$, where a closed form for $T_\star(p)$ is given in \citet[Proposition 3.1]{andersen2007moment}, in case it is finite.
Again, Hölder's inequality implies that $X^\star_T$ is $L^1$-integrable if it is $L^p$-integrable for some $p>1$.

\end{example}

\begin{rem}
$T_\star(p) = \infty$ if
\[p < \min \left( \frac{k}{\rho \sigma}, \frac{ (\sigma^2 - 2 k \sigma \rho) + \sqrt{ (\sigma^2 - 2k\sigma\rho)^2 + 4 \sigma^2 (1 - \rho^2) k^2 }}{2 \sigma^2 (1-\rho^2)}  \right).\]
Note that $(1-\rho^2)> 0 $, therefore the second term is an element of $\R_+$.\\
In the standard Heston example of \citet{herrera2021neural}, the used parameters are $\mu =2$, $\sigma=0.3$,  $k=2$, $m=4$, $\rho=0.5$ and therefore $T_\star(p) = \infty$ for all $p \leq 4.7972$.
In the second example of \citet{herrera2021neural} of a Heston model where the Feller condition is not satisfied, the used parameters are
$\mu =2$, $\sigma=3$,  $k=2$, $m=1$,  $\rho= 0.5$ and therefore $T_\star(p) = \infty$ for all $p \leq 1.0234\ldots $. Moreover, for $p=2$ we have $T_\star(2) > 0.6465$.
\end{rem}

\subsection{Stochastic process with jumps: homogeneous poisson point process}
\label{sec:Stochastic Process with Jumps: Homogeneous Poisson Point Process}
A homogeneous Poisson point process $(N^\lambda_t)_{t \geq 0}$, for $\lambda > 0$,
is defined to be the increasing stochastic process in $\R_{\geq 0}$ such that $N^\lambda_0 = 0$ and its increments are independent  Poisson distributed, i.e., $N^\lambda_t - N^\lambda_s \sim \operatorname{Poisson}(\lambda (t-s))$ for any $0 \leq s \leq t$.
It follows that its conditional expectation is $\hat N^\lambda_t = E(N^\lambda_{t} | N^\lambda_{\tau(t)}) = E( N^\lambda_{\tau(t)} + (N^\lambda_{t} - N^\lambda_{\tau(t)}) | N^\lambda_{\tau(t)}) = N^\lambda_{\tau(t)} + \lambda (t - \tau(t))$. Therefore, $f(s, \tau(t), \tilde N^{\lambda, \leq \tau(t)} )  = \lambda$.

By its definition as an increasing process, $(N^\lambda)_T^\star = N^\lambda_T $, and since $N^\lambda_T \sim \operatorname{Poisson}(\lambda (T))$ all moments exist, hence $\E[((N^\lambda)^\star_T)^p] < \infty$ for all $1 \leq p < \infty$.

\subsection{Fractional brownian motion}\label{sec:FBM}
\begin{definition}    A random vector $\left(X_1,\dots,X_n\right)\in\R^n$ is a centered Gaussian vector if any linear combination of the $X_i$ is a centered Gaussian random variable.

\end{definition}

\begin{definition}    A stochastic process $(X_t)_{t\in I}$ is a centered Gaussian process if any finite-dimensional vector $\left(X_{t_1},\dots,X_{t_p}\right)$ is a centered Gaussian vector.

\end{definition}

Since the law of a centered Gaussian vector is completely determined by its covariance function, it suffices to specify the covariance to obtain a unique Gaussian process.
In fact, one can prove that for each function $S:I\times I\rightarrow\R$ such that for all $t_1,\dots,t_p$ in $I$ and for all $\lambda_1,\dots,\lambda_p$ in $\R$,
\begin{equation*}
    \sum_{1\leq i,j\leq p} \lambda_i\lambda_j S(t_i,t_j) \geq 0,
\end{equation*}
we can construct a centered Gaussian process $(X_t)_{t\in I}$ with the given covariance function $S$, using Kolmogorov's extension theorem.

\begin{definition}    The fractional Brownian motion $B^H = (B_t^H)_{t\in\R_+}$ with Hurst index $H\in(0,1]$ is the centered Gaussian process with covariance function
    \begin{equation*}
        r_H(t,s) = \frac{1}{2} \left[t^{2H}+s^{2H} - |t-s|^{2H}\right].
    \end{equation*}

\end{definition}

Note that the case $H=\frac{1}{2}$ results in a covariance of $\frac{1}{2}\left[t+s-|t-s|\right] = \text{min}(s,t)$, which is precisely the covariance of a standard Brownian motion.
The case $H>\frac{1}{2}$ corresponds to positively correlated increments, which is a common phenomenon in financial mathematics.
Meanwhile, the case $H<\frac{1}{2}$ corresponds to negatively correlated increments.

If $H\ne \frac{1}{2}$,  the paths are non-Markovian, hence, the conditional expectation does not only include the information of the last measurement, as we have seen for the Black-Scholes, Heston and Ornstein-Uhlenbeck model.
Instead, the conditional process (of which the mean is the conditional expectation) of the fractional Brownian motion depends on the entire past trajectory.
In \citet[page~3]{FBM_cond_exp}, the law of this conditional process is described explicitly.
It requires knowledge of the whole trajectory of the fractional Brownian motion up to the current time.
Since, in our setting, we only have a finite number of discrete observations (at random times) available to train our model, it is unreasonable to expect the model to correctly approximate the conditional expectation implied by this conditional process.
However, we can also compute the true conditional expectation given our discrete observations, making use of the fact that $B^H$ is a Gaussian process, implying that the vector of observations is a Gaussian vector \citep{bj1171899318}.
Assume we have observed the values $(B^H_{t_1}, \dotsc, B^H_{t_\nu} )$ of $B^H$ at times $0 < t_1 < \dotsc < t_\nu <T$ and we want to compute the conditional expectation $\E\left[B^H_{t} \,|\, (B^H_{t_1}, \dotsc, B^H_{t_\nu} ) \right]$ for $t \geq t_\nu$.
For any fixed $t$, the vector  $(B^H_{t_1}, \dotsc, B^H_{t_\nu},  B^H_{t} )$ is centred Gaussian with covariance
\begin{equation*}
R = \begin{pmatrix}
r_H(t_1, t_1) &  \dots & r_H(t_1, t) \\
\vdots & \ddots & \vdots \\
r_H(t_1, t) & \dots & r_H(t, t)
\end{pmatrix}
=: \begin{pmatrix}
R_\nu & R_{\nu,1} \\
R_{1, \nu} & r_H(t, t)
\end{pmatrix}.
\end{equation*}
Then a simple calculation shows that
\begin{equation*}
\operatorname{cov}\left(B^H_{t} - R_{1,\nu} R_\nu^{-1} (B^H_{t_1}, \dotsc, B^H_{t_\nu}  )^\top, (B^H_{t_1}, \dotsc, B^H_{t_\nu}) \right) =
R_{1,\nu} - R_{1,\nu} R_\nu^{-1} R_\nu = 0,
\end{equation*}
implying that $ B^H_{t}  - R_{1,\nu} R_\nu^{-1} (B^H_{t_1}, \dotsc, B^H_{t_\nu})^\top$ is independent of $(B^H_{t_1}, \dotsc, B^H_{t_\nu})$, since it is a Gaussian vector.
Hence,
\begin{equation*}
\begin{split}
\E\left[B^H_{t} \,|\, (B^H_{t_1}, \dotsc, B^H_{t_\nu} ) \right] &= \E\left[ B^H_{t}  - R_{1,\nu} R_\nu^{-1} (B^H_{t_1}, \dotsc, B^H_{t_\nu})^\top  \right] +  R_{1,\nu} R_\nu^{-1} (B^H_{t_1}, \dotsc, B^H_{t_\nu})^\top \\
&= R_{1,\nu} R_\nu^{-1} (B^H_{t_1}, \dotsc, B^H_{t_\nu} )^\top \\
& = (r_H(t_1, t) , \dotsc, r_H(t_\nu, t)) R_\nu^{-1} (B^H_{t_1}, \dotsc, B^H_{t_\nu} )^\top \\
&= \frac{1}{2} (t^{2H}+t_1^{2H} - (t-t_1)^{2H} , \dotsc, t^{2H}+t_\nu^{2H} - (t-t_\nu)^{2H}) R_\nu^{-1} \\
& \quad \cdot (B^H_{t_1}, \dotsc, B^H_{t_\nu} )^\top.
\end{split}
\end{equation*}
This function is differentiable in $t$ and we get
\begin{equation*}
f(s, \tau(t), \tilde B^{H, \leq \tau(t)} )  = H (s^{2H-1} - (s-t_1)^{2H-1} , \dotsc, s^{2H-1}- (s-t_\nu)^{2H-1}) R_\nu^{-1}  (B^H_{t_1}, \dotsc, B^H_{t_\nu} )^\top,
\end{equation*}
where $\tau(t) = t_\nu$  under the assumptions above.

The following result, which is a consequence of the upper bounds on Pickands constant \citep{10.2307/24306010, DEBICKI20082046}, shows that the running maximum process of fractional Brownian motion is  $L^p$-integrable for all $1 \leq p < \infty$.

\begin{prop}
Let $H \in (0,1]$ and $B^H$ a fractional Brownian motion with Hurst parameter $H$. Let $X_T^\star := (B^H_T)^\star = \sup_{0 \leq t \leq T} |B^H_t|$ and $1 \leq p < \infty$. Then
$\E[ (X^\star_T)^p] < \infty$.
\end{prop}

\begin{proof}
Pickands constant is defined as
\begin{equation*}
\mathcal{H}_H := \lim_{t \to \infty} \tfrac{1}{t} \mathcal{H}_H(t),
\quad  \text{where} \quad
\mathcal{H}_H(t) := \E \left[ \exp \left( \sup_{0 \leq s \leq t} \left(\sqrt{2} B^H_s - s^{2H} \right) \right) \right].
\end{equation*}
According to \citet{DEBICKI20082046}, $\mathcal{H}_H$ is finitely bounded from above for every $H \in (0,1]$, therefore, there exists some $T_0$ such that for all $t \geq T_0$, $\mathcal{H}_H(t)$ is finite and bounded from above by some constant. If $T < T_0$, we can use that $\E[ (X^\star_T)^p] \leq \E[ (X^\star_{T_0})^p]$, hence we can assume without loss of generality that $T \geq T_0$.
We have
\begin{equation*}
\mathcal{H}_H(T) \geq \E \left[ \exp \left( \sup_{0 \leq t \leq T} \left(\sqrt{2} B^H_t - T^{2H} \right) \right) \right] = \E \left[ \exp \left( \sqrt{2} \sup_{0 \leq t \leq T}B^H_t \right) \right] \exp(- t^{2H})
\end{equation*}
and since $B^H$ is symmetric around $0$, implying that $-B^H \overset{d}{=}B^H$,
\begin{equation*}
\E \left[ \exp \left( \sqrt{2} X^\star_T \right) \right] = \E \left[ \exp \left( \sqrt{2} \sup_{0 \leq t \leq T} | B^H_t | \right) \right]  \leq 2 \, \E  \left[ \exp \left( \sqrt{2} \sup_{0 \leq t \leq T}  B^H_t  \right) \right] + 2.
\end{equation*}
Together, this implies that $\E \left[ \exp \left( \sqrt{2} X^\star_T \right) \right] < \infty$., which means that the moment generating function of $X^\star$ is finite. A well known  property of the moment generating function  implies that
\begin{equation*}
\E[ (X^\star_T)^p] \leq \left(\frac{p}{\sqrt{2} e }\right)^p \E \left[ \exp \left( \sqrt{2} X^\star_T \right) \right] < \infty,
\end{equation*}
since $X^\star_T$ is non-negative.
\end{proof}

\subsection{Multivariate process with incomplete observations: correlated brownian motions}
\label{sec:Multivariate Process with Incomplete Observations: Correlated Brownian Motions}
Let $A, B, C$ be three i.i.d. 1-dimensional Brownian motions and let $\alpha, \beta > 0$ with $\alpha^2+\beta^2=1$. Then we define the process $X:=(U, V)^\top := (\alpha A + \beta B, \alpha A + \beta C)^\top$. $U, V$ are again Brownian motions, but they are correlated.
First we remark that $\E[(X^\star_T)^p] < \infty$ for every $1 \leq p <  \infty$ (cf.\ \citet[Lemma 16.1.4]{cohen2015stochastic}).

In this example we assume to have incomplete observations. In particular, we allow for observation times at which only $U$ or only $V$ is observed. Therefore, let us define the (ordered) observations of $U$ as $r_i$ for $1 \leq i \leq n_U$ and those of $V$ as $s_i$ for $1 \leq i \leq n_V$, where $n_U, n_V$ are random variables similar to $n$,  representing the total number of observations of $U$ and $V$ respectively. As before, $t_i$ for $1 \leq i \leq n$ are the times at which observations are made, i.e., the times at which at least one of $U, V$ is observed.
Because the coordinates are correlated, observing only one coordinate generally also affects the conditional expectation of the other. To compute the true conditional expectations given the previous discrete (and possibly incomplete) observations, we use the fact that the increments of $A, B, C$ are independent Gaussian random variables.
Remember that $t_0 = 0$ and let $\tilde{A_i} := A_{t_i} - A_{t_{i-1}}$ for all $1 \leq i \leq n$ and similar for $B,C$. Then we define
\[v := (\tilde{A}_1, \dotsc, \tilde{A}_n, \tilde{B}_1, \dots, \tilde{B}_n, \tilde{C}_1, \dots, \tilde{C}_n)^\top,\]
which is multivariate normally distributed with $v \sim N(0, \Sigma)$, where
\[\Sigma := \operatorname{diag} \left(t_1 - t_0, \dots, t_n - t_{n-1}, t_1 - t_0, \dots, t_n - t_{n-1}, t_1 - t_0, \dots, t_n - t_{n-1} \right).\]
Let $I_l \in \R^{n \times n}$ be the lower triangular matrix with all ones  and define the matrix
\begin{equation*}
\tilde\Gamma := \begin{pmatrix}
\alpha I_l & \beta I_l & 0 \\
\alpha I_l & 0 & \beta I_l  \\
\end{pmatrix}
\in \R^{2n \times 3 n}.
\end{equation*}
Then $(U_{t_1}, \dotsc, U_{t_n}, V_{t_1}, \dots, V_{t_n})^\top = \tilde \Gamma v$.
Let $t > 0$ be the current time and let $k \in \N$ be such that $t_k \leq t < t_{k+1}$, where $t_{n+1}:=\infty$.
Moreover, let $k_U, k_V \in \N$ be such that $r_{k_U} \leq t_k < r_{k_U +1}$ and $s_{k_V} \leq t_k < s_{k_V +1}$.
In the following we compute conditional expectations of $V$. Those of $U$ follow equivalently.
First we notice that the independent increments of Brownian motion imply
\begin{equation*}
\E[V_t | \mathcal{A}_t] = \E[V_t - V_{t_k} | \mathcal{A}_{t_k} ] + \E[ V_{t_k} | \mathcal{A}_{t_k} ] = \E[ V_{t_k} | \mathcal{A}_{t_k} ] =  \E[ V_{t_k} | U_{r_1}, \dotsc, U_{r_{k_U}}, V_{s_1}, \dots, V_{s_{k_V}} ]
\end{equation*}
and therefore, $f(s, \tau(t), \tilde X^{\leq \tau(t)} ) = 0$.
If $s_{k_V} = t_k$, i.e., if $V_{t_k}$ was observed, then $ \E[ V_{t_k} | \mathcal{A}_{t_k} ] = V_{t_k}$.
Otherwise,  $s_{k_V} < t_k$ and we define
\begin{equation*}
\Upsilon := \operatorname{diag}\left( \1_{\{ t_1 \in \mathcal{R} \} }, \dots,  \1_{\{ t_{n} \in \mathcal{R}  \} }, \1_{\{ t_1 \in \mathcal{S}  \} }, \dots,  \1_{\{ t_{n} \in \mathcal{S} \} }  \right),
\end{equation*}
where $\mathcal{R}:=\{ r_1, \dotsc, r_{k_U} \}$ and $\mathcal{S} := \{ s_1, \dotsc, s_{k_V}\} \cup \{ t_k \}$, as well as $\tilde \Upsilon$ as the submatrix of $\Upsilon$ with all rows with only $0$-entries removed.
For $\Gamma := \tilde \Upsilon \tilde \Gamma$ we therefore have
\begin{equation*}
(U_{r_1}, \dotsc, U_{r_{k_U}}, V_{s_1}, \dots, V_{s_{k_V}}, V_{t_k})^\top = \Gamma v \sim N(0, \Gamma \Sigma \Gamma^\top),
\end{equation*}
where we used a well known fact about affine transformations of multivariate normal distributions (see e.g.\ \citet[Chapter~3.1]{Eaton2007Multi}).
Let
\[ \tilde \Sigma := \Gamma \Sigma \Gamma^\top =: \begin{pmatrix}
\tilde \Sigma_{11} & \tilde \Sigma_{12} \\
\tilde \Sigma_{21} & \tilde \Sigma_{22}
\end{pmatrix},\]
where $\tilde \Sigma_{11} \in \R^{(k_U + k_V) \times (k_U + k_V) }$, $\tilde \Sigma_{12} = \tilde \Sigma_{21}^\top \in \R^{(k_U + k_V) \times 1}$ and   $\tilde \Sigma_{22} = \operatorname{Var}(V_{t_k}) \in \R$.
Then, the conditional distribution of $(V_{t_k} \, | \, U_{r_1}, \dotsc, U_{r_{k_U}}, V_{s_1}, \dots, V_{s_{k_V}})$ is again normal with mean $\hat \mu := \tilde \Sigma_{21} \tilde \Sigma_{11}^{-1} (U_{r_1}, \dotsc, U_{r_{k_U}}, V_{s_1}, \dots, V_{s_{k_V}} )^\top $ and variance $\hat \Sigma :=  \tilde \Sigma_{22} -  \tilde \Sigma_{21} \tilde \Sigma_{11}^{-1} \tilde \Sigma_{12}$ \citep[Proposition~3.13]{Eaton2007Multi}.
In particular, we have  $\E[ V_{t_k} | \mathcal{A}_{t_k} ] = \hat \mu$.

\subsection{Filtering problem with brownian motions}\label{sec:Filtering Problem with Brownian Motions}
Let $X, W$ be two i.i.d. 1-dimensional Brownian motions and let $\alpha \in \R$ with which we define $h: \R \to \R, x \mapsto \alpha x$.
We use  $X$ as  the (unobserved) signal process and define the observation process $Y := h(X) + W$  (although this differs slightly from the description in Section~\ref{sec:Stochastic Filtering Problem}, it is justified by Remark~\ref{rem:more general filtering formulations}). As described in Section~\ref{sec:Stochastic Filtering with PD-NJ-ODE}, we assume that $h$ and the distribution of $X$ and $W$ are known to generate training data from it. Moreover, evaluation samples only have observations of $Y$.

$Z = (X,Y)$ is a Gaussian process with $\E[(Z^\star_T)^p] < \infty$ for every $1 \leq p <  \infty$ (cf.\ \citet[Lemma 16.1.4]{cohen2015stochastic}) and we can derive the analytic expressions of the conditional expectations similarly as in Section~\ref{sec:Multivariate Process with Incomplete Observations: Correlated Brownian Motions}.
Let $t_i$ be the observation times and define $\tilde{X}_i := X_{t_i} - X_{t_{i-1}}$ for $1 \leq i \leq n$ and similarly for $W$. Moreover, (for the samples of the training set) let us define the (ordered) observations of $X$ as $r_i$ for $1 \leq i \leq n_X$.
Then
\[v := ( \tilde{X}_1, \dotsc, \tilde{X}_n, \tilde{W}_1, \dotsc, \tilde{W}_n)^\top\]
is multivariate normally distributed with $v \sim N(0, \Sigma)$, where
\[\Sigma := \operatorname{diag} \left( t_1 - t_0, \dots, t_n - t_{n-1}, t_1 - t_0, \dots, t_n - t_{n-1} \right).\]
Let $I_l \in \R^{n \times n}$ be the lower triangular matrix with all ones  and define the matrix
\begin{equation*}
\tilde\Gamma := \begin{pmatrix}
\alpha I_l & I_l  \\
 I_l & 0 \\
\end{pmatrix}
\in \R^{2n \times 2n}.
\end{equation*}
Then $( Y_{t_1}, \dots, Y_{t_n}, X_{t_1}, \dotsc, X_{t_n} )^\top = \tilde \Gamma v$.
Let $t > 0$ be the current time and let $k \in \N$ be such that $t_k \leq t < t_{k+1}$, where $t_{n+1}:=\infty$.
Moreover, let $k_X \in \N$ be such that $r_{k_X} \leq t_k < r_{k_X +1}$.
In the following, we compute the conditional expectation of $X$, since that of $Y$ is trivially given by the last observation of $Y$, as $Y$ is observed at every observation time (cf. Section~\ref{sec:Multivariate Process with Incomplete Observations: Correlated Brownian Motions}).
The independent increments of Brownian motion imply
\begin{equation*}
\E[X_t | \mathcal{A}_t] = \E[X_t - X_{t_k} | \mathcal{A}_{t_k} ] + \E[ X_{t_k} | \mathcal{A}_{t_k} ] = \E[ X_{t_k} | \mathcal{A}_{t_k} ] =  \E[ X_{t_k} | X_{r_1}, \dotsc, X_{r_{k_X}}, Y_{t_1}, \dots, Y_{t_{k}} ]
\end{equation*}
and therefore, $f(s, \tau(t), \tilde X^{\leq \tau(t)} ) = 0$.
If $r_{k_X} = t_k$, i.e., if $X_{t_k}$ was observed, then $ \E[ X_{t_k} | \mathcal{A}_{t_k} ] = X_{t_k}$.
Otherwise,  $r_{k_X} < t_k$ and we define
\begin{equation*}
\Upsilon := \operatorname{diag}\left( \1_{\{ t_1 \in \mathcal{S}  \} }, \dots,  \1_{\{ t_n \in \mathcal{S}  \} }, \1_{\{ t_1 \in \mathcal{R}  \} }, \dots,  \1_{\{ t_{n} \in \mathcal{R} \} }  \right),
\end{equation*}
where $\mathcal{R}:=\{ r_1, \dotsc, r_{k_X} \} \cup \{ t_k \}$ and  $\mathcal{S}:=\{ t_1, \dotsc, t_{k} \}$, as well as $\tilde \Upsilon$ as the submatrix of $\Upsilon$ with all rows with only $0$-entries removed.
For $\Gamma := \tilde \Upsilon \tilde \Gamma$ we therefore have
\begin{equation*}
(Y_{t_1}, \dotsc, Y_{t_{k}}, X_{r_1}, \dots, X_{r_{k_X}}, X_{t_k} )^\top = \Gamma v \sim N(0, \Gamma \Sigma \Gamma^\top),
\end{equation*}
where we used a well known fact about affine transformations of multivariate normal distributions (see e.g.\ \citet[Chapter~3.1]{Eaton2007Multi}).
Let
\[ \tilde \Sigma := \Gamma \Sigma \Gamma^\top =: \begin{pmatrix}
\tilde \Sigma_{11} & \tilde \Sigma_{12} \\
\tilde \Sigma_{21} & \tilde \Sigma_{22}
\end{pmatrix},\]
where $\tilde \Sigma_{11} \in \R^{(k + k_X) \times (k + k_X) }$, $\tilde \Sigma_{12} = \tilde \Sigma_{21}^\top \in \R^{(k + k_X) \times 1}$ and   $\tilde \Sigma_{22} = \operatorname{Var}(X_{t_k}) \in \R$.
Then, the conditional distribution of $(X_{t_k} \, | \, Y_{t_1}, \dotsc, Y_{t_{k}}, X_{r_1}, \dots, X_{r_{k_X}})$ is again normal with mean $\hat \mu := \tilde \Sigma_{21} \tilde \Sigma_{11}^{-1} (Y_{t_1}, \dotsc, Y_{t_{k}}, X_{r_1}, \dots, X_{t_{k_X}} )^\top $ and variance $\hat \Sigma :=  \tilde \Sigma_{22} -  \tilde \Sigma_{21} \tilde \Sigma_{11}^{-1} \tilde \Sigma_{12}$ \citep[Proposition~3.13]{Eaton2007Multi}.
In particular, we have  $\E[ X_{t_k} | \mathcal{A}_{t_k} ] = \hat \mu$.

\section{Experiments}\label{sec:Experiments}
The code with all new experiments and those from  \citet{herrera2021neural} is available at \url{https://github.com/FlorianKrach/PD-NJODE}.
Additional experiments are provided in Appendix~\ref{sec:Additional Experiments}.
Moreover, further details about the experiments can be found in Appendix~\ref{sec:Experimental Details}. Since our practical implementation slightly deviates from the theoretical description, we list all differences in Appendix~\ref{sec:Differences between the Implementation and the Theoretical Description of the PD-NJ-ODE}.

First, we use synthetic datasets to confirm our theoretical results for all generalizations of the problem settings considered in \citet{herrera2021neural}.
In particular, we show that the path-dependent NJ-ODE can be applied successfully when the underlying process has jumps (Section~\ref{sec:Process with Jumps - Poisson Point Process}) or is path-dependent (Section~\ref{sec:Path Dependence -- Fractional Brownian Motion}), and when observations are incomplete (Section~\ref{sec:Incomplete Observations -- Correlated 2-Dimension Brownian Motion}).
Moreover, we verify that the (path-dependent) NJ-ODE can be used for uncertainty estimation (Section~\ref{sec:Uncertainty estimation - Brownian motion and its Square}) and for solving stochastic filtering problems (Section~\ref{sec:Stochastic Filtering of a Brownian Motion with Brownian Noise}) as was suggested in Sections~\ref{sec:Conditional Variance, Moments and Moment Generating Function} and~\ref{sec:Stochastic Filtering with PD-NJ-ODE}, respectively.
Model performance is measured using the \emph{evaluation metric} introduced in \citet[Sec. 6.1]{herrera2021neural}. This metric computes the MSE, on a discretization time grid, between the optimal prediction (given by the conditional expectation, which can be computed in these synthetic examples because the law of the underlying process is known) and the model predictions on the test samples.
It is given by
\begin{equation*}\label{equ:evaluation metric}
\operatorname{eval}(\hat X, Y^{\theta}) := \frac{1}{N_2} \sum_{j=1}^{N_2}  \frac{1}{\kappa+1} \sum_{i=0}^{\kappa} \left(  \hat X_{\frac{i T}{\kappa}}^{(j)} - Y_{\frac{i T}{\kappa}}^{\theta, j } \right)^2,
\end{equation*}
where the outer sum runs over the test set of size $N_2$ and the inner sum runs over the equidistant grid points on the time interval $[0,T]$.

In Sections~\ref{sec:Process with Jumps - Poisson Point Process} and \ref{sec:Uncertainty estimation - Brownian motion and its Square} we show results using the standard NJ-ODE model, since the respective datasets do not exhibit any of the complications that would (numerically) require  the PD-NJ-ODE. However, we remark that PD-NJ-ODE  leads (at least) to the same quality of results.

Our empirical results on more complicated datasets suggest that using the equivalent objective function as described in Remark~\ref{rem:equivalent objective function} is preferable, since the training is more stable.
With the original loss function, the training more easily ends up in local optima which are not global optima. However, these local optima do not exist for the equivalent loss function due to its slightly different structure, hence the training does not end up there.
For more details see Appendix~\ref{sec:Loss Function details}.

Finally, in Sections~\ref{ref:Physionet -- Real World Dataset} and {\ref{sec:Limit Order Book Data}}, we apply the PD-NJ-ODE model to real-world datasets.
First, we apply it to the PhysioNet dataset of patient health parameters and compare the results of PD-NJ-ODE with those of NJ-ODE.
Second, we apply it to three limit-order-book (LOB) datasets, adapting the PD-NJ-ODE architecture to address an additional classification task.

\subsection{Process with jumps-poisson point process}\label{sec:Process with Jumps - Poisson Point Process}
\begin{figure}[htp!]
\centering
\includegraphics[width=0.6\textwidth]{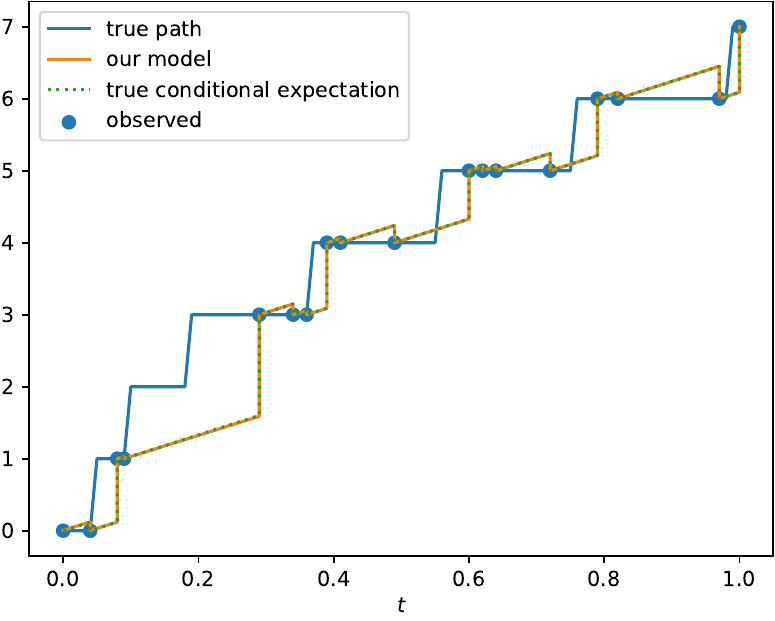}
\caption{Predicted and true conditional expectation on a test sample of the Poisson point process. All upward movements of the true path are jumps, the slope is only due to the discretization time grid.}
\label{fig:PPP}
\end{figure}

In this experiment, we use a Poisson point process, as described in Section~\ref{sec:Stochastic Process with Jumps: Homogeneous Poisson Point Process}, to show that NJ-ODE can be applied successfully to processes with jumps.
The standard NJ-ODE model without extensions already achieves an evaluation metric smaller than $2\cdot 10^{-5}$ after only 30 training epochs (Fig.~\ref{fig:PPP}). Hence, we conclude that this is another problem setting in which no extension of the NJ-ODE model is needed, although the setting is not covered by the theoretical framework in \citet{herrera2021neural} (as is also the case for Black--Scholes, Ornstein--Uhlenbeck, and Heston processes).

\subsection{Uncertainty estimation -- Brownian motion and its square}\label{sec:Uncertainty estimation - Brownian motion and its Square}
\begin{figure}[htp!]
\centering
\includegraphics[width=0.49\textwidth]{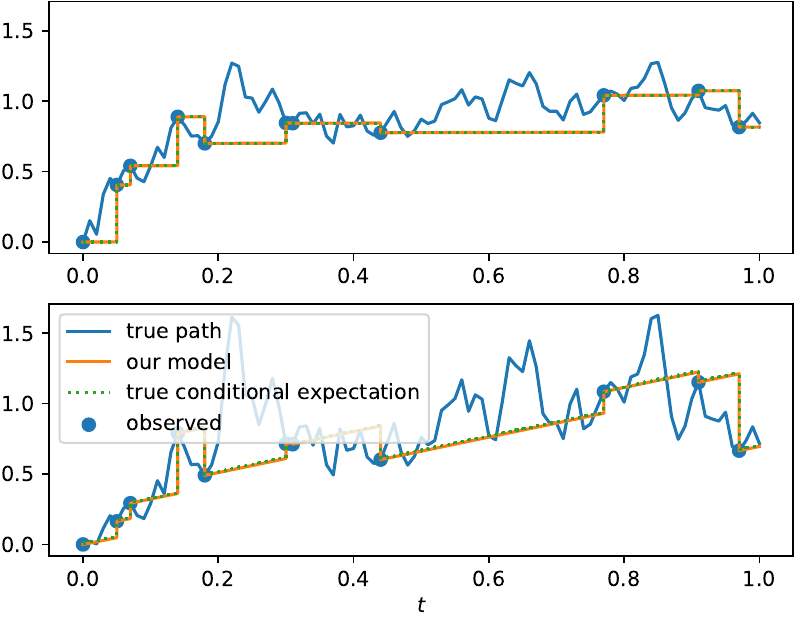}
\includegraphics[width=0.49\textwidth]{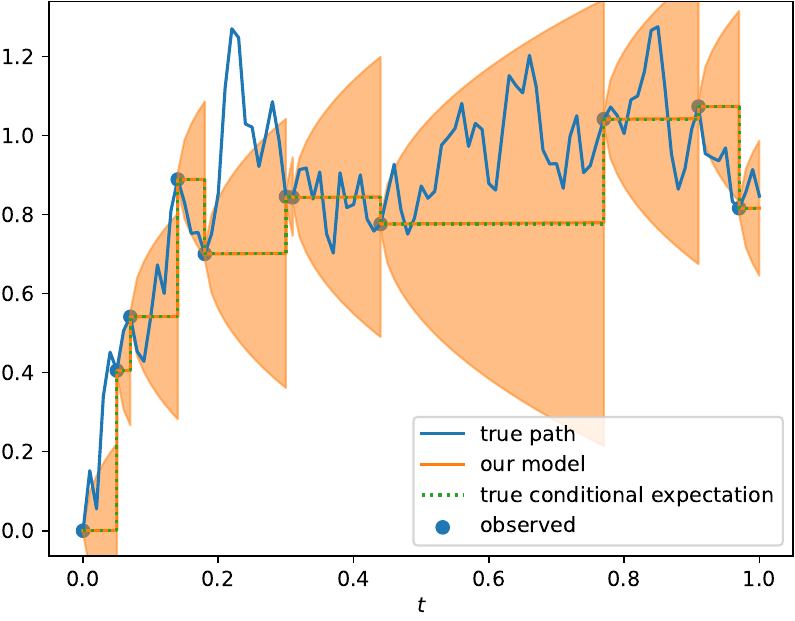}
\caption{Left: a test sample of a Brownian motion $X$ (top) and its square $X^2$ (bottom) together with the predicted and true conditional expectation. Right: the same test sample of the Brownian motion $X$ with a confidence interval given as $\hat{\mu}_t \pm \hat{\sigma}_t$. }
\label{fig:BMVar}
\end{figure}

We apply NJ-ODE to a Brownian motion $X$ and its square $X^2$, as described in Example~\ref{example:BM and Cond Var}, to show that our model can be used to compute the conditional variance, which gives rise to an uncertainty estimate of the prediction.
In Fig.~\ref{fig:BMVar}, we show the predictions of the two-dimensional process $(X, X^2)$, as well as the prediction of $X$ together with a confidence region defined as $\hat{\mu}_t \pm \hat{\sigma}_t$, where $\hat{\mu}_t$ is NJ-ODE's predicted conditional expectation and $\hat{\sigma}_t^2$ is its predicted conditional variance at time $t$. The evaluation metric on the two-dimensional dataset becomes smaller than $5 \cdot 10^{-5}$.

\subsection{Path dependent process -- fractional brownian motion}\label{sec:Path Dependence -- Fractional Brownian Motion}
\begin{table}
\vspace{-0.0cm}
\begin{center}
\caption{Minimal evaluation metrics on the test set of FBM (with $H=0.05$) within the $200$ epochs of training for different NJ-ODE models.}
\begin{tabular}{| c | r | r | r | r |}
\toprule
 & NJ-ODE & NJ-ODE (with sig.) & NJ-ODE (with RNN) & PD-NJ-ODE \\
\midrule
min. evaluation metric & $8.1 \cdot 10^{-2}$ & $1.0 \cdot 10^{-2}$ & $1.1 \cdot 10^{-2}$ & $0.5 \cdot 10^{-2}$ \\
\bottomrule
\end{tabular}
\label{tab:NJODE FBM Comparison}
\end{center}
\end{table}

\begin{figure}[htp!]
\centering
\includegraphics[width=0.49\textwidth]{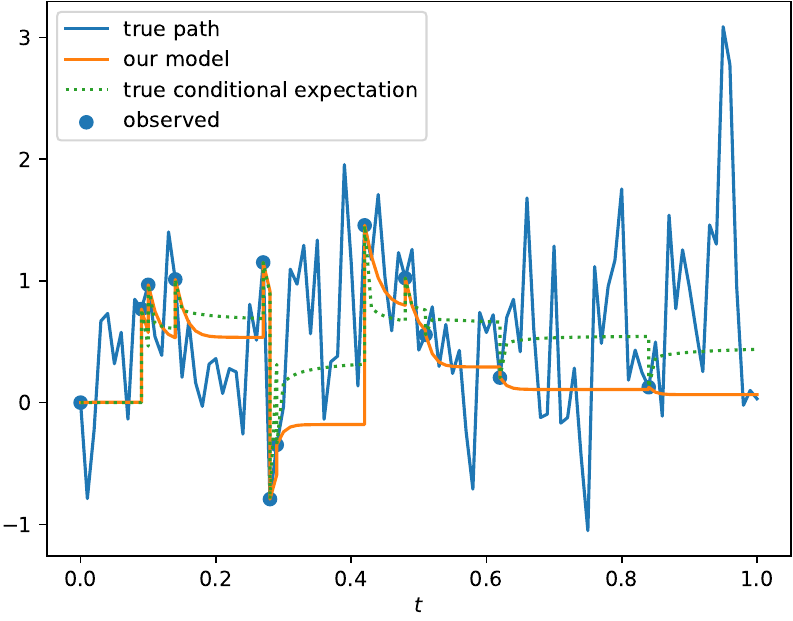}
\includegraphics[width=0.49\textwidth]{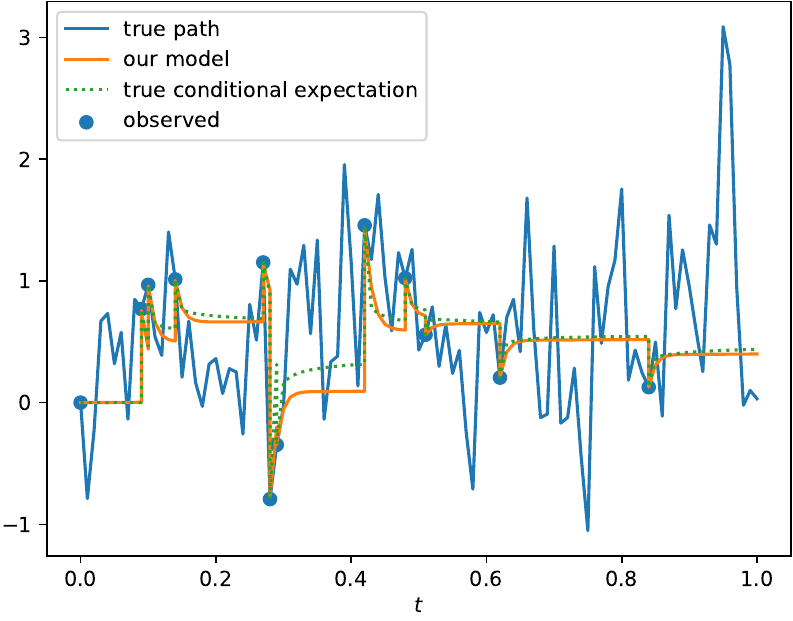}\\
\includegraphics[width=0.49\textwidth]{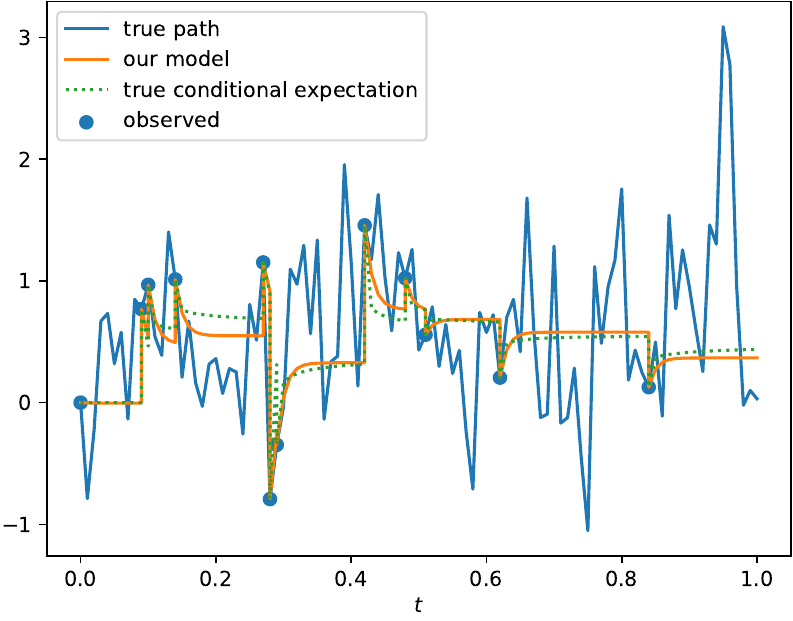}
\includegraphics[width=0.49\textwidth]{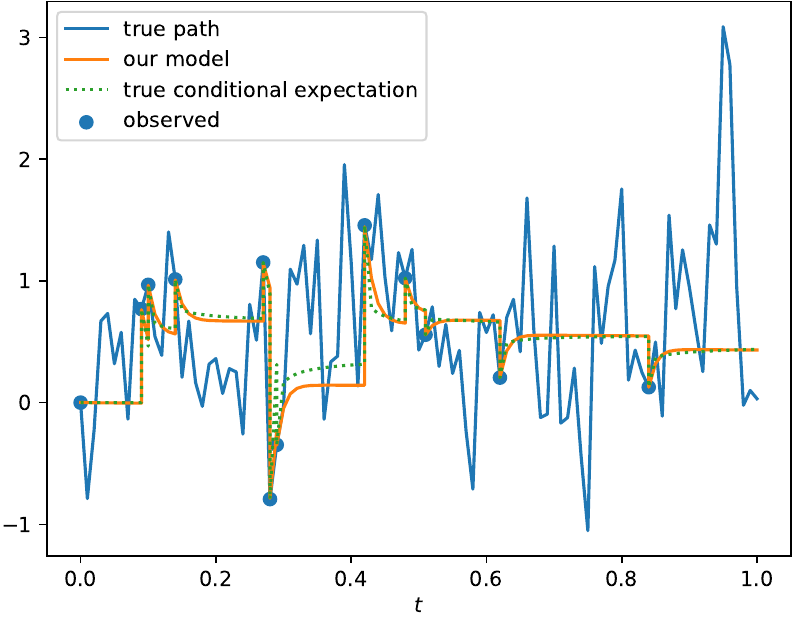}
\caption{NJ-ODE (top-left), NJ-ODE with signature input (top-right), NJ-ODE  with recurrent jump network (bottom-left) and PD-NJ-ODE (bottom-right) on a test sample of a fractional Brownian motion with Hurst parameter $H=0.05$. }
\label{fig:FBM}
\end{figure}

Path-dependent processes are one of the main application areas for the extension of NJ-ODE. Here, we use a fractional Brownian motion (FBM) with Hurst parameter $H=0.05$, which yields a rough process with high negative autocorrelation (cf. Section~\ref{sec:FBM}).
To show that PD-NJ-ODE performs better than the standard NJ-ODE, we compare them as well as 2 intermediate versions, where once the signature is added as input to NJ-ODE and once a recurrent jump network is used in NJ-ODE.
The same architecture is used for all four models.
The optimal evaluation metric for each of the models is given in Table~\ref{tab:NJODE FBM Comparison}. PD-NJ-ODE achieves the best evaluation metric, for the two intermediate models it is approximately doubled and for NJ-ODE it is about $16$ times larger.\\
Although our theoretical results imply that the NJ-ODE with signature should be enough, the performance can depend on the truncation level used for the signature input.
All truncation levels between $m \in \{1, \dotsc, 10 \}$ were tested. For $m=1$ and $m=2$ the minimal evaluation metric was $1.1 \cdot 10^{-2}$ and $0.6 \cdot 10^{-2}$. For all $m \geq 3$, it was $0.5 \cdot 10^{-2}$.
The computation time grows with increasing truncation level $m$, since it depends on the input dimension of the neural networks which increases according to \eqref{eq:sig_nb_terms} for growing $m$.
Therefore, we choose to use a truncation level of $2$ or $3$ for all of our experiments, since this seems to be a good trade-off between model performance and computation time. Here we report results with $m=3$. \\
Since truncation levels up to $m=10$ are still rather small, it is not surprising that the recurrent structure can carry additional path information not captured by the truncated signature and therefore improve performance.\\
On the other hand, these results also show that the recurrent structure alone does not carry all the necessary path information.
These differences in the quality of the predictions can also be seen in Fig.~\ref{fig:FBM}.

\subsection{Incomplete observations -- Correlated two-dimensional brownian motion}\label{sec:Incomplete Observations -- Correlated 2-Dimension Brownian Motion}
\begin{figure}
\centering
\includegraphics[width=0.6\textwidth]{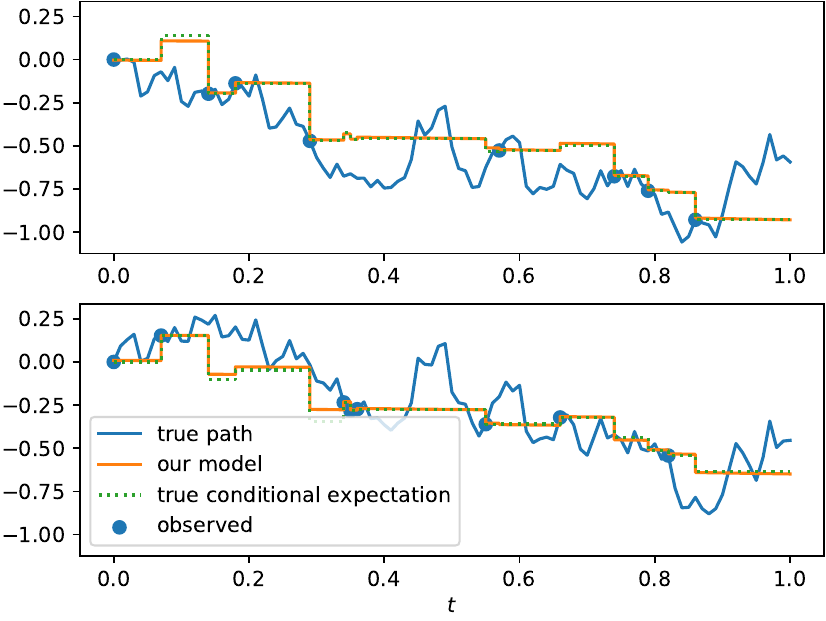}
\caption{Predicted and true conditional expectation on a test sample of the 2-dimensional correlated Brownian motion (first coordinate plotted on top, second on bottom). The PD-NJ-ODE adjusts its prediction well for both coordinates when observing only one of them.}
\label{fig:BM2D}
\end{figure}

To test the model on a synthetic dataset with incomplete observations, we consider the example of a two-dimensional correlated Brownian motion, as described in Section~\ref{sec:Multivariate Process with Incomplete Observations: Correlated Brownian Motions}, with $\alpha^2 = 0.9$. We first sample observation times from the discretization grid as usual, and at each observation time one of the two coordinates is chosen at random for observation (cf. Appendix~\ref{sec:Synthetic Datasets}, where $\lambda=0$ is used).
The derivations in Section~\ref{sec:Multivariate Process with Incomplete Observations: Correlated Brownian Motions} show that the incomplete observations also introduce some path dependence.
The path-dependent NJ-ODE model achieves a minimal evaluation metric of $5.2 \cdot 10^{-3}$, and we see that its prediction of one coordinate responds well when the other coordinate is observed, even after multiple consecutive observations of the same coordinate (Fig.~\ref{fig:BM2D}).

\subsection{Stochastic filtering of a Brownian motion with Brownian noise}\label{sec:Stochastic Filtering of a Brownian Motion with Brownian Noise}
\begin{figure}
\centering
\includegraphics[width=0.6\textwidth]{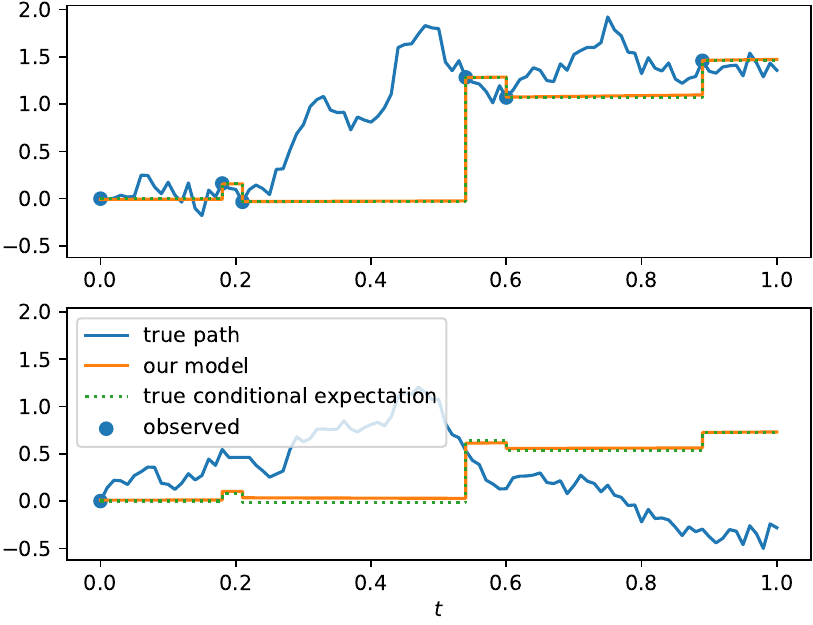}
\caption{Predicted and true conditional expectation on a test sample of the stochastic filtering dataset. The first coordinate (plotted on top) is the observation process $Y = \alpha X + W$ and the second coordinate (bottom) is the signal process $X$ (which is never observed on test samples).}
\label{fig:Stochastic Filtering}
\end{figure}

PD-NJ-ODE is applied to the stochastic filtering problem that was described in Section~\ref{sec:Filtering Problem with Brownian Motions}, where $\alpha=1$ and  observation probabilities $p_k = 0.25$, for the $X$-coordinates of the training set, are used.
The model achieves a minimal evaluation metric of $5.4 \cdot 10^{-4}$ and we see that it learns to use new observations of $Y$ to  update the predictions of $X$ (Fig.~\ref{fig:Stochastic Filtering}).

\subsection{PhysioNet}\label{ref:Physionet -- Real World Dataset}
We test the PD-NJ-ODE model on the extrapolation task of \cite{ODERNN2019} using the PhysioNet Challenge 2012 dataset \citep{physionet} and compare its results with those of NJ-ODE. Table~\ref{tab:physionet} shows that PD-NJ-ODE outperforms NJ-ODE. Although the results for PD-NJ-ODE are computed using single-hidden-layer neural networks with $50$ hidden nodes, compared with two-hidden-layer networks with $50$ hidden nodes for NJ-ODE, the number of parameters is much larger because the signature truncated at level $2$ is used as an additional input (for the $41$-dimensional underlying process, this amounts to $1'723$ additional inputs).
These results suggest that there is some (small) path-dependence in the PhysioNet dataset, which could be dealt with by the PD-NJ-ODE model.

\begin{table}[t]
\vspace{-0.0cm}
\begin{center}
\caption{Mean and standard deviation of MSE on the test set of physionet. Results of baselines were reported by \cite{ODERNN2019} and \cite{herrera2021neural}. Where known, the number of trainable parameters is reported.}
\begin{tabular}{| l | c | c |}
\toprule
& Physionet -- MSE $(\times 10^{-3})$ & \# params \\
\midrule
RNN-VAE & $3.055 \pm 0.145 $ & - \\ 
Latent ODE (RNN enc.) & $3.162 \pm 0.052$ & - \\ 
Latent ODE (ODE enc) & $2.231 \pm 0.029$ &  $163'972$\\ 
Latent ODE + Poisson & $2.208 \pm 0.050 $ & $181'723$ \\
\midrule
NJ-ODE &  1.945 $\pm$ 0.007 & $24'423$ \\ 
PD-NJ-ODE &  \textbf{1.930 $\pm$ 0.006} & $201'691$\\
\bottomrule
\end{tabular}
\label{tab:physionet}
\end{center}
\vspace*{-0.0cm}
\end{table}

\subsection{Limit order book datasets}\label{sec:Limit Order Book Data}
Stock and cryptocurrency exchanges use a limit order book (LOB) to track all limit orders\footnote{A limit order is an order to buy (sell) a certain amount of an asset at or below a specified maximum price (at or above a specified minimum price).} placed by agents who want to buy or sell assets traded on the exchange.
For each asset, the LOB has a buy side and a sell side and lists, for each side, all price levels together with the respective order volumes at which agents would like to buy or sell. The midprice is defined as the mean of the best bid (buy order) and ask (sell order) prices. A level-$n$ LOB, for $n \in \N$, is a truncated version that includes only the $n$ best (highest) bid prices and the $n$ best (lowest) ask prices. In the following, we always work with level-$10$ LOBs.
Whenever a new order is made or an order is cancelled, the LOB is updated. New limit orders are added to the book, while new market orders are directly executed against the best available limit orders. Hence, LOBs are updated irregularly in time, making them a natural example to apply the PD-NJ-ODE.

We test our model on cryptocurrency LOB data. The first dataset (denoted ``BTC'') is based on one day of Bitcoin LOB data, kindly provided by Covario. The other datasets (denoted ``BTC1sec'' and ``ETH1sec'') are based on roughly 12 days of Bitcoin (BTC) and Ethereum (ETH) LOB snapshots recorded at a frequency of one second (the state of the LOB is saved once per second rather than after every update). Using LOB snapshots at a predefined frequency usually entails some information loss compared with using the full LOB. However, these datasets are publicly available \citep{HFLOB}, which is why we include them here. For all datasets, any data point at which the first 10 levels of the LOB did not change relative to the previous point is deleted.
The datasets are split into non-overlapping samples, each comprising $100$ consecutive data points as input and the data point $10$ steps ahead as the prediction target (label).

\begin{table}[t]
\vspace{-0.0cm}
\begin{center}
\caption{Minimal MSEs (smaller is better) during the training of each model (if applicable) are reported for different LOB datasets.}
\begin{tabular}{| l | c | c | c |}
\toprule
&  BTC & BTC1sec & ETH1sec \\
\midrule
last observation & $0.11808 $ & $1350.44 $ & $ 2.58909 $ \\
best LinReg / RF & $0.12198 $ & $1355.04$ & $ 2.58253 $ \\
\midrule
PD-NJ-ODE & \textbf{0.11743} & \textbf{1343.91}  & \textbf{2.56636} \\ 
\bottomrule
\end{tabular}
\label{tab:LOB MSE}
\end{center}
\vspace*{-0.0cm}
\end{table}

\begin{table}[t]
\vspace{-0.0cm}
\begin{center}
\caption{
The true mean value and PD-NJ-ODE's mean predicted value  of the midprices $10$ steps ahead for the 3 different datasets and their subsets by labels.}
\begin{tabular}{| l | r  | r | r |  r  | r | r |}
\toprule
&  \multicolumn{2}{c|}{BTC} & \multicolumn{2}{c|}{BTC1sec} & \multicolumn{2}{c|}{ETH1sec} \\
& \multicolumn{1}{c|}{true} & prediction & \multicolumn{1}{c|}{true} & prediction & \multicolumn{1}{c|}{true} & prediction \\
\midrule
overall & $0.01143$ & $0.01123$ 			   & $ -2.40790  $ & $ -2.39125 $ & $ -0.12932  $ & $  -0.10787 $\\
decrease & $ -0.41470  $ & $ -0.31175  $ & $ -47.28358  $ & $ -28.80446 $ & $ -2.19563 $ & $  -1.25667 $ \\
stationary & $0.08016  $ & $ 0.08585  $  & $ -2.75307 $ & $ -3.27602 $ & $ -0.34508 $ & $ -0.36719  $ \\ 
increase & $ 0.44849  $ & $ 0.31417  $    & $ 40.63428 $ & $ 23.33839 $ & $ 2.24031 $ & $ 1.30459  $ \\
\bottomrule
\end{tabular}
\label{tab:LOB true and predicted means}
\end{center}
\vspace*{-0.0cm}
\end{table}

The common task in the literature is to predict price movements, i.e., whether the midprice increases, decreases or stays roughly constant \citep{tran2017tensor, tsantekidis2017forecasting, dixon2017classification, ntakaris2018benchmark, passalis2018temporal, tsantekidis2020using, zhang2019deeplob}.
As a baseline, we use the state-of-the-art DeepLOB model of \citet{zhang2019deeplob}, which was shown to outperform many other models.\\
To compare with these results, we extend the PD-NJ-ODE model with a classifier network, which maps the latent variable $H_{t_n}$, after the last data point has been processed, to the probabilities that the label $10$ steps ahead belongs to one of the classes (increase, decrease, or stationary).
This classifier network is trained using a cross-entropy loss, first together with the remaining PD-NJ-ODE architecture and then again on its own.
We retrain the classifier because the training of the PD-NJ-ODE model is relatively time-consuming, while training the classifier alone is very fast. Since the $50$ epochs for which the PD-NJ-ODE is trained are not sufficient for the classifier to reach its best performance,  we retrain it for another $1000$ epochs on its own.

The natural task for our model is the regression problem of forecasting the midprice $10$ steps ahead.
Because baseline models for this task are scarce in the literature, we compare the PD-NJ-ODE midprice prediction with three simple baselines. First, we use the midprice of the last observation as the prediction for the midprice $10$ steps ahead, i.e., the stationary prediction. Second, we fit linear regression models to all features of the training dataset or subsets thereof (midprices and bid/ask prices but no volumes; midprices only; midprices from only the last 10 observation times; or the midprice from only the last observation time).
As a final baseline, we fit random forest (RF) regression models to subsets of the training-dataset features (the midprice, bid/ask prices, and volumes from the last 10 observation times; the midprice and bid/ask prices but no volumes from the last 10 observation times; or only the midprice from the last 10, 5, or 3 observation times).
For all datasets, the best-performing linear regression model uses only the midprice of the last observation and outperforms the best RF model; therefore, only the results of this model are reported.

Table~\ref{tab:LOB MSE} shows that, in the regression task, the PD-NJ-ODE model outperforms the best baseline method by about $0.5\%$ on each of the three datasets. Although this improvement is relatively small, it is important to note that all linear regression models perform worse than the stationary prediction on both Bitcoin datasets, while on the Ethereum dataset only the simplest linear model, using only the midprice at the last observation time as a predictor, performs slightly better than the stationary prediction.
Moreover, all RF models perform worse than the stationary prediction.
In particular, this implies that even very simple linear models overfit the noise in the LOB data. By contrast, the PD-NJ-ODE model manages to extract at least a little more information from the LOB paths without fitting their noise.

\begin{figure}
\centering
\includegraphics[width=0.7\textwidth]{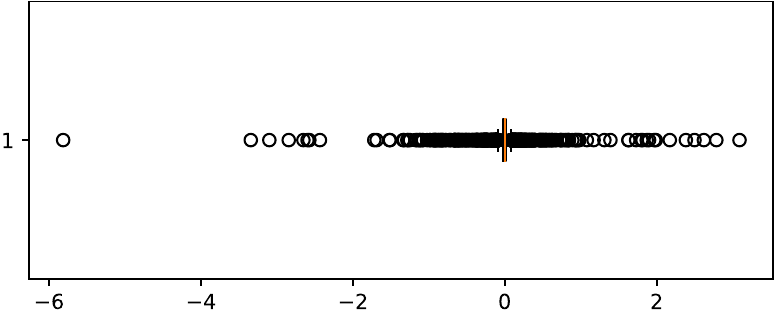}
\caption{ 
Boxplot of the prediction errors of the PD-NJ-ODE model on the BTC dataset.
}
\label{fig:LOB error distr BTC}
\end{figure}

In Table~\ref{tab:LOB true and predicted means}, we show our model's mean prediction and the true mean of the midprice $10$ steps ahead for the entire datasets and for the subsets corresponding to the three labels. Taking the means is a useful way to average out random sample noise. At the same time, however, some genuine information is also averaged out (see, for example, the difference between the overall mean and the means of the three labels).
For all three datasets, the overall mean prediction is relatively close to the true mean, while the absolute values of the mean predictions for the decrease and increase labels are significantly smaller than the true values. This can be explained by the boxplots of the prediction errors (i.e., the differences between predicted and corresponding true values) shown in Fig.~\ref{fig:LOB error distr BTC} for the BTC dataset and in Fig.~\ref{fig:LOB error distr BTC1sec and ETH1sec} in the appendix for the other two datasets. Although the distributions are closely concentrated around $0$, there are many large outliers on both sides, which increase the true absolute means for the decrease and increase labels. By contrast, the PD-NJ-ODE model appears to learn the main behaviour without overfitting these large-valued outliers, thereby keeping the predicted absolute means smaller in those two groups. In particular, this suggests that PD-NJ-ODE is both robust and flexible.

The results of the classification task for the three datasets are shown in Table~\ref{tab:LOB F1 score}, where we compare the weighted F1-scores of the models, as in \cite{zhang2019deeplob}. For the two Bitcoin datasets, PD-NJ-ODE (with a retrained classifier) and DeepLOB yield very similar results. On the Ethereum dataset, our model achieves slightly better results than on ``BTC1sec'', whereas DeepLOB performs significantly worse; we have no explanation for this difference.\footnote{The DeepLOB model was trained several times to exclude the possibility that this was due to an unfortunate initialization or a similar issue.}
Overall, we conclude that, although classification is not the core task of our model, the results obtained by adding a simple classifier network on top of the PD-NJ-ODE model are already very promising.

\begin{table}[t]
\vspace{-0.0cm}
\begin{center}
\caption{Maximal F1-scores (in $\%$, larger is better) during the training of each model are reported for different LOB datasets.}
\begin{tabular}{| l | c | c | c |}
\toprule
&  BTC & BTC1sec & ETH1sec \\
\midrule
DeepLOB & $61.59$ & \textbf{52.24} & $ 44.97$ \\
\midrule
PD-NJ-ODE (no retraining) & $52.34$ & $ 52.02$ & $ 52.48$  \\ 
PD-NJ-ODE (classifier retrained) & \textbf{62.06} & no improvement & \textbf{54.08} \\ 
\bottomrule
\end{tabular}
\label{tab:LOB F1 score}
\end{center}
\vspace*{-0.0cm}
\end{table}

\section{Conclusion}
We extended the NJ-ODE model to work for much more general datasets and settings.
In \citet{herrera2021neural}, there were examples for which NJ-ODE worked well empirically, although they were not covered by the theoretical setting. Here, we aligned the theoretical and empirical results by imposing the weakest possible constraints on the data-generating process under which PD-NJ-ODE can be applied successfully. Indeed, if the conditional expectation is not continuously differentiable, it is clear that the proposed framework cannot approximate it arbitrarily well, since neural networks cannot approximate the corresponding functions arbitrarily well.
On the other hand, if the integrability assumption is not satisfied, the loss function is not well defined.
Moreover, using multiple synthetic datasets, we showed empirically that PD-NJ-ODE works in all the more complicated settings now permitted by the theoretical results.
Finally, applying PD-NJ-ODE to limit order book data yielded very promising results, which will be explored further in future work.

\if\addackn1
\section*{Acknowledgement}
	This paper grew out of Marc N\"ubel's master's thesis project, carefully supervised by Calypso Herrera and Florian Krach. We are grateful to Calypso Herrera for many helpful discussions and contributions during this phase of the project, as well as for proofreading the first version of the manuscript. Moreover, the authors thank Andrew Allan, Robert A. Crowell, and F\'elix B. Tambe Ndonfack for helpful discussions and feedback; William Andersson for his meticulous proofreading at later stages of the manuscript; and Jakob Heiss for his helpful suggestions on making the paper more precise.
	The authors also thank Covario for providing the BTC LOB dataset.
    Finally, the authors thank the anonymous reviewer for their thoughtful feedback and careful proofreading that significantly improved the paper.
\fi

\bibliographystyle{iclr2021_conference}

\if\inclapp1
\clearpage
\appendix
\section*{Appendix}

\section{A Generalized Version of the Filtering Problem}\label{sec:A Generalized Version of the Filtering Problem}
The filtering problem of Section~\ref{sec:Stochastic Filtering Problem} can be generalized in two simple ways supported by the PD-NJ-ODE model without requiring any additional assumptions.
First, the $Y$-coordinates need not be observed completely. It is not even necessary for every $Y$-coordinate to be observable with positive probability at every observation time. To satisfy our assumptions, we require only that the same coordinates of $Y_0$ are always observed. This generalisation accommodates the possibility that certain sensors may malfunction or be switched off from time to time.
On the other hand, some of the $X$-coordinates might be observed from time to time, which means that the ground truth is sometimes (partly) observable. Again, to satisfy our assumptions, we assume that always the same coordinates of $\varphi(X_0)$ are observed.
Without loss of generality, we also assume that at least one coordinate is observed at each observation time, since otherwise it would not be an observation time and could be omitted. However, it is possible, for example, that only some $X$-coordinates and no $Y$-coordinates are observed at certain observation times.

Let  $\mathcal{Z}_t$  be the $\sigma$-algebra describing the currently available information (of evaluation samples). Moreover, let $\varphi : \R^{d_X} \to \R^{d_\varphi}$ be any integrable measurable function that can be evaluated with incomplete observations (e.g. a function that is applied element-wise).
\begin{rem}
In the case that the $X$-coordinates are never observed in the evaluation samples, $\varphi$ can be any integrable measurable function.
\end{rem}
Then, the filtering problem amounts to computing
\begin{equation*}
\E[ \varphi(X_t) \, | \, \mathcal{Z}_t ].
\end{equation*}

To train the PD-NJ-ODE model to approximate this random function, we generate a training dataset similarly as in Section~\ref{sec:Applying PD-NJ-ODE to the Filtering Problem}, with the difference that the generated observation mask has the probability  $p_k \in (0,1)$ for $k \geq 1$ for each coordinate of $M_k$ to be $1$  individually. At $t=0$ exactly those coordinates of $M_0$ corresponding to the coordinates of $(\varphi(X_0), Y_0)$ that are always observed, are $1$.
\begin{rem}
In the case that the $X$-coordinates are never observed in the evaluation samples, the coordinates of $M_k$, $k\geq 1$, corresponding to $X$ are generated either all $0$ or all $1$ simultaneously (as before), while the coordinates corresponding to $Y$ are sampled individually.
\end{rem}
Then the following results follow similarly as Corollary~\ref{cor:convergence in filtering problem}.

\begin{cor}
Assume that the training samples are generated as described above and that $Z = (\varphi(X), Y)$ satisfies Assumptions~\ref{assumption:4} and~\ref{assumption:5}.
Let $\theta^{\min}_{m,N} \in \Theta^{\min}_{m,N} := \arg \inf_{\theta \in \Theta_m}\{ \hat\Phi_N(\theta)\}$ for every $m, N \in \N$.
Then, one can define an increasing sequence $(N_m)_{m \in \N}$ in $\N$ such that for every $1 \leq k \leq K$ the following statements hold.
\begin{enumerate}
\item $G^{\theta_{m, N_m}^{\min}}$ converges to $\hat{Z} = (\widehat{\varphi(X)}, \hat{Y})$
in the metric $d_k$ as $m \to \infty$.
\end{enumerate}

In the case that the $X$-coordinates are never observed in the evaluation samples, we write $\mathcal{Y}_t = \mathcal{Z}_t$ (even though here  the $Y$-coordinates might not be observed completely). Then  the following statements hold.
\begin{enumerate}
\setcounter{enumi}{1}
\item $G^{\theta_{m, N_m}^{\min}}$ converges to $\hat{Z}$
in the metric $\tilde{d}_k$ as $m \to \infty$.

\item
It holds that
\begin{equation*}
\tilde{d}_k \left(  G^{\theta_{m, N_m}^{\min}}(Z), G^{\theta_{m, N_m}^{\min}}(Y) \right) = \tilde{d}_k \left( \hat{Z} ,  \left( \E[ Z \, | \,  \mathcal{Y}_t] \right)_{t \in [0,T]} \right) = 0 .
\end{equation*}
\end{enumerate}
\end{cor}

\section{Additional Examples of Processes Satisfying the Assumptions}\label{sec:Additional Examples of Processes Satisfying the Assumptions}

\subsection{Stochastic functional differential equations}\label{sec:Stochastic Functional Differential Equations}

Following the definitions in \citet[Section 16.2]{cohen2015stochastic}, let $\mathcal{D}$ be the space of c\`adl\`ag adapted processes on $\R^d_X$ and, for $2 \leq p < \infty$, let $S^p \subseteq \mathcal{D}$ be the set of all processes $X \in \mathcal{D}$ for which
\begin{equation*}
\| X \|_{S^p} := \| X^\star_T \|_{L^p} = \E[ (X_T^\star)^p ]^{1/p} < \infty,
\end{equation*}
where $X^\star = \max_i\{ X_i^\star \}$ for multidimensional processes and where we use the finite time horizon $T$ instead of $\infty$. We look at functions $f : \Omega \times [0,T] \times \mathcal{D} \to \R^{u \times v}, (\omega, t, X) \mapsto f(\omega, t, X) = f(X)$ that are uniformly Lipschitz, i.e., there exists some $L \in \R$ such that for any $X, Y \in \mathcal{D}$  and $t \in [0,T]$,
\begin{equation}\label{equ:functional lipschitz}
(f(X) - f(Y))^\star_t \leq L (X -Y )^\star_{t-}.
\end{equation}
For such functions we write $f \in \operatorname{Lip}(L)$. It is clear from this definition that $f(X)_t$ only depends on $(X_s)_{0 \leq s \leq t}$. Moreover, for such functions we have that $\| f(X) - f(Y) \|_{S^p} \leq L  \| X-Y \|_{S^p}$.

Let $2 \leq p <\infty $, $d_W \in \N$ and let $(W_t)_{t\in[0,T]}$ be a $d_W$-dimensional Brownian motion on
the probability space $(\Omega,\mathcal{F},\mathbb{F}:= \{ \mathcal{F}_t\}_{0\leq t\leq T},\P )$.
We assume that the stochastic process $X:=(X_t)_{t\in[0,T]}$ is defined as the continuous solution of the stochastic functional differential equation
\begin{equation}\label{eq:SFDE}
    dX_t
    =\mu\left(\omega, t, X \right)dt
    +\sigma\left(\omega, t, X \right)d{W}_t = \mu(X)_t dt + \sigma(X)_t dW_t
\end{equation}
for all $0\leq t \leq T$ with (random) starting point $X_0$ taking values in $\R^{d_X}$ such that $\| X_0 \|_{L^p}<\infty$, drift function $\mu: \Omega \times [0,T] \times \mathcal{D} \to \R^{d_X}$ and diffusion function $\sigma: \Omega \times [0,T] \times \mathcal{D} \to \R^{d_X \times d_W}$.
We assume that $\mu$ and $\sigma$ are uniformly Lipschitz in the sense of \eqref{equ:functional lipschitz} and that there exists a constant $C > L$ such that $\| \mu(0) \|_{S^p} +  \| \sigma(0) \|_{S^p} < C$.
Moreover, we assume that we always observe all coordinates of $X$ simultaneously.
These assumptions imply that $\mu$ and $\sigma$ have linear growth,
\begin{align*}
\| \mu(X) \|_{S^p} &= \| \mu(0) \|_{S^p} + (\| \mu(X) \|_{S^p} - \| \mu(0) \|_{S^p})
\leq  \| \mu(0) \|_{S^p}  + \| \mu(X)  - \mu(0) \|_{S^p} \\
& \leq   \| \mu(0) \|_{S^p}  + L \| X \|_{S^p} \leq C (1+ \| X \|_{S^p} ),
\end{align*}
and similarly for $\sigma$.

It is easy to see that $(W_t)_{0 \leq t \leq T}$ and $(t)_{0 \leq t \leq T}$ are both $\alpha$-sliceable for any $\alpha > 0$  \citep[Definition 16.3.8]{cohen2015stochastic}, in fact even with deterministic stopping times.
Therefore, \citet[Lemma 16.3.10]{cohen2015stochastic} implies that there exists a unique solution to \eqref{eq:SFDE} and that this solution satisfies
\begin{equation*}
\| X \|_{S^p} \leq \tilde{C} (\| X_0 \|_{L^p} + \| \mu(0) \|_{S^p} +  \| \sigma(0) \|_{S^p}),
\end{equation*}
for some $\tilde C$ depending only on $L$ and $W$. In particular, we can choose $C$ such that $\| X \|_{S^p} < C$.
This implies the integrability of the drift and diffusion components
\begin{align}\label{equ: integrability of drift and diffusion}
\E\left[ \int_0^T |\mu(X)_t| + |\sigma(X)_t |^2 \, dt  \right] &= \int_0^T \E\left[  |\mu(X)_t| + |\sigma(X)_t |^2  \right] \, dt  \nonumber\\
&\leq  \int_0^T \left( \| \mu(X) \|_{S^p} + \| \sigma(X) \|_{S^p}^2 \right) \, dt \nonumber\\
&\leq 2 T C^2(1+ C)^2,
\end{align}
where we used Fubini's Theorem in the first and HÃ¶lder's inequality in the second step.

\eqref{equ: integrability of drift and diffusion} yields that the process
\[
    \tilde{M}_t:= \int_0^t \sigma(s,X_s,X_{<s})d{W}_s,\quad 0\leq t\leq T,
\]
is a square-integrable martingale by \citet[Lemma before Thm. 28, Chap. IV]{Pro1992}, since the Brownian motion ${W}$ is square-integrable with $d[W^i,W^j]_t = \delta_{i,j} dt $.
Using the martingale property of $\tilde{M}$ and the tower property for $\mathcal{A}_{\tau(t)} \subseteq \mathcal{F}_{\tau(t)}$, we have ($\tilde{\omega}$-wise) for every $t\in[0,T]$,
\begin{align}
    \hat{X}_t
    &= \mathbb{E}_\mathbb{P}\left[\left(X_t-X_{\tau(t)}\right)
    +X_{\tau(t)}\vert\mathcal{A}_{\tau(t)}\right]\nonumber\\
    &= X_{\tau(t)}
    +\mathbb{E}_\mathbb{P}\left[\int_{\tau(t)}^t \mu(X)_r \, dr
    \vert\mathcal{A}_{\tau(t)}\right] +\mathbb{E}_\mathbb{P}\left[\int_{\tau(t)}^t \sigma(X)_r \, dW_r
    \vert\mathcal{A}_{\tau(t)}\right]\nonumber\\
    &= X_{\tau(t)}
    +\int_{\tau(t)}^t \mathbb{E}_\mathbb{P}\left[\mu(X)_r \,
    \vert\mathcal{A}_{\tau(t)}\right]dr +\mathbb{E}_\mathbb{P}\left[ \mathbb{E}_\mathbb{P}\left[  M_t - M_{\tau(t)}
   \vert\mathcal{F}_{\tau(t)}\right]  \vert\mathcal{A}_{\tau(t)}\right]\nonumber\\
    &= X_{\tau(t)}
    +\int_{\tau(t)}^t \mathbb{E}_\mathbb{P}\left[\mu(X)_r \,
    \vert\mathcal{A}_{\tau(t)}\right]dr,\label{eq:SFDE_cond_exp}
\end{align}
where we used Fubini's theorem (for conditional expectation) in the penultimate step, which is justified by \eqref{equ: integrability of drift and diffusion}.
Let us define $\Delta:=\left\{(t,r)\in[0,T]^2\vert t+r\leq T\right\}$ and the function
\begin{align*}
    \tilde{\mu}:\Delta\times BV([0,T])
    &\rightarrow\R^{d_X}\\
    ((t,r), \xi )
    &\mapsto \mathbb{E}_\mathbb{P}\left[\mu(\omega, t+r ,X) \vert \tilde X^{\leq t} =\xi\right],
\end{align*}
then the Doob-Dynkin Lemma \citep[Lemma 2]{taraldsen2018optimal} implies that we can rewrite (\ref{eq:SFDE_cond_exp}) as
\begin{equation}
\hat{X}_t
    = X_{\tau(t)} + \int_{\tau(t)}^t \tilde{\mu} \left(  \tau(t), r-\tau(t), \tilde X^{\leq \tau(t)}  \right) dr .
\end{equation}
To satisfy Assumption~\ref{assumption:4}, we need that $\tilde{\mu}$ is continuous.
While this might not be true in general, the following examples give cases where this can be shown.

\begin{example}  Under the additional assumption that $\mu$ and $\sigma$ depend only on the current value of $X$ and not on its entire path, \citet[Theorem 8]{gubinelli}, together with \citet[Proposition B.4]{herrera2021neural}, proves that $\tilde \mu $ is continuous. Importantly, here we need neither the boundedness assumption for $\mu$ nor the integrability assumption for $\sigma$; hence, this generalizes the example from Section~\ref{sec:Ito Diffusion with Regularity Assumptions}, as described in Remark~\ref{rem:generalization ito diff}. Note that, by adding further \emph{unobserved} factors to the state space of $X$ that follow Markovian dynamics satisfying mild regularity conditions, we can incorporate many path-dependence structures in the drift and thereby obtain continuous dependence of $ \tilde \mu $ (with respect to those factors) in an analogous manner.

\end{example}

\begin{example}If we additionally assume that $\mu(X)_t = \alpha X_t + \beta$ for some $\alpha, \beta \in \R$, i.e., that it is linear in the current value of $X$, while $\sigma$ can be general and path-dependent, we can show that $\tilde \mu$ is continuous.
Indeed, then it follows   from \eqref{eq:SFDE_cond_exp} that
\begin{equation}\label{equ:FSDE-exa-cond-exp}
\mathbb{E}_\mathbb{P}\left[X_t \,
    \vert\mathcal{A}_{\tau(t)} \right]= X_{\tau(t)}
    +\int_{\tau(t)}^t \left( \alpha \mathbb{E}_\mathbb{P}\left[X_r \,
    \vert\mathcal{A}_{\tau(t)}\right] + \beta \right) dr.
\end{equation}
\eqref{equ:FSDE-exa-cond-exp} is equivalent to the ordinary differential equation
\begin{equation*}
\begin{split}
y'(t) &= \alpha y(t) + \beta, \quad t \geq \tau(t), \\
y(\tau(t)) &= y_{\tau} := X_{\tau(t)},
\end{split}
\end{equation*}
by defining $y(r):= \mathbb{E}_\mathbb{P}\left[X_r \, \vert\mathcal{A}_{\tau(t)}\right]$ for $r \geq \tau(t)$.
The unique solution to this initial value problem is given by $y(t) = -\frac{\beta}{\alpha} + \left(\frac{\beta}{\alpha} + y_{\tau}  \right) e^{\alpha (t - \tau)}$ hence the conditional expectation between $\tau(t)$ and the next observation is given by
\begin{equation*}
\mathbb{E}_\mathbb{P}\left[X_r \, \vert\mathcal{A}_{\tau(t)}\right] = F(r, \tau, \tilde X^{\leq \tau}) = -\frac{\beta}{\alpha} + \left(\frac{\beta}{\alpha} + X_{\tau(t)}  \right) e^{\alpha (t - \tau)},
\end{equation*}
which is continuous and differentiable in $t$ leading to
\begin{equation*}
\tilde{\mu} \left(  \tau, r-\tau, \tilde X^{\leq \tau}  \right) = \alpha \mathbb{E}_\mathbb{P}\left[X_r \, \vert\mathcal{A}_{\tau}\right]  + \beta = \left( \beta + \alpha X_{\tau}  \right) e^{\alpha (r - \tau)} = f (r, \tau, \tilde X^{\leq \tau})  = \frac{\partial }{\partial r} F(r, \tau, \tilde X^{\leq \tau}).
\end{equation*}

\end{example}

\subsection{Predicting a time-lagged version of an observed process}\label{sec:Predicting a Time-Lagged Version of an Observed Process}
Let $(X_t)_{t\geq 0}$ be a $1$-dimensional stochastic process, let $\alpha > 0$ and define $Y_t :=X_{t - \alpha}$, with the convention that $X_t = 0$ for all $t\leq 0$. If $X$ satisfies the integrability assumption, then also $Z := (X,Y)$ does.
We are interested in predicting $Y$ optimally given observations of $X$. This problem is related to the filtering problem in that only one process (here $X$) is observed and predictions are made for the other process (here $Y$). In contrast to the filtering problem, however, observations of $X$ also appear as observations of $Y$ after a time lag of $\alpha$.
However, it is clear that this observation structure does not satisfy our assumption that every coordinate is observed with positive probability at any observation time.
This is a problem because the (theoretical) loss function no longer has a unique minimizer. The first coordinate of a minimiser must still be given by the conditional expectation (since the assumptions hold for this coordinate); however, its second coordinate need only equal the observations of the first coordinate after a time lag of $\alpha$ (i.e., at times $t_i+\alpha$, where $t_i$ are the observation times of the first coordinate). The behaviour of the second coordinate between these points does not affect the loss, since no observations are made there.
We can address this problem in a manner similar to the filtering framework if we assume that we can sample a more general training dataset in which $Y$ may be observed between observations of $X$. Importantly, such observations of $Y$ occur only at times $t$ for which $t < \tau_X(t) + \alpha$, where $\tau_X(t)$ is the last observation time of $X$ before $t$, since otherwise the observation of $Y$ would affect the prediction of $X$. If, for any $t \in [0,T]$ with $\tau_X(t) \leq t < \tau_X(t) + \alpha$, the probability of observing $Y_t$ lies in $(0,1)$, then the minimizer of the loss function is again uniquely given by the conditional expectation process.
As in the filtering problem, the trained model can be evaluated on the original samples, where only the process $X$ is observed.

\begin{example}[Brownian Motion and its Time-Lagged Version]\label{exa:Brownian Motion and its Time-Lagged Version}
Let $X$ be a  $1$-dimensional  Brownian motion. Then $X$ satisfies $\E[(X^\star_T)^p] < \infty$ for every $1 \leq p <  \infty$ (cf.\ \citet[Lemma 16.1.4]{cohen2015stochastic}) and therefore also  $Z$.
The analytic expression of the conditional expectation can be derived as follows. For $X$ it is simply given by its last observation (cf. Section~\ref{sec:Multivariate Process with Incomplete Observations: Correlated Brownian Motions}). For $Y$ we have to distinguish between 3 cases. If $t \leq \alpha$, then $\E[Y_t | \mathcal{A}_t ] = 0$, since $Y_t = X_{t - \alpha} = 0$.
If $t \geq \tau_X(t) + \alpha$, i.e., if the last observation of $X$ was longer ago than $\alpha$, then (cf.\ \citet[Lemma 16.1.4]{cohen2015stochastic})
\[\E[Y_t | \mathcal{A}_t ] = \E[( X_{t-\alpha} - X_{\tau_X(t)}) +  X_{\tau_X(t)}| \mathcal{A}_t ] = X_{\tau_X(t)}.\]
Finally, if $h < t < \tau_X(t) + \alpha$, then $s := t - \alpha$ lies between two observation times of $X$  (which both happened before $t$).
Let us assume that there are $k$ observations until $t$ and that $1 \leq \ell \leq k$ is such that $t_0 < \dotsb < t_{\ell-1} < s < t_\ell < \dotsb <t_k$. Then $v := (X_s, X_{t_1}, \dotsc, X_{t_k})$ is a Gaussian vector with $v \sim N(0, \tilde \Sigma)$, where
\[ \tilde \Sigma :=  \begin{pmatrix}
\tilde \Sigma_{11} & \tilde \Sigma_{12} \\
\tilde \Sigma_{21} & \tilde \Sigma_{22}
\end{pmatrix},\]
with $\tilde \Sigma_{11} = \operatorname{Var}(X_{s}) = s \in \R$, $\tilde \Sigma_{12} = \tilde \Sigma_{21}^\top  = (t_1, \dotsc, t_{\ell-1}, s, \dotsc, s) \in \R^{1 \times k}$ and $\tilde \Sigma_{22} \in \R^{k \times k}$, given by $(\tilde \Sigma_{22})_{i,j} = t_{\min(i,j)}$. Note that this follows from the definition of a Brownian motion, which implies that $\operatorname{Cov}(X_a, X_b) = \min(a,b)$.
Then, the conditional distribution of $(X_s \, | \, X_{t_1}, \dotsc, X_{t_k})$ is again normal with  mean $\hat \mu := \tilde \Sigma_{12} \tilde \Sigma_{22}^{-1} (X_{t_1}, \dotsc, X_{t_{k}} )^\top $ and variance $\hat \Sigma :=  \tilde \Sigma_{11} -  \tilde \Sigma_{12} \tilde \Sigma_{22}^{-1} \tilde \Sigma_{21}$ \citep[Proposition~3.13]{Eaton2007Multi}.
In particular, we have
\[\E[ Y_{t} | \mathcal{A}_{t} ] = \E[ X_{s} | X_{t_1}, \dotsc, X_{t_k}  ] = \hat \mu. \]
A direct computation shows that
\[ (\tilde \Sigma_{22}^{-1} )_{i,j}= \begin{cases}
(t_i - t_{i-1})^{-1} + (t_{i+1} - t_i)^{-1}, & \text{if } i=j < k, \\
 (t_k - t_{k-1})^{-1}, & \text{if } i=j = k, \\
 - (t_i - t_j)^{-1}, & \text{if } i=j+1, \\
 - (t_j - t_i)^{-1}, & \text{if } j=i+1, \\
 0, & \text{otherwise},
\end{cases}\]
with which we can compute that
\[\hat \mu = \tilde \Sigma_{12} \tilde \Sigma_{22}^{-1} (X_{t_1}, \dotsc, X_{t_{k}} )^\top = X_{t_{\ell -1}} + (X_{t_\ell} - X_{t_{\ell -1}}) \frac{s - t_{\ell - 1}}{t_\ell - t_{\ell -1}}.\]
In particular this means that the conditional expectation between any two observations is given by the linear interpolation of these observations.

\end{example}

\subsection{Chaotic systems}\label{sec:Chaotic Systems}
A chaotic system is a deterministic dynamical system which is sensitive to initial conditions, topologically transitive and has dense periodic orbits \citep{hasselblatt_katok_2003}. In particular, although the system is in principle deterministic, small changes in the initial condition lead to vastly different trajectories of the system.
A chaotic system, where the initial point is chosen from some distribution supported on a compact set, with discrete observation times (where the initial point is observed completely) falls into our framework and satisfies  Assumptions~\ref{assumption:4} and~\ref{assumption:5}, if it has $C^1$ trajectories. Indeed, since $X_0$ is observed, the conditional expectation is given by the deterministic trajectory starting at $X_0$. Hence, Assumption~\ref{assumption:4} is satisfied.
Moreover, since the trajectories are continuous as functions of the initial point and time, they are bounded on $\operatorname{supp}(X_0) \times [0,T]$ (since a continuous function on a compact subset is bounded) which implies that also Assumption~\ref{assumption:5} is satisfied.

\begin{example}[Double Pendulum]\label{exa:Double Pendulum}
One of the best known examples of a chaotic system is a double pendulum as depicted in Fig.~\ref{fig:double pendulum}. Here, we shortly explain its dynamics following \citet{DoublePendulum}.
\begin{figure}[htp!]
\centering
\includegraphics[width=0.49\textwidth]{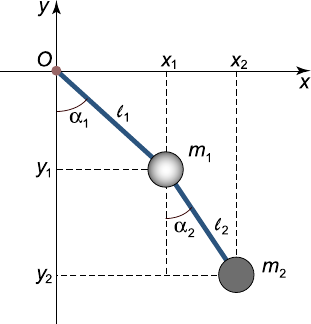}
\caption{A schematic representation of a double pendulum. Picture copied from \citet{DoublePendulum}.}
\label{fig:double pendulum}
\end{figure}
The dynamical system is at any point determined completely by a $4$-dimensional state vector $(\alpha_1, \alpha_2, p_1, p_2)$, where $(\alpha_1, \alpha_2)$ determine the current position of both pendulums and $(p_1, p_2)$ are the so-called generalized momenta, which are related to the velocities of both pendulums. The differential equation describing the dynamics of this state vector is
\begin{align*}
\alpha_1^\prime & = \frac{p_1 l_2 - p_2 l_1 \cos(\alpha_1 - \alpha_2)}{l_1^2 l_2 A_0}, \\
\alpha_2^\prime & = \frac{p_2(m_1+m_2) l_1 - p_1 m_2 l_2 \cos(\alpha_1 - \alpha_2)}{m_2 l_1 l_2^2 A_0 } ,\\
p_1^\prime &= - (m_1 + m_2) g l_1 \sin(\alpha_1) - A_1 + A_2, \\
p_2^\prime &= - m_2~g l_2 \sin(\alpha_2) +A_1 - A_2,
\end{align*}
where
\begin{align*}
A_0 &= [m_1 + m_2 \sin^2(\alpha_1 - \alpha_2)], \\
A_1 &= \frac{p_1 p_2 \sin(\alpha_1 - \alpha_2)}{l_1 l_2 A_0}, \\
A_2 &= \frac{[p_1^2 m_2 l_2^2 - 2 p_1 p_2 m_2 l_1 l_2 \cos(\alpha_1 - \alpha_2) + p_2^2 (m_1 + m_2) l_1^2 ] \sin(2(\alpha_1 - \alpha_2))}{2 l_1^2 l_2^2 A_0^2},
\end{align*}
and $g$ is the gravitational acceleration constant.
As initial points  $X_0$ we only consider positions where the double pendulum is straight, i.e., both pendulums have the same angle $\alpha_1 = \alpha_2$ and the generalized momenta  $p_1, p_2$ are $0$. In particular, we can therefore sample the initial point by sampling $\alpha$ from some distribution on $[0, 2\pi]$.

\end{example}

\section{Additional Experiments}\label{sec:Additional Experiments}

\subsection{Observation intensity depending on the underlying process}\label{sec:Observation Intensity depending on the Underlying Process}
For a Black--Scholes model $X$, we use the following method to randomly sample observation times. As usual, the process is sampled on a grid with step size $0.01$ and for each of the grid points an independent Bernoulli random variable is drawn to determine whether it is used as observation time or not (cf. Appendix~\ref{sec:Synthetic Datasets}). The only difference is that the success probability of the Bernoulli random variable is  $p = .05 + 0.4 \tanh(|X_t|/10) $, where $X_t$ is the value of the process at the given grid point.
We note that this example is not covered by the extended setting of \cite{NJODE3}. Nevertheless, the PD-NJ-ODE model applied to this dataset learns to correctly predict the process (Fig.~\ref{fig:DepIntensityBS}) achieving a minimal evaluation metric of $6.2 \cdot 10^{-4}$.
\begin{figure}[htp!]
\centering
\includegraphics[width=0.6\textwidth]{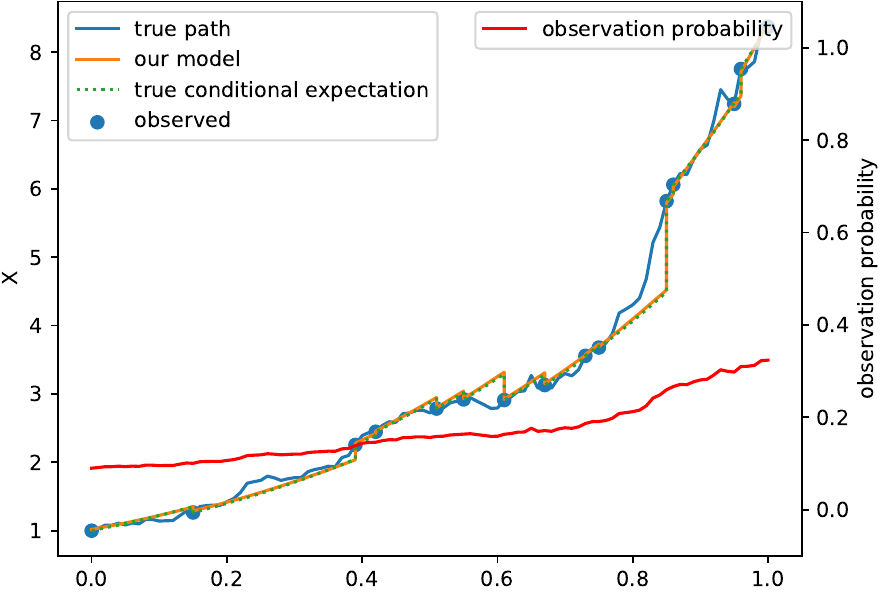}
\caption{Predicted and true conditional expectation on a test sample of the Black--Scholes dataset, where the observation intensity depends on the values of the process.}
\label{fig:DepIntensityBS}
\end{figure}

\subsection{Predicting a time-lagged Brownian motion}\label{sec:Predicting a Time-Lagged Brownian Motion}
To test how well the PD-NJ-ODE model performs in storing information about past events and reusing it at a later time, we apply it to a dataset where the second coordinate is a time-lagged version of the first coordinate, as described in Example~\ref{exa:Brownian Motion and its Time-Lagged Version}, where $\alpha = 0.19$.
This learning problem is complicated, since the model needs to learn to store past observations of $X$ (together with their observation times) and to decide which of them to use to predict $Y$ depending on their observation times and the current time. Moreover, the model needs to learn to distinguish when a new observation of $X$ changes the current prediction of $Y$ (i.e., when the previous observation of $X$ is longer ago than $\alpha$) and when it does not.
Therefore, it is not surprising that the model has some difficulties learning the correct predictions, which is also reflected in the relatively high value $1.0 \cdot 10^{-3}$ of the minimal evaluation metric. Nevertheless, we see in Fig.~\ref{fig:BM and Time-lag} that the model learns to approximate the correct behaviour.
\begin{figure}
\centering
\includegraphics[width=0.49\textwidth]{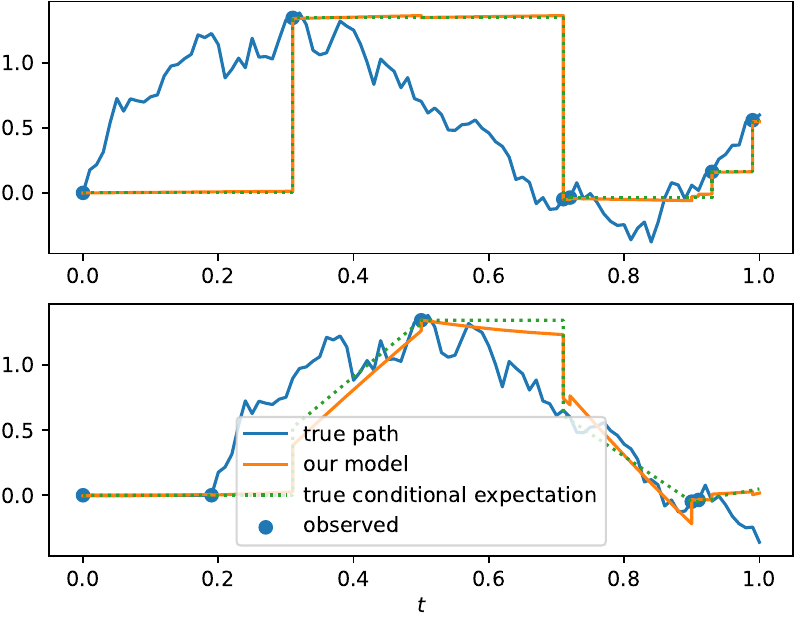}
\includegraphics[width=0.49\textwidth]{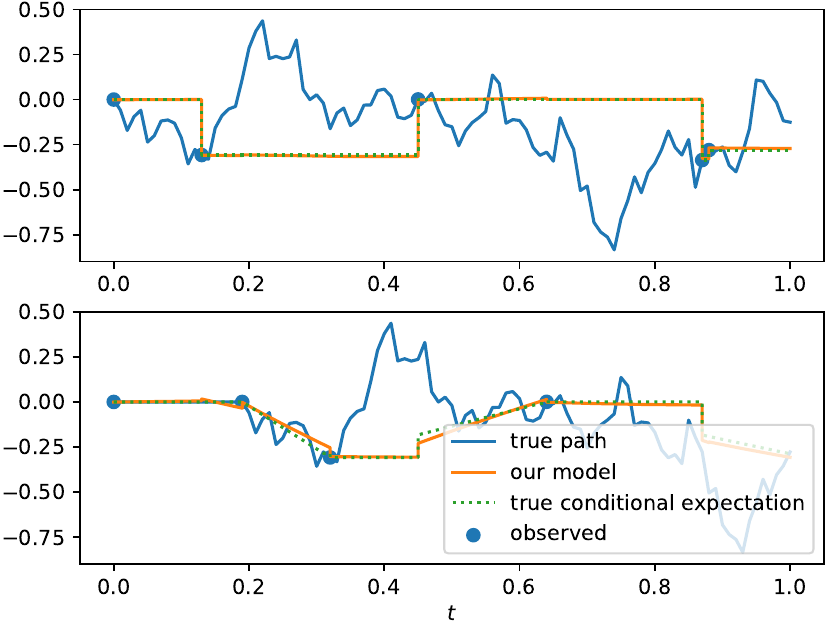}
\caption{Predicted and true conditional expectation on two test samples of the dataset, where the first coordinate (plotted on top) is a Brownian motion $X$ and the second coordinate (bottom) is its time-lagged version with lag $\alpha=0.19$.}
\label{fig:BM and Time-lag}
\end{figure}

\subsection{Chaotic system -- Double Pendulum}\label{sec:Chaotic System -- Double Pendulum}
We apply the PD-NJ-ODE to a Double Pendulum, as described in Example~\ref{exa:Double Pendulum}, where we choose $m_1=m_2=l_1=l_2 =1$ and sample the initial angle for the starting point $X_0$ from $\alpha \sim N(\pi, 0.2^2)$, i.e., normally distributed around to highest point the Pendulum could reach. In the generated dataset, we always observe $X_0$ and always have complete observations.
The PD-NJ-ODE model achieves a minimal evaluation metric of $5.1 \cdot 10^{-2}$. In particular we see that it learns to predict the deterministic dynamics of the chaotic system well, even though not free of error. At new observations, the prediction error gets reverted (Fig.~\ref{fig:Double Pendulum Experiment} left).
If only the initial point is observed, the error grows over time (Fig.~\ref{fig:Double Pendulum Experiment} right).
We note that for chaotic system dataset it is of high importance to chose the step size in the dataset small enough, such that the model can see and learn the entire dynamic (cf. Appendix~\ref{sec:Details for Double Pendulum}).
\begin{figure}
\centering
\includegraphics[width=0.49\textwidth]{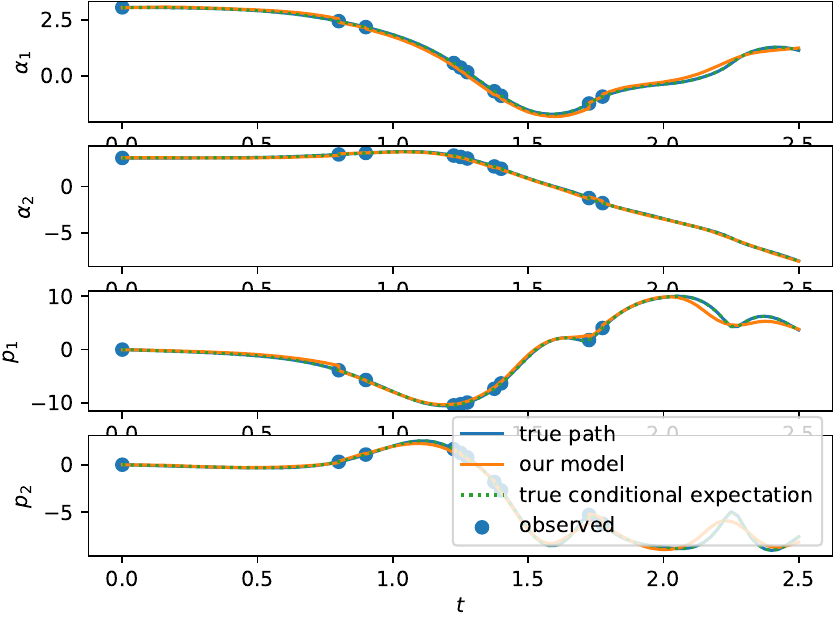}
\includegraphics[width=0.49\textwidth]{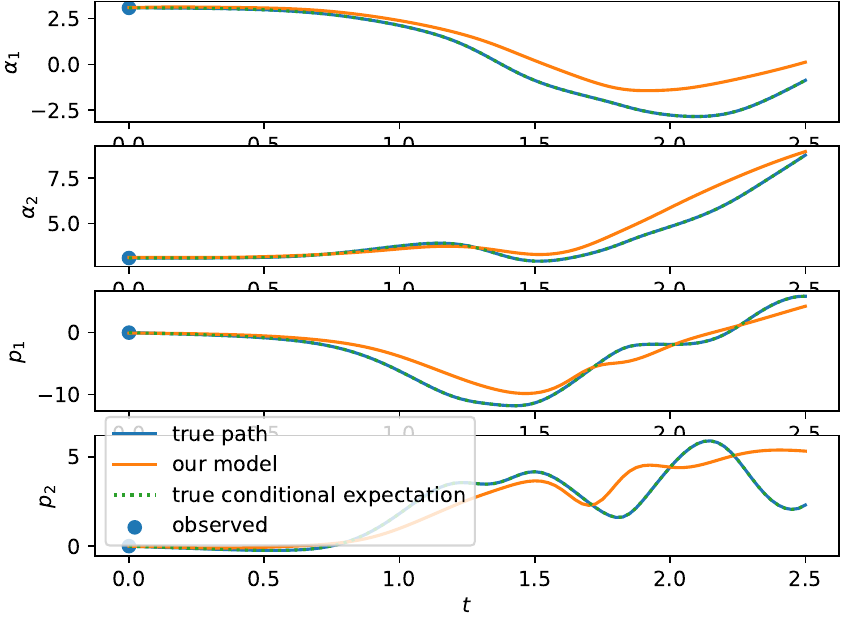}
\caption{Predicted conditional expectation on two test samples of the Double Pendulum dataset. In the sample on the left, new observations are made over time, such that the model can revert its error. In the sample on the right, only the initial point is observed, which leads to a growing error over time. 
Since the system is deterministic, the true conditional expectation coincides with the true path.}
\label{fig:Double Pendulum Experiment}
\end{figure}

\vfill\pagebreak
\section{Experimental Details}\label{sec:Experimental Details}

\subsection{Implementation details}

\subsubsection{Differences between the implementation and the theoretical description of the PD-NJ-ODE}\label{sec:Differences between the Implementation and the Theoretical Description of the PD-NJ-ODE}
Below we list all differences between the theoretical description of the PD-NJ-ODE and our implementation of it that can be found at \url{https://github.com/FlorianKrach/PD-NJODE}.
\begin{itemize}
	\item The bounded output neural networks (cf. Definition~\ref{def:bounded output NN}) $f_{\theta_1}$ and $\rho_{\theta_2}$ in \eqref{equ:PD-NJ-ODE} needed to derive our theoretical results are replaced by standard neural networks (without the bound from Definition~\ref{def:bounded output NN}) in our implementation. If the last activation function of this standard neural network is bounded (e.g., $\tanh$ is used), then the same theoretical guarantees hold (since boundedness is all we need, while the explicit form is irrelevant).
	\item For the neural network $\rho_{\theta_2}$, we use the self-imputed observation $M_{t_i }\odot X_{t_i } + (1 - M_{t_i }) \odot Y_{t_i }$ and the mask $M_{t_i }$ as additional inputs at an observation time $t_i$. Moreover, the user can choose whether to use time as an input, comprising the current time $t_i$ and the previous observation time $t_{i-1}$, and whether to use a recurrent structure with $H_{t_i-}$ as an input.
  \item We do not explicitly use $\tilde X^{\star}_t, n_t,\delta_t$ as inputs to the neural network, since they can easily be reconstructed from the extended input used and are needed only to construct the compensating neural network that ensures integrability, which is not a practical concern.
	\item For the neural network $f_{\theta_1}$ we use as additional input for times $t_i < t < t_{i+1}$  either the last self-imputed observation $M_{t_i }\odot X_{t_i } + (1 - M_{t_i }) \odot Y_{t_i }$ or the models prediction $Y_{t_i}$ at the last observation time (it is the user's choice which one is used). Moreover, the user can choose whether a $\tanh$ should be applied (element-wise) to the inputs before passing them to the network, such that they lie within $[-1,1]$.
	\item For both networks $f_{\theta_1}$ and $\rho_{\theta_2}$  one can choose to use the coordinate-wise last observation times instead of the overall last observation time $\tau$ for the time input.
	\item The user can choose whether to use a residual connection from input to output for the networks $\rho_{\theta_2}$ and $\tilde g_{\tilde \theta_3}$. In the case that a recurrent network is used for $\rho_{\theta_2}$, no residual connection is added to it.
	\item The user can choose whether to use the signature as input for $f_{\theta_1}$ and $\rho_{\theta_2}$ and up to which truncation level.
	We directly compute the signature $\pi_m (\tilde X^{\leq \tau(t)})$ instead of $\pi_m (\tilde X^{\leq \tau(t)}-X_0 )$ and before computing the signature we add the current time as additional coordinate to $\tilde X^{\leq \tau(t)}$.
	\item The user can choose which architectures to use for the neural networks  $f_{\theta_1}$, $\rho_{\theta_2}$ and $\tilde g_{\tilde \theta_3}$.
	\item By choosing whether to use a recurrent structure for $\rho_{\theta_2}$ and whether to use the signature as an input, the user can select PD-NJ-ODE, NJ-ODE, or either of the two intermediate versions. Whenever we refer to the PD-NJ-ODE model, we use both the signature and the recurrent structure.
	\item The user can choose to use gradient clipping or output clamping, which is not done by default.
	\item The solution to the SDE \eqref{equ:PD-NJ-ODE} is approximated with the Euler method for some step size $\Delta t$.
	\item Since the $2$-norms appearing in the objective functions \eqref{equ:Psi}--\eqref{equ:appr loss function} involve computing a square-root, which leads to differentiability issues at $0$, we add a small regularizing constant ($\epsilon = 10^{-10}$) to the quantities in the square-roots.
\end{itemize}

\subsubsection{Remarks on the signature}
First, we note that the ``not-tree-like'' condition is easily satisfied in practice by including a monotonically increasing coordinate in the data.
To this end, we concatenate the current time to the data as an additional dimension, which is standard practice when working with signatures (see, for example, \citet{fermanian2020embedding}).
If the function to be learnt in Theorem~\ref{thm:universal_approx_sig} depends significantly on higher degree terms, and the truncation level is chosen too low, some information will inevitably be lost.
For a more elaborate setup, \cite{deep_signature_transform} discuss methods to reduce this issue by applying an augmentation to the original data stream before computing the signature.
In our applications, we address this issue differently. We use a recurrent structure, which can learn to extract and store certain path-dependent features of the data.
In particular, it could learn to approximate a higher-degree term of the signature that carries significant information.

Several Python packages are available for computing the signatures of the interpolated paths considered here, including \textit{ESig}, \textit{iisignature}, and \textit{signatory}. In our implementation, we use the \textit{iisignature} package developed by \cite{reizenstein2018iisignature}.

\subsubsection{Synthetic datasets}\label{sec:Synthetic Datasets}
We use the same method to generate datasets as in \citet{herrera2021neural}. In particular, if not mentioned otherwise, we consider the time interval $[0,1]$, i.e., $T=1$, and a discretization time grid with step size $0.01$ which leads to $101$ time points on this interval. For each path $i$ and each of these time points $t$, an independent Bernoulli random variable $O_{i,t} \sim Bernoulli(p)$ is drawn and the time point is used as observation time if $O_{i,t} =1$. Our standard choice for the success probability is $p=.1$.
The paths themselves are sampled using either a closed-form method, if available, or the Euler scheme adapted to the discretization time grid. By default, we sample $20'000$ paths, which are split into an $80\%$ training set and a $20\%$ test set.

For each observation time (where $O_{i,t}=1$) in synthetic datasets with incomplete observations (masked data), an independent random variable $N_{i,t} \sim 1 + Poisson(\lambda)$ is first drawn to specify the number of coordinates observed at that time. Then $\min(d_X, N_{i,t})$ coordinates are drawn at random (without replacement) from the set of all coordinates to be observed.
If $\lambda = 0$, then exactly one (randomly chosen) coordinate is observed at each observation time.

\subsubsection{Training}\label{sec-app:Training}
We always use the Adam optimizer with the standard choices $\beta = (0.9, 0.999)$ and weight decay of $0.0005$. If not mentioned otherwise, our standard choice for the learning rate is $0.001$, a dropout rate of $0.1$ is used for every layer and training is performed with a mini-batch size of $200$ for $200$ epochs.

\subsubsection{Objective function}\label{sec:Loss Function details}
The (original) loss function \eqref{equ:Psi},
\begin{equation*}
\Psi: \mathbb{D} \to \R, \nonumber
Z \mapsto \Psi(Z) := \E_{\P\times\tilde\P}\left[ \frac{1}{n} \sum_{i=1}^n  \left(  \left\lvert M_t \odot ( X_{t_i} - Z_{t_i} ) \right\rvert_2 + \left\lvert M_t \odot (Z_{t_i} - Z_{t_{i}-} ) \right\rvert_2 \right)^2 \right],
\end{equation*}
uses the distance between the predictions before and after the jump to make the neural ODE learn the correct behaviour between jumps.
In contrast, the equivalent loss function (cf. Remark~\ref{rem:equivalent objective function}),
\begin{equation*}
\tilde \Psi: \mathbb{D} \to \R, \nonumber
Z \mapsto \Psi(Z) := \E_{\P\times\tilde\P}\left[ \frac{1}{n} \sum_{i=1}^n  \left(  \left\lvert M_t \odot ( X_{t_i} - Z_{t_i} ) \right\rvert_2 + \left\lvert M_t \odot (X_{t_i} - Z_{t_{i}-} ) \right\rvert_2 \right)^2 \right],
\end{equation*}
uses the distance between the observation and the prediction before the jump.
As shown in the left panel of Fig.~\ref{fig:LossFuncComp}, the original loss function can have local minima at which the model does not jump to $X_{t_i}$ upon receiving a new observation but instead jumps to a point between $X_{t_i}$ and $Z_{t_{i}-} $, where the term $(Z_{t_i} - Z_{t_{i}-} ) $ is smaller than it would be if $Z_{t_i} = X_{t_i}$.
When the weights are stuck at this local minimum, the model achieves a minimal evaluation metric of $1.8 \cdot 10^{-2}$.
For the equivalent loss function, the value $Z_{t_{i}-}$ before the jump is compared directly with $X_{t_i}$, so jumping to a point $Z_{t_i} \ne X_{t_i}$ provides no local advantage (right panel of Fig.~\ref{fig:LossFuncComp}). The model achieves a minimal evaluation metric of $1.0 \cdot 10^{-2}$, an improvement by nearly a factor of $2$.
The used models here are NJ-ODEs with the signature as additional input, both with the same architecture for each neural network  with latent dimension $d_H = 50$ and 2 hidden layers with $\tanh$ activations and $200$ nodes.
\begin{figure}[htp!]
\centering
\includegraphics[width=0.49\textwidth]{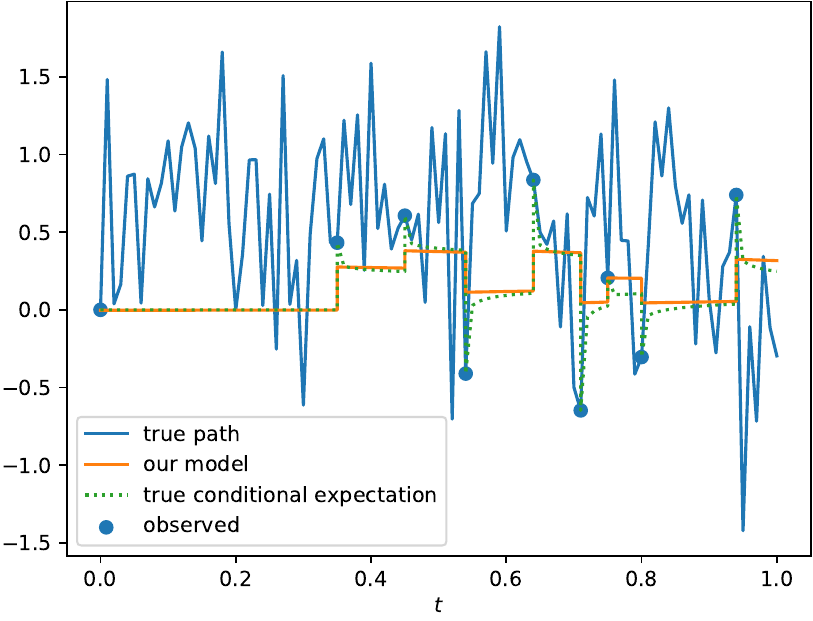}
\includegraphics[width=0.49\textwidth]{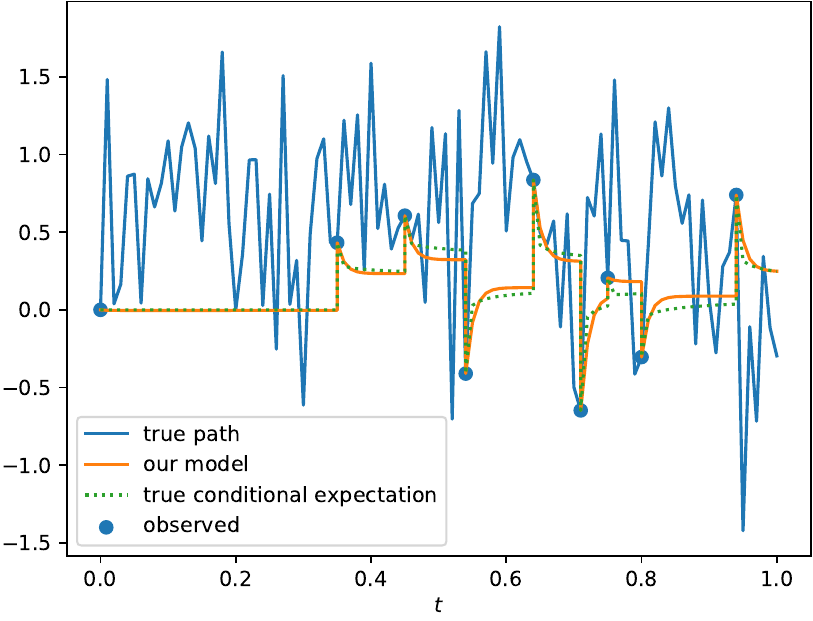}
\caption{The same test sample of a fractional Brownian motion, when NJ-ODE (with signature) is trained with the original loss function (left) and the equivalent loss function (right). The equivalent loss function prevents NJ-ODE from ending up in local minima.}
\label{fig:LossFuncComp}
\end{figure}

\subsection{Details for poisson point process}
\textbf{Dataset.}
The easiest way to sample a homogeneous Poisson point process is to use the fact that its interarrival times are i.i.d. exponential random variables with mean $1/\lambda$. Hence, for each path, we sample as many i.i.d. random variables $E_k \sim Exp(\lambda)$ as needed such that $\sum_k E_k \geq T$. The cumulative sums $(\sum_{k \leq i} E_k)_{i \geq 1}$ are then the time points at which the process increases by $1$. Between these points, the process is constant, and its starting point is $0$. Finally, the standard discretization time grid is applied to obtain one sample of the dataset.

\textbf{Architecture.}
We use the standard NJ-ODE with the following architecture. The hidden size is $d_H = 10$, and all three neural networks have the same structure: two hidden layers with a $\tanh$ activation function and $50$ nodes.

\subsection{Details for uncertainty estimation}
\textbf{Architecture.}
Since the dataset is relatively simple, we test a less complex network to show that it can already be sufficient. In particular, we use the standard NJ-ODE with the following architecture. The hidden size is $d_H = 50$, the readout network is a linear map, and the other two neural networks have the same structure: one hidden layer with a $\tanh$ activation function and $50$ nodes.

\subsection{Details for dependent observation intensity}\label{sec:Details for Dependent Observation Intensity}
\textbf{Dataset.}
We use the Euler scheme to sample paths from a Black--Scholes model (geometric Brownian motion) with drift $\mu = 2$, volatility $\sigma = 0.3$, and starting value $X_0 = 1$.

\textbf{Architecture.}
We use the PD-NJ-ODE with the following architecture. The latent dimension is $d_H = 50$, and all three neural networks have the same structure; two hidden layers with a $\tanh$ activation function and $50$ nodes.
The signature is used up to truncation level $3$.

\subsection{Details for fractional Brownian motion}\label{sec:Details for Fractional Brownian Motion}
\textbf{Architecture.}
We compare NJ-ODE, NJ-ODE with a signature, NJ-ODE with a recurrent jump network, and PD-NJ-ODE, all using the following architecture. The latent dimension is $d_H = 50$, and all three neural networks have the same structure: two hidden layers with a $\tanh$ activation function and $200$ nodes.
All truncation levels $m \in \{ 1, \dotsc, 10\}$ were tested, and level $3$ gave the best trade-off between model performance and computation time. Therefore, results are shown only for truncation level $3$.

\subsection{Details for correlated 2-dimensional Brownian motion }
\textbf{Dataset.} We use the same methods as described in Appendix~\ref{sec:Synthetic Datasets}, but to account for the higher complexity of the dataset, we generate $100'000$ samples instead of only $20'000$.

\textbf{Architecture.}
We use the PD-NJ-ODE with the following architecture. The latent dimension is $d_H = 100$, the readout network is a linear map, and the other two neural networks have the same structure; one hidden layer with a $\tanh$ activation function and $100$ nodes. The signature is used up to truncation level $2$. This architecture achieves a minimal evaluation metric of $5.2 \cdot 10^{-3}$. Changing to $\operatorname{ReLU}$ activation functions nearly halves this value to $2.8 \cdot 10^{-3}$.

\subsection{Details for stochastic filtering}\label{sec:Details for Stochastic Filtering}
\textbf{Dataset.} We use the same methods as described in Appendix~\ref{sec:Synthetic Datasets}, but to account for the higher complexity of the dataset, we generate $40'000$ samples instead of only $20'000$ and we generate training and testing datasets separately. In the training dataset, the $Y$-coordinates are observed at every observation time, while the $X$-coordinates are observed with probability $p_k = 0.25$ (for each observation time a $Bernoulli(p_k)$ random variable is drawn).
In the test set, with $4'000$ samples, the $Y$-coordinate is again observed at every observation time and the $X$-coordinate is never observed.

\textbf{Architecture.}
We use the PD-NJ-ODE with the following architecture. The latent dimension is $d_H = 200$, and all three neural networks have the same structure; one hidden layer with a $\tanh$ activation function and $100$ nodes. The signature is used up to truncation level $2$.

\subsection{Details for double pendulum}\label{sec:Details for Double Pendulum}
\textbf{Dataset.}
We use the same methods as described in Appendix~\ref{sec:Synthetic Datasets}, but with $T=2.5$ and step size $0.025$, which leads again to $101$ time points.
We chose a larger step size than usual because the pendulums would otherwise move very slowly, and the resulting dataset would not appear very chaotic when sampled over $100$ steps. However, it is important not to choose a step size that is too large, as the model would then be unable to observe and learn the dynamics at a sufficiently fine scale.
To illustrate this, we generate a second dataset with $101$ time points, where $T=10$ and the step size is $0.1$. A comparison of the trained models on test samples from the two datasets is given in Fig.~\ref{fig:Double Pendulum Experiment 2}, which shows that the model cannot learn and reproduce the more volatile parts of the longer trajectories as accurately.

\textbf{Architecture.}
We use the PD-NJ-ODE, with the following architecture.
The latent dimension is $d_H = 400$, and all three neural networks have the same structure; one hidden layer with a $\tanh$ activation function and $200$ nodes. Empirically, the model performed best when using the recurrent jump network but no signature terms as inputs.

\begin{figure}
\centering
\includegraphics[width=0.49\textwidth]{images/double-pendulum-small_step-path-4}
\includegraphics[width=0.49\textwidth]{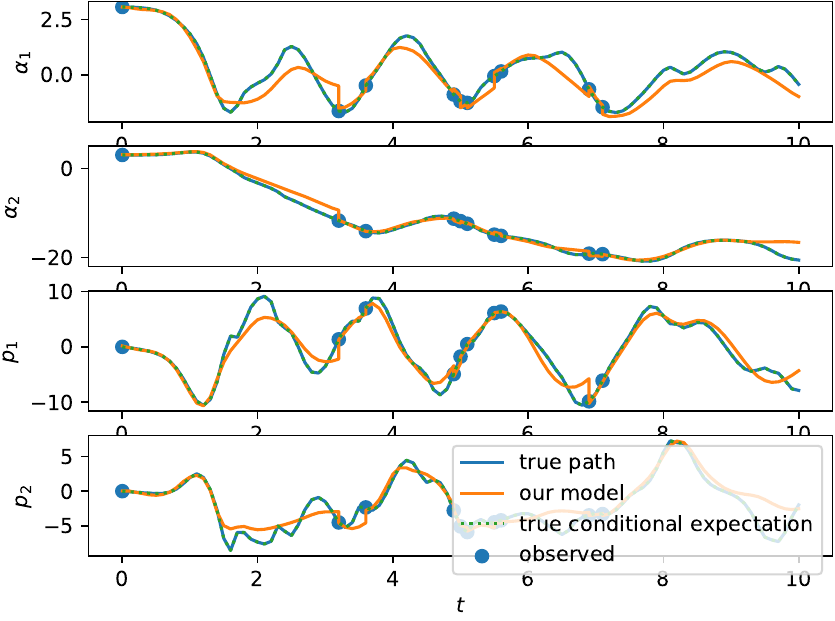}
\caption{Predicted conditional expectation on test samples of the Double Pendulum dataset. Since the system is deterministic, the true conditional expectation coincides with the true path. Left: dataset with $T=2.5$. Right: dataset with $T=10$. Both using $100$ steps.}
\label{fig:Double Pendulum Experiment 2}
\end{figure}

\subsection{Details for Brownian motion and its time-lagged version}
\textbf{Dataset.} We use the same methods as described in Appendix~\ref{sec:Synthetic Datasets}, but we generate the training and test datasets separately. In both datasets, the $X$-coordinate is observed at every observation time $t_i$, and the $Y$-coordinate is observed at $t_i + \alpha$, where $\alpha = 0.19$.
In the test set, with $4'000$ samples, the $Y$-coordinate is not observed at any additional time points.
Only in the training dataset, the $Y$-coordinate is additionally observed with probability $0.5$ at any $t_i+ \Delta t$, where $\Delta t$ is the used step size. We note that this is not general enough to satisfy the assumptions (cf. Section~\ref{sec:Predicting a Time-Lagged Version of an Observed Process}). However, it suffices for the model to learn the correct behaviour and empirically leads to better results than using a dataset, where the $Y$-coordinate could be observed at all times between $t_i$ and $t_i+ \alpha$. Indeed, the optimal evaluation metric on this dataset was $1.2 \times 10^{-3}$.

\textbf{Architecture.}
We use the PD-NJ-ODE with the following architecture. The latent dimension is $d_H = 400$, the readout network is a linear map, and the other two neural networks have the same structure; one hidden layer with a $\operatorname{ReLU}$ activation function and $200$ nodes. The signature is used up to truncation level $2$.

\subsection{Details for PhysioNet}
\textbf{Dataset.} Details on the dataset are given in \citet[Appendix F.5.3]{herrera2021neural}. The exact same setup is used.

\textbf{Architecture.}
We use the PD-NJ-ODE with the following architecture. The hidden size is $d_H = 50$, and all neural networks have the same structure; one hidden layer with a $\tanh$ activation function and $50$ nodes. The signature is used up to truncation level $2$. Because the network size grows exponentially with the signature truncation level, no larger levels were tested.

\textbf{Training.} The training was done as specified in Section~\ref{sec-app:Training}, except that a batch size of $50$ was used for $175$ epochs.
Five runs of the same network with random initializations were performed, and their mean and standard deviation were computed.

\textbf{Results.} We report the minimum MSE on the test set over the $175$ epochs. If, instead, the MSE from the epoch with the minimum training loss is reported, the result is $1.957 \pm 0.018 (\times 10^{-3})$, which also outperforms NJ-ODE.

\subsection{Details for limit order book dataset}
\textbf{Datasets.}
The widely used benchmark dataset for midprice forecasting \citep{ntakaris2018benchmark} is unfortunately unsuitable in our context because timestamps (or time differences between LOB updates) are not included. Hence, the PD-NJ-ODE model could not be applied without adding an artificial time variable.\\
Therefore, we test our model on the cryptocurrency LOB datasets ``BTC'', ``BTC1sec'', and ``ETH1sec'', as described in Section~\ref{sec:Limit Order Book Data}.
The datasets contain $10'297$ (``BTC''), $8'669$ (``BTC1sec''), and $8'753$ (``ETH1sec'') samples, of which the first $80\%$ are used for training and the final $20\%$ for testing (thereby avoiding look-ahead bias).
While ``BTC'' is based on the complete LOB for one day (July 2, 2020), ``BTC1sec'' and ``ETH1sec'' are based on snapshots recorded once per second over roughly 12 days (April 7--19, 2021). The median time step in the ``BTC'' dataset is approximately $ 0.025$, which means that, on average, the order book is updated $40$ times per second.
However, not every update affects the first $10$ levels of the order book in which we are interested. Overall, comparing the number of samples of the datasets, we can conclude that there are roughly $ 14$ updates to the first $10$ levels of the order book each second. \\
The labels (increase, decrease, and stationary) for the classification task are computed as outlined in \citet[Equation~4]{zhang2019deeplob}, where the threshold $\alpha$ is chosen as the empirical $\frac{2}{3}$-quantile of the dataset so that there are roughly equal numbers of samples for each label. Importantly, the labels are computed before any other preprocessing is applied to the data, as such preprocessing might otherwise lead to different labels.\\
For DeepLOB, the dataset is normalized using $z$-scores, as in \citet{zhang2019deeplob}. For PD-NJ-ODE, the dataset is not normalized; instead, each sample is shifted so that it starts at $X_0=0$ (this shift worsens DeepLOB's performance, whereas not shifting the samples does not significantly change PD-NJ-ODE's performance). Moreover, time is shifted so that each sample starts at $t_0=0$. Furthermore, we use only the midprice and bid/ask prices up to level 10, not the volume, as inputs to PD-NJ-ODE.

\textbf{Architecture.}
We use the PD-NJ-ODE with the following architecture. The hidden size is $d_H = 100$, and all neural networks have the same structure; one hidden layer with a $\tanh$ activation function and $50$ nodes. The signature is used up to truncation level $2$, but no recurrent jump network is used.
On top of this architecture a classifier network, again with the same structure as the networks above, but with a final softmax activation (to produce outputs in $[0,1]$ that can be interpreted as probabilities), is used to map the last latent variable $H_{t_n}$ to the class probabilities $(p_{inc}, p_{stat}, p_{dec}) \in [0,1]^3$.
During the additional retraining of the classifier, we also test more complex classification networks, which yield better results. In particular, for the results shown in Table~\ref{tab:LOB F1 score}, we retrained three different classifier networks with the following architectures:
\begin{itemize}
\item 1 hidden layer with $\tanh$ activation function and $50$ nodes (same as in combined training),
\item 2 hidden layers with $\tanh$ activation functions and $200$ nodes,
\item 4 hidden layers with $\tanh$ activation functions and $200$ nodes,
\end{itemize}
and report the results of the best-performing one, although all architectures achieve very similar results (less than a $3\%$ deviation from the reported F1-scores).

\textbf{Training and Evaluation.}
The model is first trained for $50$ epochs, where the sum (without weighting) of the PD-NJ-ODE loss and the cross-entropy loss of the classifier is minimized with the standard Adam optimizer. The batch size is chosen to be $50$ and the learning rate is $0.01$.
The model is trained to forecast all inputs, i.e., the midprice and all bid and ask prices up to level $10$; however, only the midprice is considered when evaluating the MSE.\\
The  classifier is retrained for $1000$ epochs with the cross-entropy loss alone, also with the standard Adam optimizer. The batch size is again $50$ and the learning rate is $0.001$.

\textbf{Training of baseline model DeepLOB.}
We use the training procedure that was suggested in \citet{zhang2019deeplob}. In particular, the model is trained with a batch size of $64$ for $50$ epochs and learning rate $0.0001$.

\textbf{Training of linear regression and random forest baseline models.}
Standard linear regression models using an intercept and without regularization are used. The random forest regression models all use $100$ trees with a maximum depth of $8$.

\textbf{Additional Results}
\begin{figure}[htp!]
\centering
\includegraphics[width=0.49\textwidth]{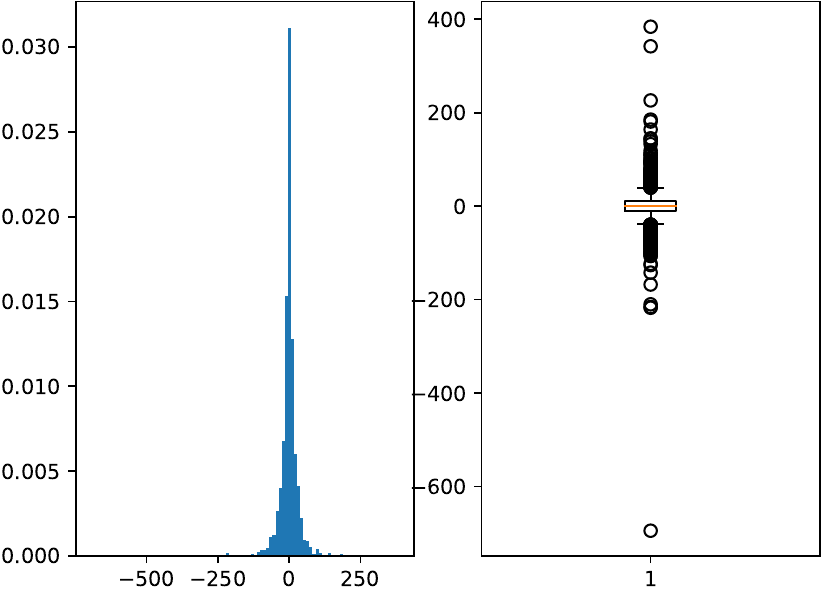} \hspace{0.2cm}
\includegraphics[width=0.475\textwidth]{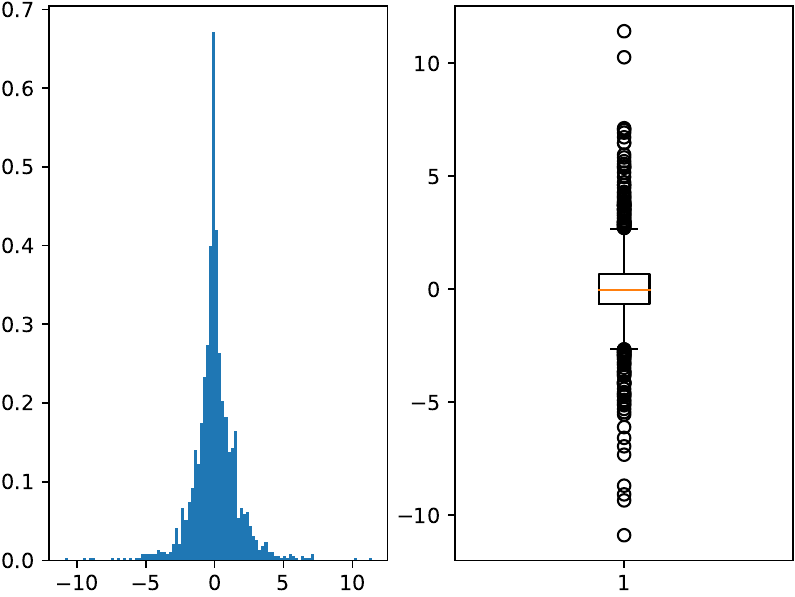}
\caption{ 
Distribution of the prediction errors of the PD-NJ-ODE model  as (density) histogram and boxplot on the BTC1sec (left) and ETH1sec (right) dataset.
}
\label{fig:LOB error distr BTC1sec and ETH1sec}
\end{figure}

\section{Direct Neural Network Approximation of the Conditional Expectation}\label{sec:Direct Neural Network Approximating of the Conditional Expectation}
Revisiting the setting and assumptions in Section~\ref{sec:Problem Setting}, one might ask why we do not approximate the conditional expectation directly by approximating the functions $F_j$ for $1 \leq j \leq d_X$. The assumption that all functions $F_j$ are continuous is needed in any case; hence, using neural networks to approximate them should, in principle, be possible within this setting. Moreover, in our approach, we also approximate the functions $F_j$ using the neural network $\rho$, although only at observation times.
From a theoretical perspective, the clear benefit of this direct approximation is that it works under weaker assumptions. In particular, continuity of $F_j$, rather than continuous differentiability in its first coordinate $t$, would suffice.\\
By contrast, our approach uses additional domain knowledge; namely, that the learning problem can be split into two intertwined subproblems, one of which is to learn the continuous evolution between any two observation times and the other to learn the updates (jumps) at observation times when new information becomes available.
For the continuous learning part, we have the additional knowledge that this is the limit of a recursive problem, which amounts to learning the neural ODE network. In particular, instead of learning the function at every point in time, we only need to learn the dynamics that update the value of the function through time.

In practice, using this domain knowledge makes the learning task much easier.
To quantify this, we compare PD-NJ-ODE with the direct approximation of $F_j$ by a neural network (which we refer to as the NJ-model) on the simple Black--Scholes (geometric Brownian motion) dataset. While PD-NJ-ODE achieves a minimal evaluation metric of $5 \times 10^{-4}$ on the test set, the NJ-model achieves $1 \times 10^{-2}$, i.e., a value larger by a factor of $20$.

\subsection{Implementation details}
\textbf{Dataset.}
We use a Black--Scholes model with the  parameters  described in Section~\ref{sec:Details for Dependent Observation Intensity} and sample the observation times as described in Section~\ref{sec:Synthetic Datasets}.

\textbf{Architecture.}
For the PD-NJ-ODE, we use the following architecture. The latent dimension is $d_H = 50$, and all three neural networks have the same structure; one hidden layer with a $\tanh$ activation function and $100$ nodes. The signature is used up to truncation level $2$. This architecture has approximately $27$K trainable parameters.\\
For the NJ-model, we use only the neural network $\rho$, which maps directly to the output space, i.e., without a latent space and thus without a readout map. We first test the same network as above: one hidden layer with a $\tanh$ activation function and $100$ nodes. When using the network structure suggested by the functions $F_j$, i.e., without a recurrent structure and with only the signature as an input, the minimal evaluation metric is $8 \times 10^{-2}$. A recurrent structure similar to that in PD-NJ-ODE, which uses the output from the previous time as an input, improves this value to $2 \times 10^{-2}$. We note that this network structure also uses some domain knowledge, but in a way for which the weaker assumptions remain sufficient. Since this model has only about $1$K trainable parameters, comparing it with the PD-NJ-ODE model above might be considered unfair. Therefore, we test two larger architectures: first, a network with one hidden layer, a $\tanh$ activation function, and $2000$ nodes (resulting in $24$K trainable parameters); and second, a network with two hidden layers, each with a $\tanh$ activation function and $200$ nodes (resulting in $42$K trainable parameters). While the first architecture did not improve the results, the second led to the reported evaluation metric of $1 \times 10^{-2}$.
\fi

\end{document}